\documentclass[journal]{IEEEtran}
\usepackage{graphicx}
\usepackage{subfloat}

\usepackage{float}
\usepackage{amsmath}
\usepackage{CJK}
\usepackage{amssymb}
\usepackage{indentfirst}
\usepackage{mathtools}
\usepackage{subfigure}
\usepackage{stfloats}
\usepackage{mathrsfs}
\usepackage{bm}
\usepackage{blindtext}
\usepackage{algorithm}
\usepackage{algpseudocode}
\usepackage{amsmath}

\usepackage{amsmath,amssymb}
\DeclareSymbolFont{EulerExtension}{U}{euex}{m}{n}
\DeclareMathSymbol{\euintop}{\mathop} {EulerExtension}{"52}
\DeclareMathSymbol{\euointop}{\mathop} {EulerExtension}{"48}

\usepackage[Symbol]{upgreek}
\usepackage{booktabs}
\usepackage{multirow}

\usepackage{array}
\newcommand{\PreserveBackslash}[1]{\let\temp=\\#1\let\\=\temp}

\ifCLASSINFOpdf
\else
\fi
\begin{document}
%

\title{An Ensemble Rate Adaptation Framework for Dynamic Adaptive Streaming over HTTP}
%
%
%

\author{Hui~Yuan,~\IEEEmembership{Senior Member,~IEEE,}
        Xiaoqian~Hu,
        Junhui~Hou,~\IEEEmembership{Member,~IEEE,}
        Xuekai~Wei,
        and~Sam~Kwong,~\IEEEmembership{Fellow,~IEEE}
\thanks{Manuscript Received.}
\thanks{This work was supported in part by the National Key R\&D Program of China under Grants 2018YFC0831003, in part by the National Natural Science Foundation of China under Grants 61571274 and 61871342; in part by the Shandong Natural Science Funds for Distinguished Young Scholar under Grant JQ201614; in part by the Young Scholars Program of Shandong University (YSPSDU) under Grant 2015WLJH39.}
\thanks{H. Yuan is with the School of Control Science and Engineering, Shandong University, Ji'nan 250061, China (Email: huiyuan@sdu.edu.cn).}
\thanks{X. Hu is with the School of Information Science and Engineering, Shandong University, Qingdao, 266237, China (Email: huxiaoqian11@163.com).}
\thanks{J. Hou (corresponding author), X. Wei and S. Kwong are with the Department of Computer Science, City University of Hong Kong, Kowloon, Hong Kong (Email: jh.hou@cityu.edu.hk, xuekaiwei2-c@my.cityu.edu.hk, and cssamk@cityu.edu.hk).}

\thanks{\textbf{Copyright (c) 2019 IEEE. Personal use of this material is permitted. However, permission to use this material for any other purposes must be obtained from the IEEE by sending an email to pubs-permissions@ieee.org.}}}

\maketitle
\begin{abstract}
Rate adaptation is one of the most important issues in dynamic
adaptive streaming over HTTP (DASH). Due to the frequent
fluctuations of the network bandwidth and complex variations of
video content, it is difficult to deal with the varying network
conditions and video content perfectly by using a single rate
adaptation method. In this paper, we propose an ensemble rate adaptation framework for DASH, which aims to leverage the advantages of multiple methods involved in the framework to improve the quality of experience
(\emph{QoE}) of users. The proposed framework is simple yet very
effective. Specifically, the proposed framework is composed of two
modules, i.e. the method pool and method controller. In the method
pool, several rate adaptation methods are integrated. At each
decision time, only the method that can achieve the best \emph{QoE}
is chosen to determine the bitrate of the requested video segment.
Besides, we also propose two strategies for switching methods, i.e.,
InstAnt Method Switching, and InterMittent Method Switching, for the method controller to determine which method can provide the best \emph{QoE}s. Simulation results demonstrate that, the proposed framework always achieves the highest
\emph{QoE} for the change of channel environment and video
complexity, compared with state-of-the-art rate adaptation methods.
\end{abstract}


%
\IEEEpeerreviewmaketitle


\section{Introduction}
\IEEEPARstart{I}{n} the last few years, video streaming has become the dominant source of traffic over the Internet. To face this growing demand of media traffic, new efficient video streaming techniques have been developed, such as the popular Dynamic Adaptive Streaming over HTTP (DASH) [1].  In contrast to RTP/UDP, DASH is easy to configure and, in particular, can greatly simplify the traversal of firewalls and network address translators. In addition, it can be easily deployed with content delivery networks at a relatively low cost. Therefore, DASH has been widely adopted for providing uninterrupted video streaming services for users [2].

\begin{figure}
\centering
\includegraphics[width=8.76cm]{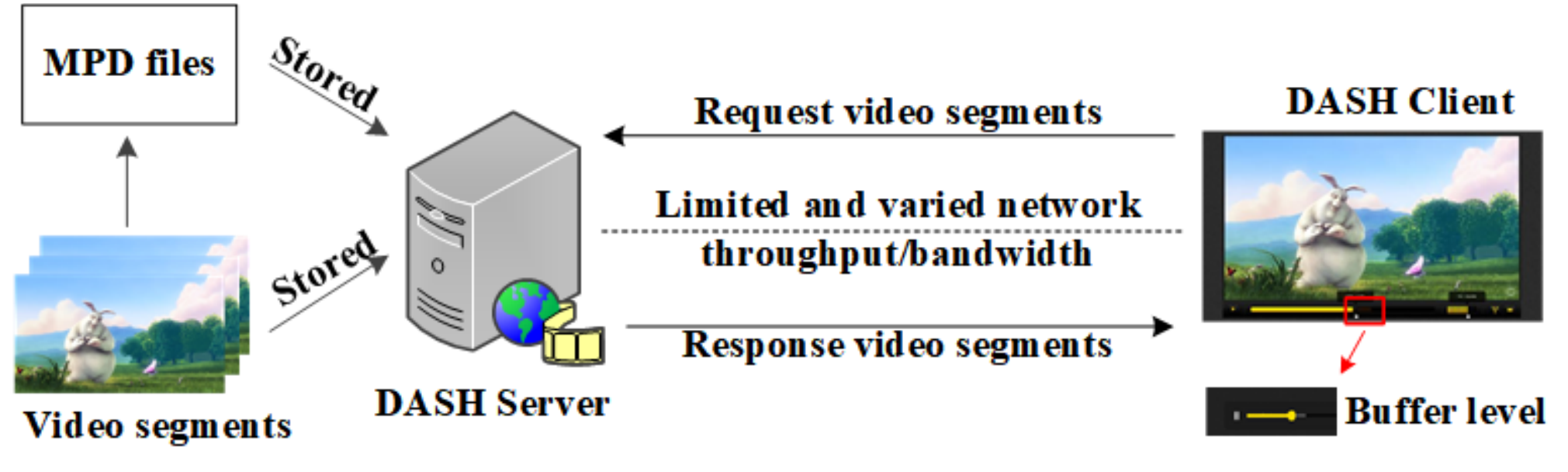}
\caption{Structure of a DASH based video delivery system.}
\label{fig1}
\end{figure}

The key concept of DASH is that each video is encoded into several
representations with different bitrates. These representations are
then divided into small segments or chunks (typically with display
time of 1-10 seconds). As shown in Fig. 1, at the beginning of a DASH
connection, a manifest file (MPD file) that records the information
of all the available chunks of a video, e.g., URL addresses, chunk
lengths, quality levels, resolutions, etc., is first downloaded by
the client. Then the client will download the MPD file and
dynamically request video segments with various bitrates based on
its rate adaptation logic. It is noteworthy that the rate adaptation
logic of clients is not specified in the DASH protocol. Naive DASH
adaptation logics usually suffer from large quality variations or
playback buffer underflow [3], which drastically deteriorates the
quality of experience (\emph{QoE}) of users [4]. Because the network
bandwidth is highly dynamic, it is challenging to provide stable and
high quality videos to ensure satisfactory user experience all
the time. Therefore, how to design appropriate rate adaptation
methods to improve user \emph{QoE} is one of the most important
research topics (see Section \uppercase\expandafter{\romannumeral2}
for details) for DASH.

Different from the traditional quality of service (\emph{QoS})-based
video streaming like RTP/UDP, DASH is designed as a
\emph{QoE}-compatible streaming protocol which can satisfy the
greedy requirements of users by using appropriate rate adaptation
methods. Arising from the characteristics of human visual system
[5], \emph{QoE} is affected by lots of factors, e.g., the initial
playback delay, the received video quality, smoothness, the number
and duration of playback stalling events, the instantaneous
interactions between playback stalling, video quality [6-12].

Due to the complexity of network variations, it is difficult to
utilize a single method to adapt to the network variations
perfectly for all time. Therefore, to achieve the optimal
\emph{QoE} for users as much as possible, we propose an ensemble rate
adaptation framework for DASH. Our contributions are as follows.
\begin{enumerate}
  \item We propose an ensemble rate adaptation framework for DASH to adapt complex network variations in practice, such that the optimal QoE for users can be achieved.
  \item We propose two control strategies to select the best method at different decision times to adapt to diverse network conditions.
  \item To fit the proposed framework, we also propose a QoE model by taking the average video quality, the quality variation, re-buffering events, and the initial playout delay into account.
\end{enumerate}

The rest of the paper is organized as follows. Related work is
discussed in Section \uppercase\expandafter{\romannumeral2}. The
proposed framework is presented in detail in Section
\uppercase\expandafter{\romannumeral3}. Simulation results and
conclusions are given in Sections
\uppercase\expandafter{\romannumeral4} and
\uppercase\expandafter{\romannumeral5}, respectively.


\section{Related Work}
Existing rate adaptation methods for DASH can be roughly divided into two
categories, i.e., content-dependent methods (usually for multi-view videos and 360-degree videos)[13-16] and content-independent methods [6][17-34]. The
content-independent methods can be further classified into two
categories, i.e. methods for a single server [6][17-35] and methods
for multi-servers [18][36][37]. Since only content-independent
methods with a single server are investigated in this paper, only
those related works are discussed.

Dr\"{a}ler and Karl [3] introduced the quality-first-based method.
In this method, users will always request the video segment whose
bitrate best matches the estimated channel bandwidth. If there is
not enough channel capacity, it tries to download segments with the
next lowest video quality level, until there is no capacity or
buffer space left. This algorithm favors downloading segments at
higher video quality levels at the expense of having more buffering
segments. On the contrary, Huang \emph{et al}. [19][20] proposed a
buffer-first-based adaptation method. In this method, users will
always consider filling a predefined buffer length (evaluated by the remaining playout time of the downloaded video segments). When the buffer length is larger than the predefined buffer length, video segments with higher bitrates will
be requested. The buffer-first-based method avoids the challenge of
estimating channel bandwidths, and stabilizes the buffer occupancy
to ensure smooth video playback. However, such method usually
requests video segments with lower quality than the
quality-first-based method.

Meng \emph{et al}. [21] proposed a proportional-integral-derivative
(PID)-based rate adaption method for DASH. The performance of this
method is overwhelming for channels with stable parameters, but it
is difficult to determine the appropriate control parameters for the
PID model when the channel state fluctuates. By considering the
influence of the current requested video segment on the next
requested video segment, Zhou \emph{et al}. [22] and Bokani \emph{et
al}. [23] proposed a Markov decision process (MDP)-based rate
adaptation methods, respectively, where the request
process for a video segment is considered to be an MDP. To maximize
the \emph{QoE}, Mart\'{\i}n \emph{et al}. [24] proposed a
Q-Learning-based rate adaptation method based on the
Markov process assumption, in which the requested video
segments can be dynamically adjusted based on the perceived network
state. In spite of the pretty good performance of the Q-Learning
based method, it needs a long time to converge. To overcome this limitation,
Claeys \emph{et al}. [25] proposed an HTTP Adaptive Streaming client
based on Frequency Adjusted Q-Learning (FAQ). They simplified the MDP model and improved the learning rate of the process. However, this simplification leads to suboptimal streaming performance. In other words, there is a tradeoff between
convergence speed and learning accuracy. Chiariotti \emph{et al}.
[17] proposed an online learning-based rate adaptation method. In
this method, the developed learning process exploits a parallel
learning technique that improves the learning rate and limits
sub-optimal choices, leading to a fast and accurate learning process
that converges to the best state rapidly. Unfortunately, it is hard to make the best request decision when the channel parameters frequently change.

All the methods are expected to achieve the optimal rate adaptation
and guarantee the best user \emph{QoE} by only using one
model. Due to the uncertainty of network throughput/bandwidth
requirements, a single model is unlikely to perform well at all
times. Therefore, we propose an ensemble rate adaptation framework for DASH.


\section{Proposed Ensemble Learning-based Framework}

\subsection{Overview of the Proposed Framework}

\begin{figure}
\centering
\includegraphics[width=8.76cm]{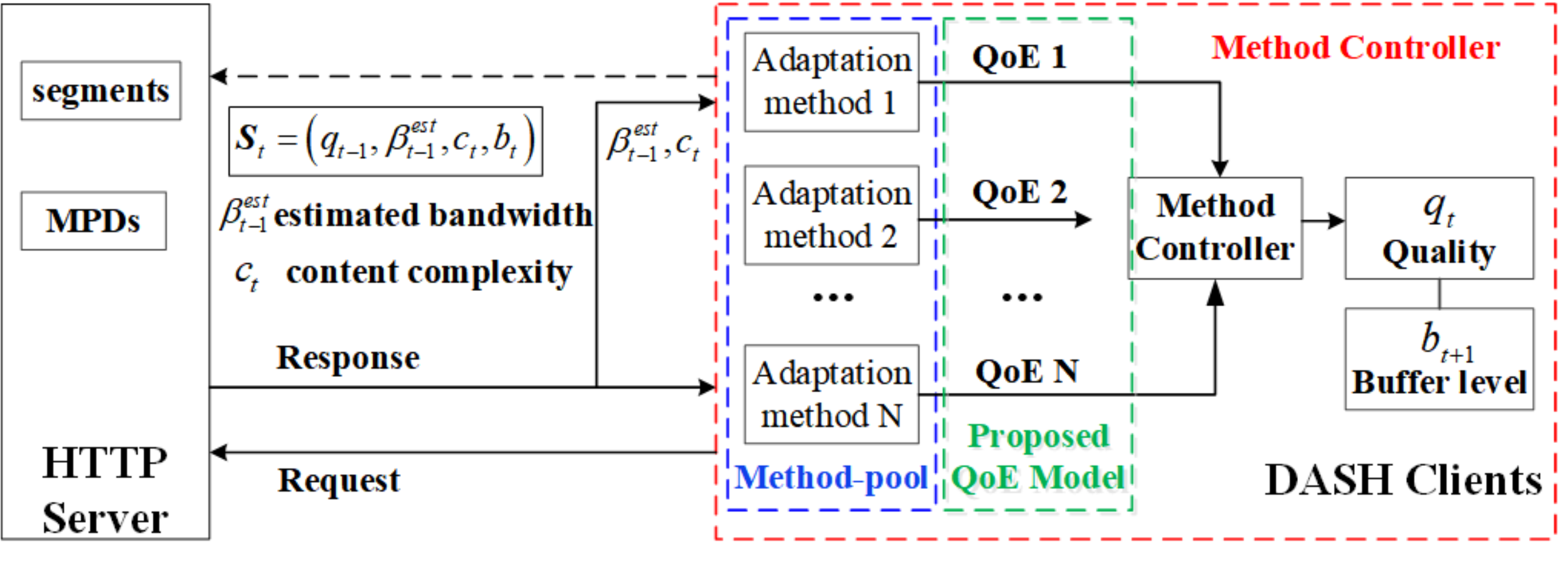}
\caption{Proposed ensemble learning based rate adaptation framework
for DASH.} \label{fig2}
\end{figure}

As shown in Fig. 2, the proposed framework is composed of two
modules: the method pool and the method controller. In the method pool,
several rate adaptation methods are integrated. At each decision
time, the method that can achieve the best \emph{QoE} is chosen to
determine the bitrate of the requested video segment. It is also worth pointing out  the remaining unselected methods will also update their parameters based on the actual system information (e.g., estimated throughputs, current buffer level, video complexity, etc.) to simulate their request procedure and calculate their
corresponding estimated \emph{QoE}s. The method controller will
finally decide which method is the best according to their
\emph{QoE}s. In the following sections, we will introduce the method
pool, the method controller, and the proposed QoE model in detail.

For clear representation, a summary of notations is given in TABLE I.

\begin{table}[htbp]
  \centering
  \caption{NOTATIONS}
\includegraphics[width=8.76cm]{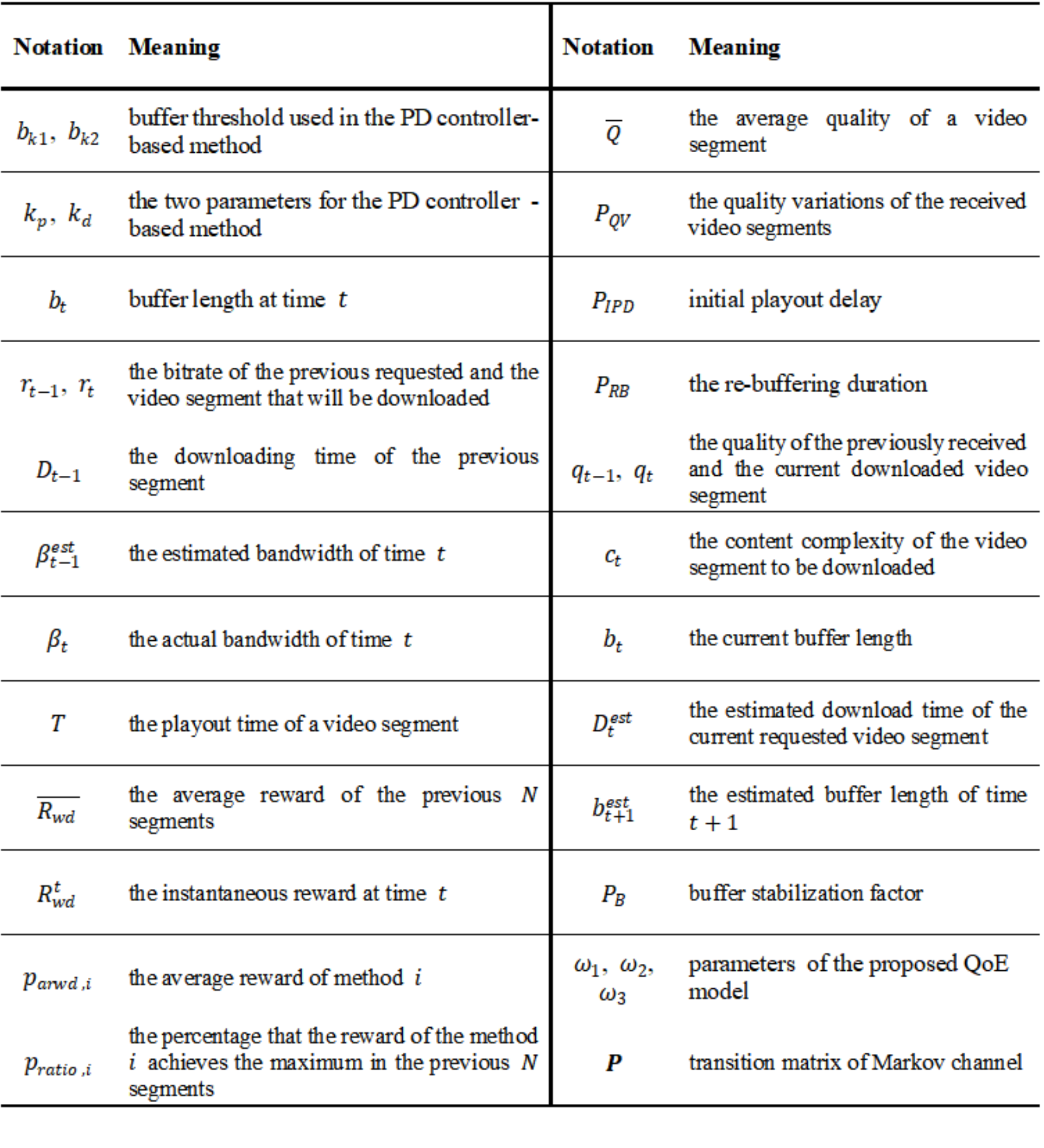}
\label{table1}
\end{table}

\subsection{Method Pool}
In this part, we introduce three methods as components of the method pool module of the proposed ensemble rate adaptation framework, i.e., a rate-based method [3], a proportion differentiation (PD) controller-based method [18], and an online learning-based method [17]. It is worth pointing out that we select these three methods as a typical implementation example, and other methods can also be integrated into the method pool.

\begin{itemize}
\item[(\emph{\romannumeral1})] \emph{The rate-based method}
\end{itemize}

The rate-based method will request the video segment whose bitrate
is closest to the estimated bandwidth [3]. As the rate-based method
is simple yet effective to some extent, it has been extensively used
in many adaptive streaming systems, e.g. \emph{Microsoft} Smooth
Streaming [38-39], \emph{Adobe} HTTP Dynamic Streaming [40] and
Apple's HTTP Live Streaming [41].

\begin{itemize}
\item[(\emph{\romannumeral2})] \emph{The PD controller-based adaptation method}
\end{itemize}

In this method, two buffer thresholds, $b_{k1}$ and $b_{k2}$, are
defined to restrict fluctuations of the user buffer. First, the
video segment with the lowest bitrate is downloaded. When the buffer
length at time $t$ (denoted by $b_{t}$) is larger than $b_{k1}$ and
smaller than $b_{k2}$, the requested bitrate will remain unchanged.
When the $b_{t}$ is less than $b_{k1}$, the requested bitrate will
be calculated as
\begin{equation}\label{E1}
r_{t}=r_{t-1}+\frac{1}{T}\times\beta_{t-1}^{est}[k_{p}\times(b_{t}-b_{k1})+k_{d}\times\frac{T-D_{t-1}}{D_{t-1}}],
\end{equation}
whereas, when the $b_{t}$ is larger than $b_{k2}$, the requested
bitrate will be calculated as
\begin{equation}\label{E2}
r_{t}=r_{t-1}+\frac{1}{T}\times\beta_{t-1}^{est}[k_{p}\times(b_{t}-b_{k2})+k_{d}\times\frac{T-D_{t-1}}{D_{t-1}}],
\end{equation}
where $r_{t-1}$ is the bitrate of the previous requested video
segment; $r_{t}$ is the bitrate of the video segment that will be
downloaded; $D_{t-1}$ is the downloading time of the previous
segment; $\beta_{t-1}^{est}$ is the estimated bandwidth; $T$ is the
playout time of a video segment, e.g., 2 seconds;  $k_{p}$ and
$k_{d}$ are the two parameters of the PD controller. Based on [18],
the two parameters $k_{d}$ and $k_{p}$ can be set as
\begin{equation}\label{E3}
\left\{
  \begin{array}{lr}
  k_{p}=\eta\times\sqrt{T^{2}-k_{d}^{2}},& \\
  \eta\geq\frac{1}{T}\sqrt{\frac{T+k_{d}}{T-k_{d}}}\ln\frac{20T}{T+k_{d}},&\\
  0<k_{d}<T.&
  \end{array}
\right.
\end{equation}

\begin{itemize}
\item[(\emph{\romannumeral3})] \emph{The online learning-based adaptation method}
\end{itemize}

By assuming that the channel bandwidth experienced by the users as
well as the video characteristic variation follows a Markov model,
the online learning-based method [17] formulates the optimal action
selection problem as a Markov Decision Process (MDP). The system
dynamics in the MDP model are a priori unknown and thus learned
through a Reinforcement Learning technique [17], in which the
\emph{QoE} of users is dynamically maximized.

At each decision time, the method controller of our framework will
select a method from the method pool to request a video segment so
that the \emph{QoE} of users can be maximized.

\subsection{Method Controller}

As aforementioned, there are multiple rate adaptation methods in the
method pool. For the first several decision times (each decision
time corresponds to a video segment), a default method is selected.
Then, at the subsequent decision times, the method controller selects one method to request a video segment. At the same time, the remaining unselected methods will also perform a "virtual request" based on the estimated bandwidth and then update their "virtual buffers". The method controller will always check the states of all
the rate adaptation methods and select the one that can achieve the
highest \emph{QoE}. In order to select an appropriate rate
adaptation method, we also design two strategies to enable the
controller to effectively switch from one method to another, i.e.,
\textbf{InstAnt Method Switching (IAMS)}, and \textbf{InterMittent
Method Switching (IMMS)}.

For the \textbf{IAMS} strategy shown in Fig. 3, the method controller will
always check the average reward (denoted as $\overline{R_{wd}}$) of
the previous $N$ decision times (segments) of all the methods in the
method pool, and then select the method that can achieve the maximum
average reward to request a video segment at the current decision
time. Note that the reward of a method in the previous decision time
may be ``virtual'' or ``actual'', which depends on whether the
method was selected at that decision time or not.

\begin{figure}
\centering
\includegraphics[width=8.76cm]{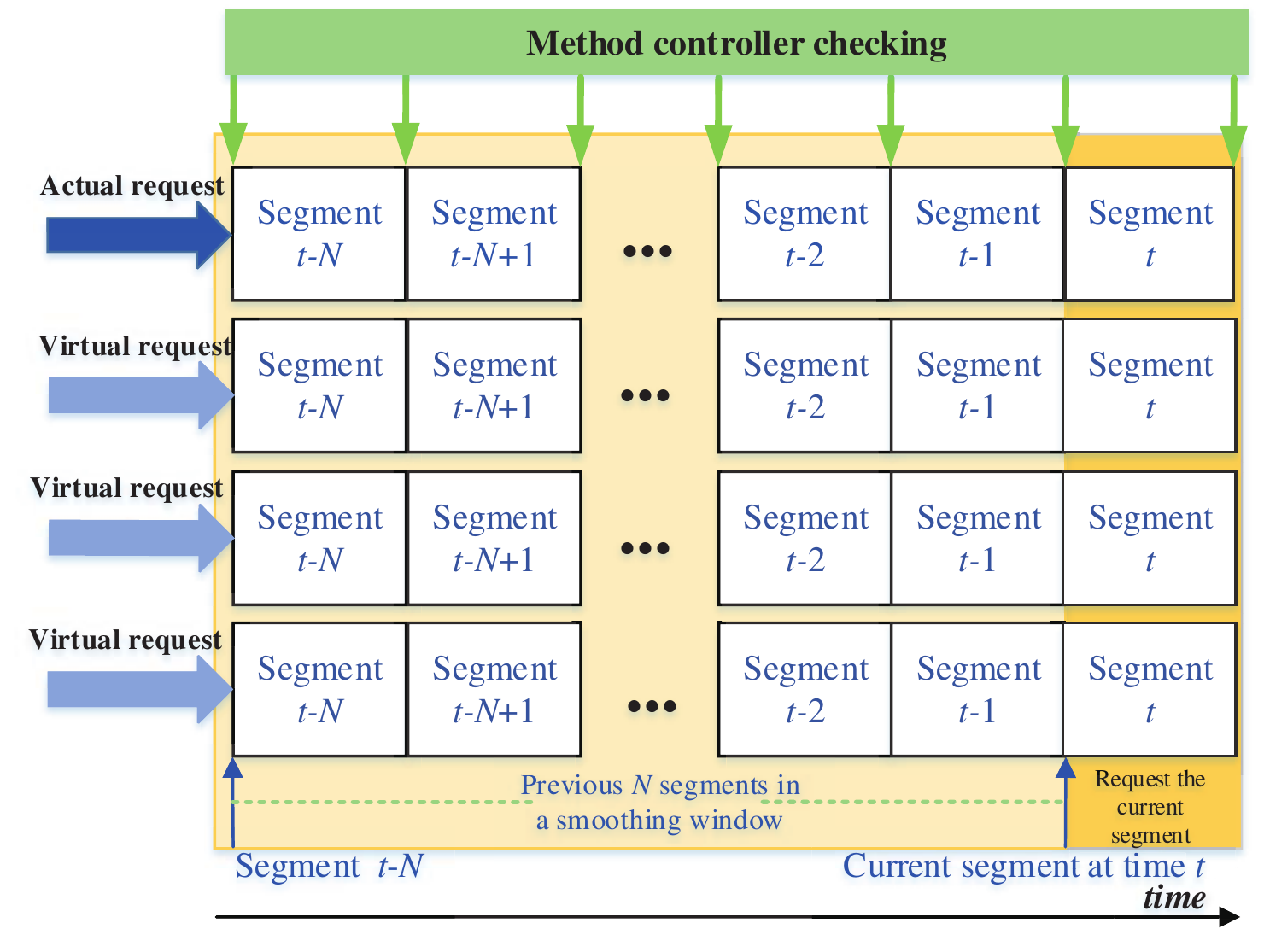}
\caption{Instant method switching.} \label{fig3}
\end{figure}

The \textbf{IAMS} strategy may result in frequent method switching.
Therefore, we also propose the \textbf{IMMS} strategy as shown in Fig. 4. In this method, the method controller will intermittently check the product of the
average reward of each method and the ratio that the reward of a
method achieves the maximum in the previous $N$ segments:
\begin{equation}\label{E4}
p_{product,i}=p_{arwd,i} \times p_{ratio,i},
\end{equation}
where $p_{arwd,i}$ denotes the average reward of method $i$, and
$p_{ratio,i}$ is the percentage that the reward of the method
$i$ achieves the maximum in the previous $N$ segments. Then, the
method controller will select the method that can achieve the
maximum product to request a video segment for the next $N$
segments.

\subsection{The proposed QoE Model}

The user \emph{QoE} must be considered and estimated well when
designing rate adaptation methods. Typically, by taking the visual
memory [42] into  into account, the user \emph{QoE} can be estimated
based on four factors [6-7] [17] [24-25] [43-45]: the average quality
level, the quality variations during the video playout, the video
freezes, and the startup delay. By investigating the existing
\emph{QoE} models [6-7] [24-25] [43-49], the \emph{QoE} model can be
written as
\begin{equation}\label{E5}
QoE=\overline{Q}-\omega_{0}P_{IPD}-\omega_{1}P_{QV}-\omega_{2}P_{RB},
\end{equation}
where $\overline{Q}$ and $P_{QV}$ denote the average quality and the
quality variations of the received video segments, respectively;
$P_{IPD}$ and $P_{RB}$ denote the initial playout delay and the re-buffering time, respectively; $\omega_{0}$, $\omega_{1}$, and $\omega_{2}$ are the model parameters.

Based on Eq.(5), to estimate the user \emph{LT-QoE}, the system state that can be derived from $\overline{Q}$, $P_{QV}$ and $P_{IPD}$ must be available. In the proposed rate adaptation framework, the system state at time $t$ is defined as
$\emph{\textbf{S}}_{t}=(q_{t-1},\beta_{t-1}^{est},c_{t},b_{t})$,
where $q_{t-1}$ is the quality of the previously received video
segment, $\beta_{t-1}^{est}$ is the estimated bandwidth, $c_{t}$ is
the content complexity of the video segment to be downloaded, and
$b_{t}$ is the current buffer length. Based on the defined system
state, the average quality $\overline{Q}$ is replaced with the
quality of the video segment that may be downloaded at the decision
time $t$, i.e. $q_{t}$, while the quality variations $P_{QV}$ are
simplified as $|q_{t}-q_{t-1}|$. The re-buffering time $P_{RB}$ is
represented with $max\{(D_{t}^{est}-b_{t}),0\}$ where $D_{t}^{est}$
is the estimated download time of the video segment requested at
time $t$:
\begin{equation}\label{E6}
D_{t}^{est}=r_{t}\times{T}/{\beta_{t-1}^{est}},
\end{equation}
To stabilize the buffer length to a predefined value $(b_{0})$, we
add another item $(P_{B})$ in the \emph{QoE} model, which is
expressed as
\begin{equation}\label{E7}
P_{B}=g(b_{t+1}^{est}-b_{0})\times|b_{t+1}^{est}-b_{0}|,
\end{equation}
where
\begin{equation}\label{E8}
b_{t+1}^{est}=b_{t}+T-D_{t}^{est},
\end{equation}
is the estimated buffer length of time $t+1$, and $g(\cdot)$ is a
function defined as:
\begin{equation}\label{E9}
g(\cdot)=
\left\{
\begin{array}{lr}
-1\qquad\, b_{t+1}^{est}-b_{0}<0, &\\
-0.25\quad b_{t+1}^{est}-b_{0}\geq0. &
\end{array}
\right.
\end{equation}
When $b_{t+1}^{est}$ is smaller than $b_{0}$, the risk of playout
interruption (re-buffering) is large. Thus, we use a large penalty
coefficient of value $-1$ for $|b_{t+1}^{est}-b_{0}|$ to select
video segments with lower bitrates. When $b_{t+1}^{est}$ is larger
than $b_{0}$, the risk of playout interruption (re-buffering) is
small, and we use a small penalty coefficient of value $-0.25$ for
$|b_{t+1}^{est}-b_{0}|$ to select video segments with higher
bitrates. Since the start up delays are manually set as a fixed
value (e.g. 2s, 5s, etc.), $\omega_{0}$ is usually set to be zero.
Finally, the user \emph{QoE} (we also call it as the reward at time
$t$, $R_{wd}^{t}$) model used in the proposed rate adaptation
framework is formulated as
\begin{equation}\label{E10}
\begin{split}
R_{wd}^{t}=QoE=q_{t}-\omega_{1}\times|q_{t}-q_{t-1}|-\omega_{2}\\\times
max\{(D_{t}^{est}-b_{t}),0\}-\omega_{3}\times P_{B},
\end{split}
\end{equation}
where $\omega_{1}$, $\omega_{2}$, and $\omega_{3}$ are the model
parameters.

\begin{figure}
\centering
\includegraphics[width=8.76cm]{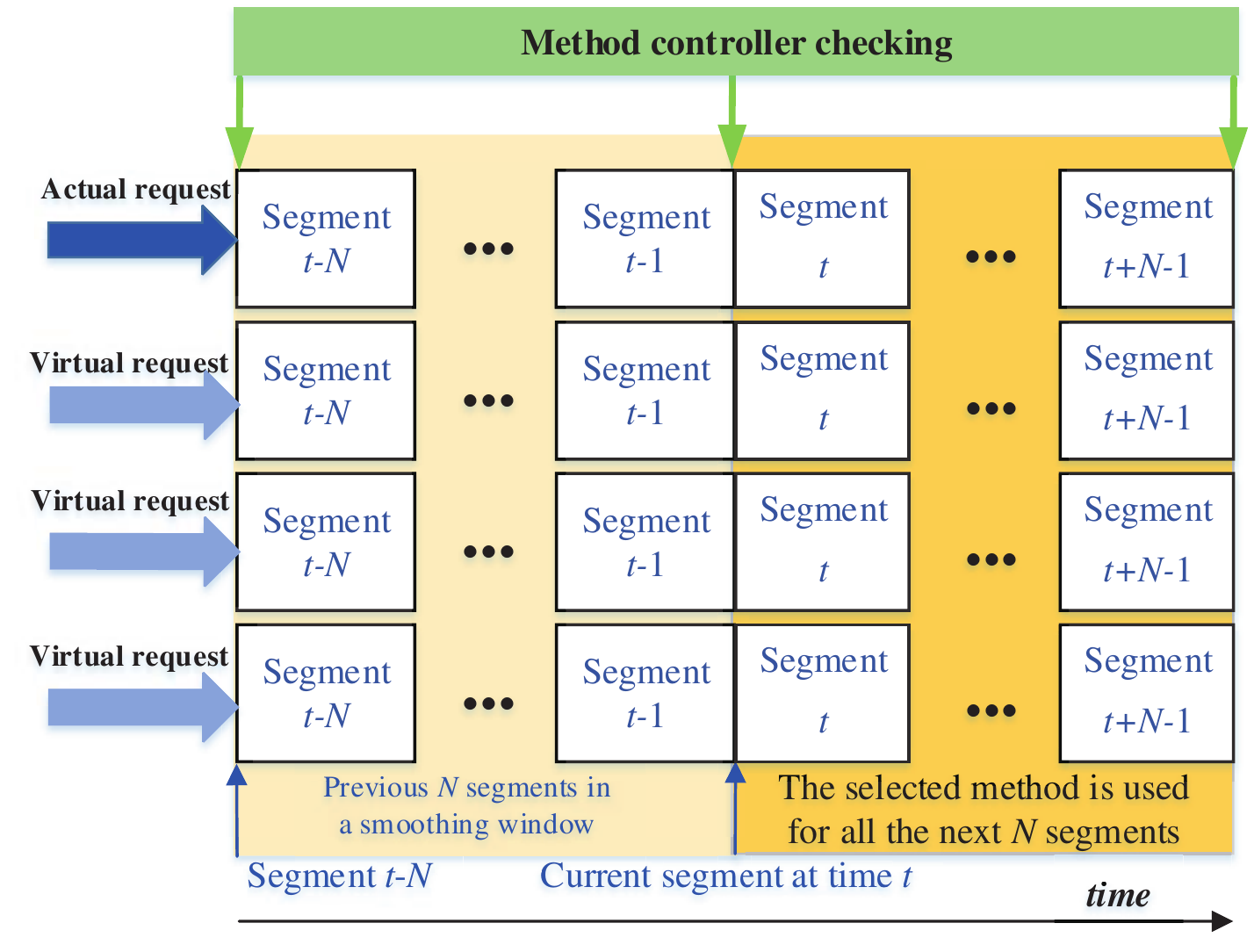}
\caption{Intermittent method switching.} \label{fig4}
\end{figure}



\begin{table}[htbp]
  \centering
  \caption{QUALITY LEVELS AND CORRESPONDING BITRATES}
\includegraphics[width=8.76cm]{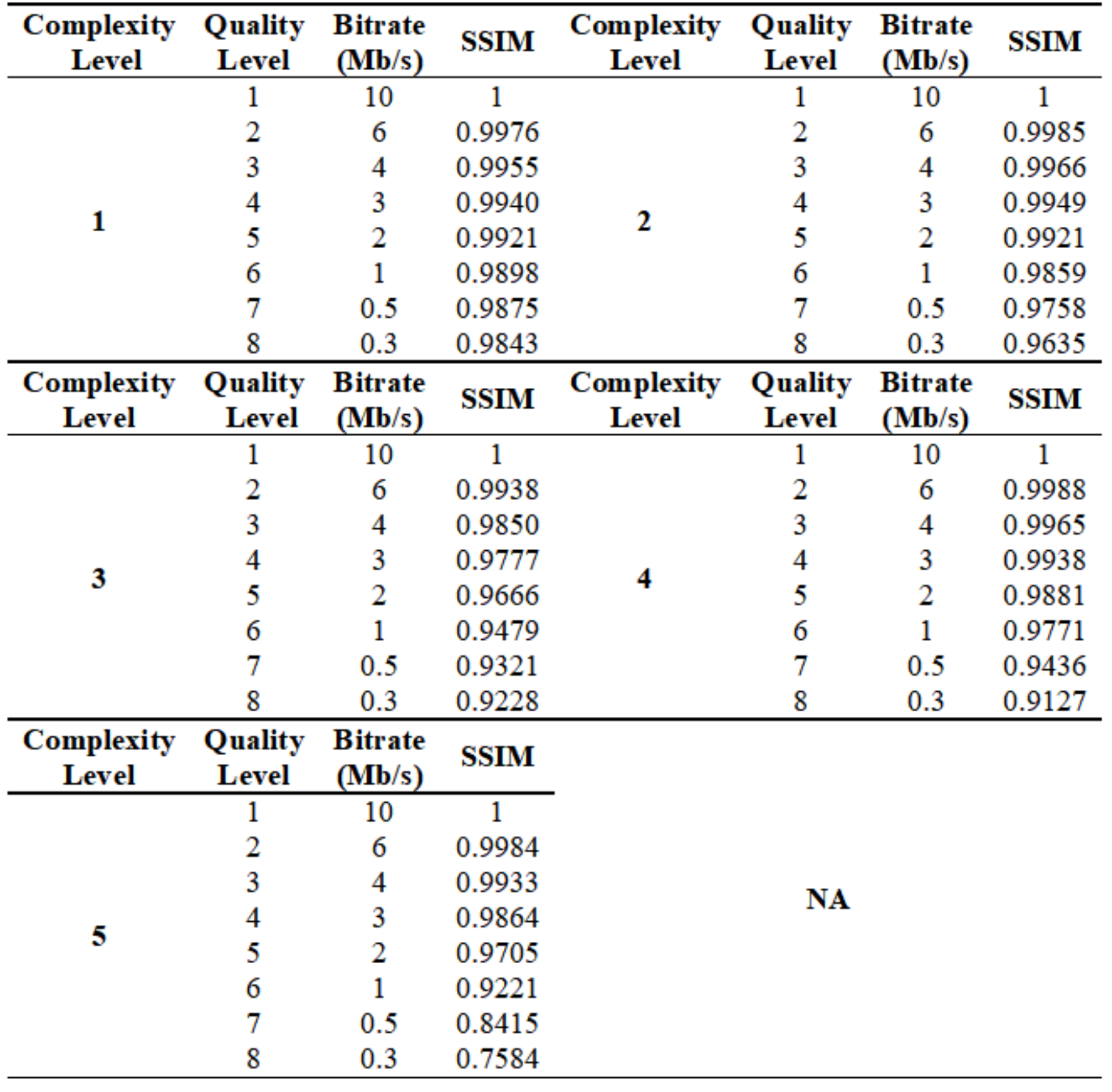}
\label{table2}
\end{table}

\section{Simulation Results and Analysis}
\subsection{Simulation Settings}

The experiments were conducted over the video sequences taken from
the \textbf{EvalVid CIF video trace reference database}. The quality
levels, complexity level, and the corresponding set of available
bitrates are described in TABLE II. Note the complexity levels are
predefined for all the video traces based on [17]. The Structural
SIMilarity index (SSIM) [50] is used as the quality metric for video
segments. The value of SSIM ranges from 0 to 1, and when the SSIM
value is smaller than 0.9, we consider the quality to be poor.

The parameter $\omega_{1}$ in Eq. (10) was set to 2 to get a good tradeoff between the average quality and the quality switches. Parameters $\omega_{2}$, $\omega_{3}$, and $b_{0}$ are used to adjust the buffer penalty in Eq. (10). We then empirically set $b_{0}=8$s, $\omega_{2}=50$, and $\omega_{3}=0.0001$ to achieve a sufficiently low probability of re-buffering events and yet allow the framework to efficiently use the buffer in order to decrease quality variations. Because of the common choice of DASH controllers, we also limit the largest buffer length to be $b_{max}=20$s. Finally, the exponential discount factor $\gamma$ was set as 0.9 to achieve a good tradeoff between the immediate and future rewards.


In simulation, four kinds of channel environments were considered:
the constant channel, the short-term fluctuating channel, the long-term
fluctuating channel, and the Markov channel (see Fig. 5). For the Markov
channel [22] [51] [52], assuming that there are $k$ states, the
following $k\times k$ transition matrix $\textbf{\emph{P}}$ is set
as
\begin{equation}\label{E11}
\emph{\textbf{P}}=\begin{pmatrix}
p_{11} & \cdots & p_{1k} \\
\vdots & \ddots & \vdots \\
p_{k1} & \cdots & p_{kk} \\
\end{pmatrix},
\end{equation}
where $p_{ij}$ denotes the transition probability from state $i$ to
state $j$:
\begin{equation}\label{E12}
p_{ij}= \left\{
\begin{array}{lr}
2p/3,\qquad |i-j|=1, &\\
p/3,\qquad\,\,\, |i-j|=2.& \end{array} \right.
\end{equation}
A large $p$ corresponds to a highly dynamic channel. By varying $p$,
we can simulate different dynamic channels. In the simulations, we
set $p$ to 0, 0.25 and 0.5 [17]. To verify the
performance of the proposed ensemble rate adaptation
framework, we have compared it with other methods using MATLAB
simulation.

\begin{figure}
\centering
\includegraphics[width=1.7in]{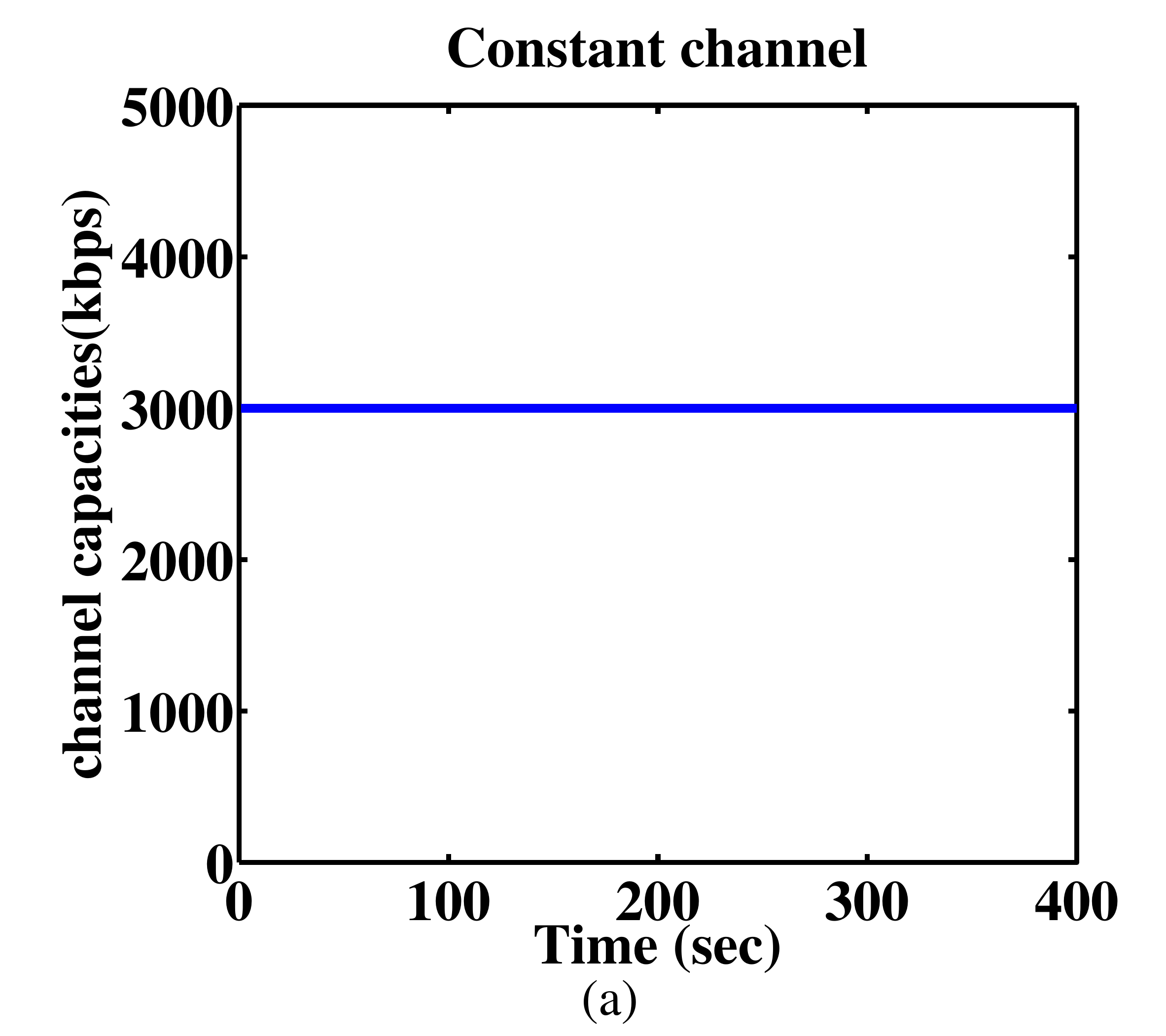}
\includegraphics[width=1.7in]{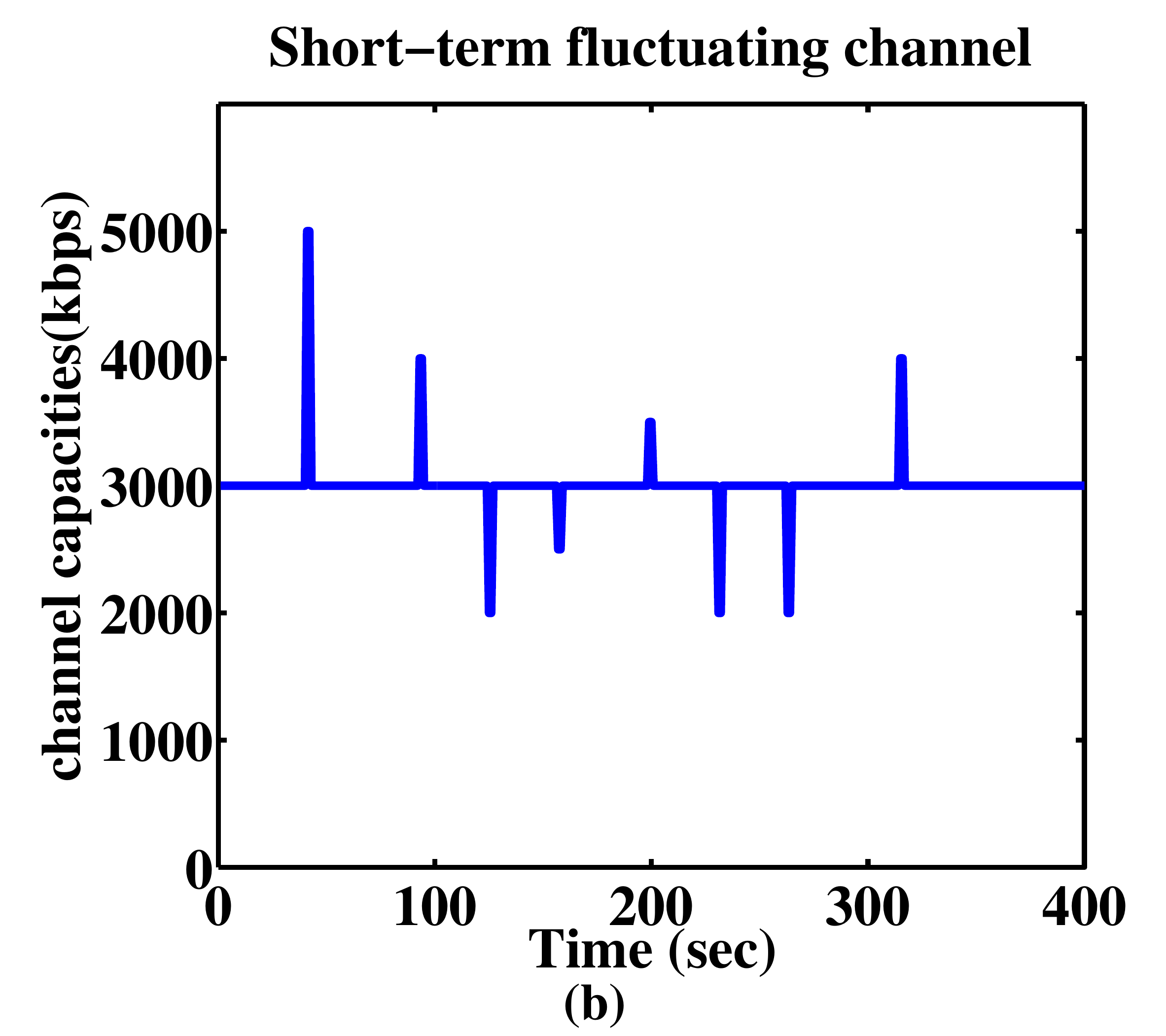}
\includegraphics[width=1.7in]{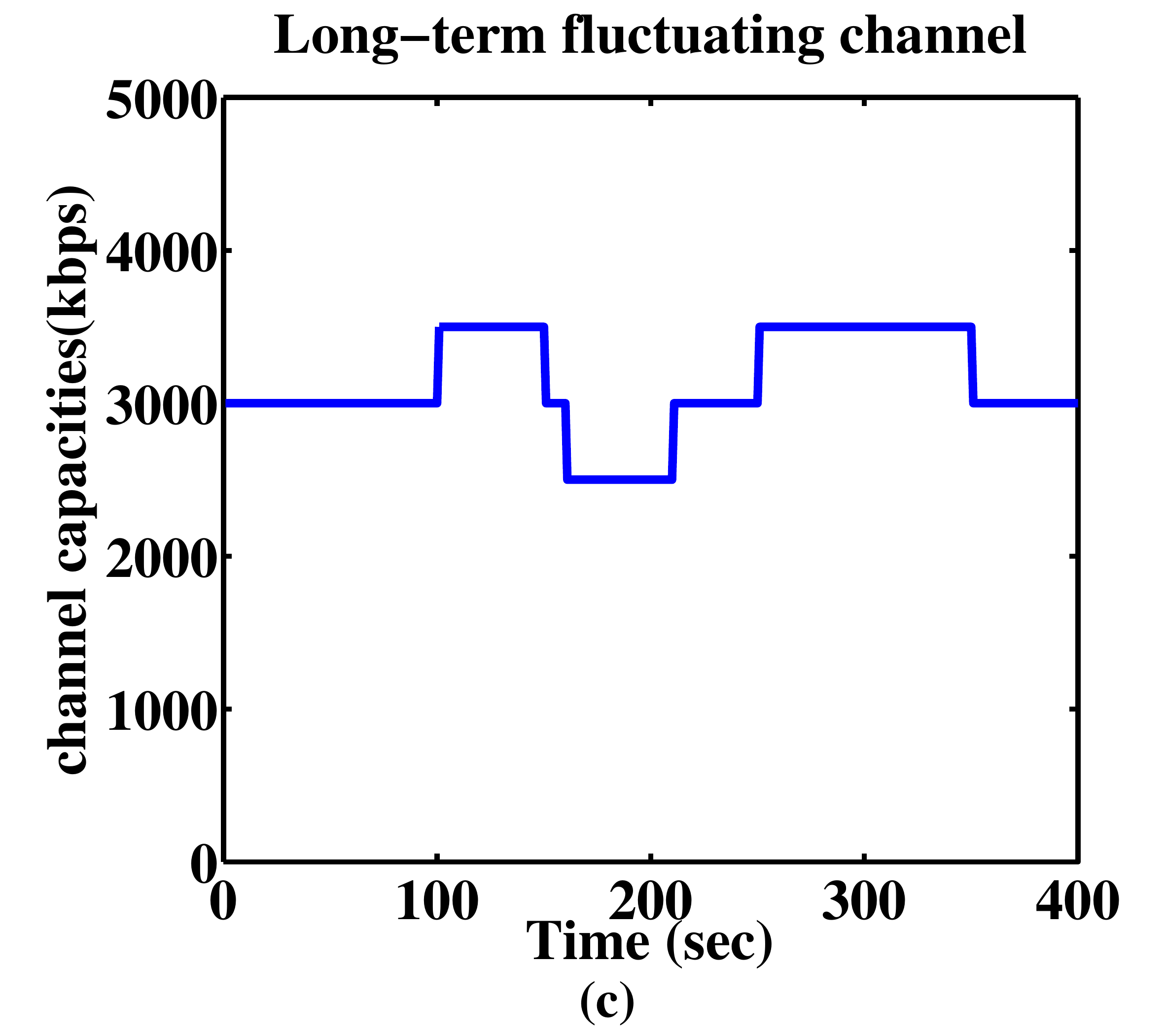}
\includegraphics[width=1.7in]{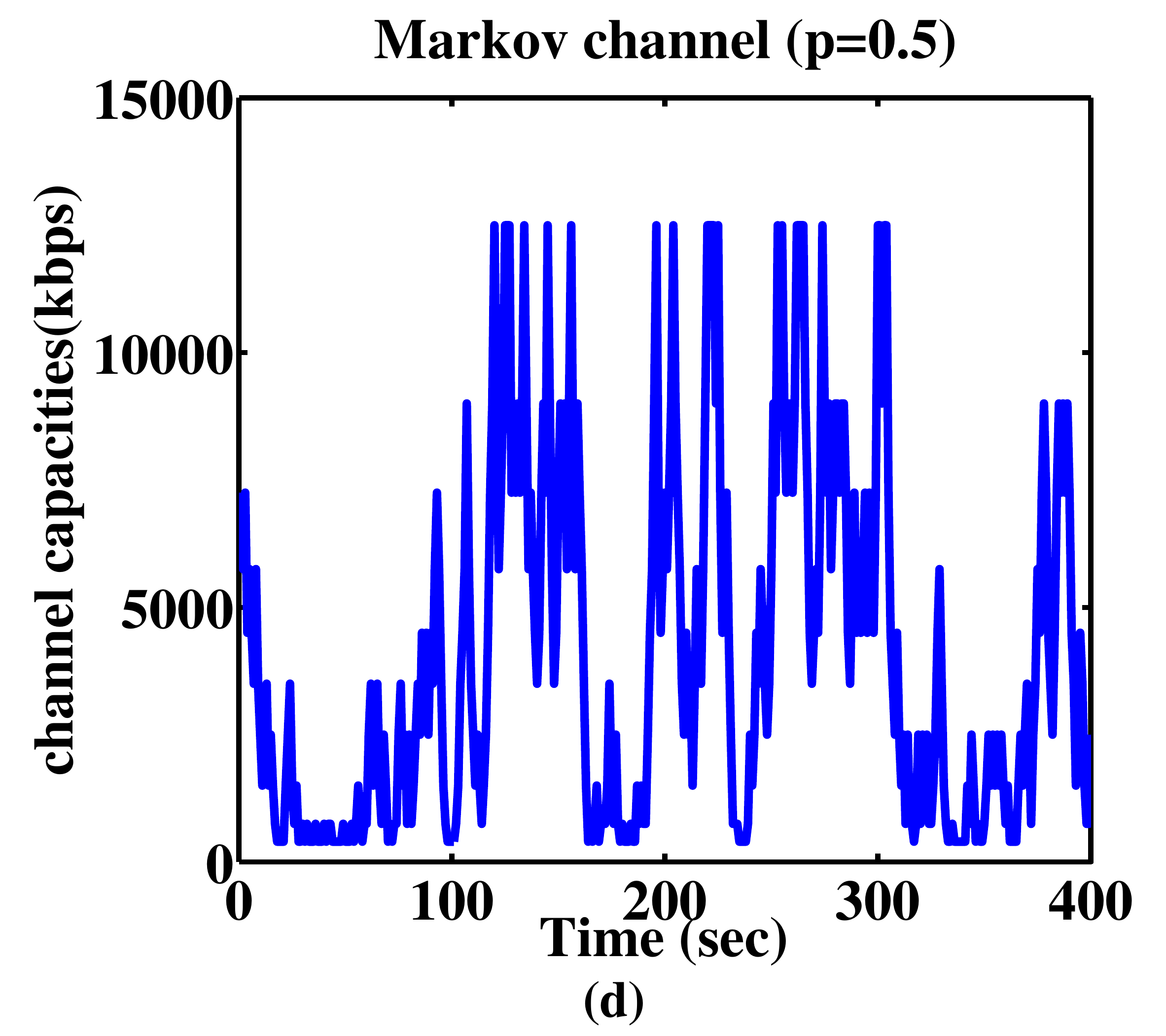}
\caption{Four kinds of channel environments.} \label{fig5}
\end{figure}

\subsection{Simulation Results}
\begin{itemize}
\item[(\emph{\romannumeral1})] \emph{Results of the proposed framework with \textbf{IAMS}}
\end{itemize}

First, we compared the methods in a static scenario in which both
the channel capacity ($\beta_{t}^{est}=3Mbps$) and video complexity
($c=4$) are constant. Then, we performed simulations on the other
three channels: the short-term fluctuating channel, long-term
fluctuating channel, and Markov channel, as shown in Figs. 5 (b),
(c), and (d), respectively. For the proposed framework (denoted by
\emph{\textbf{Proposed Framework with IAMS}}), the method controller
will check the average reward denoted as ($\overline{R_{wd}}$) of
the previous $N_{IAMS}$ (that was empirically set to be 2) decision
times (segments) of all the methods in the method pool.

\begin{figure}
\centering
\includegraphics[width=8.76cm]{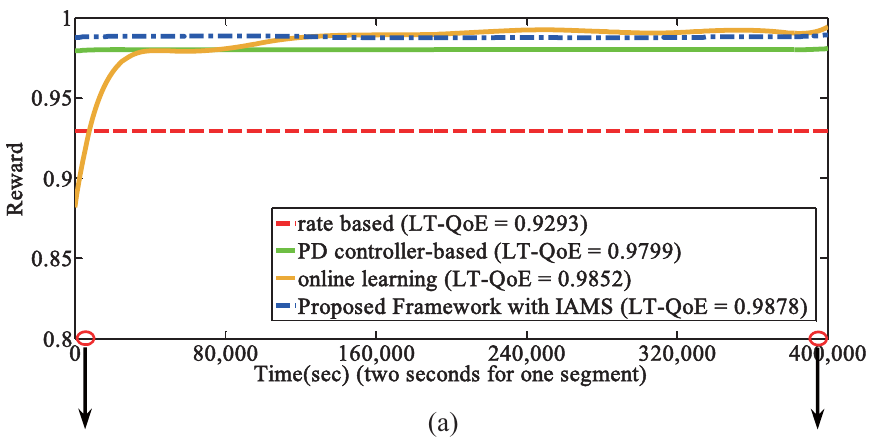}
\includegraphics[width=4.2cm]{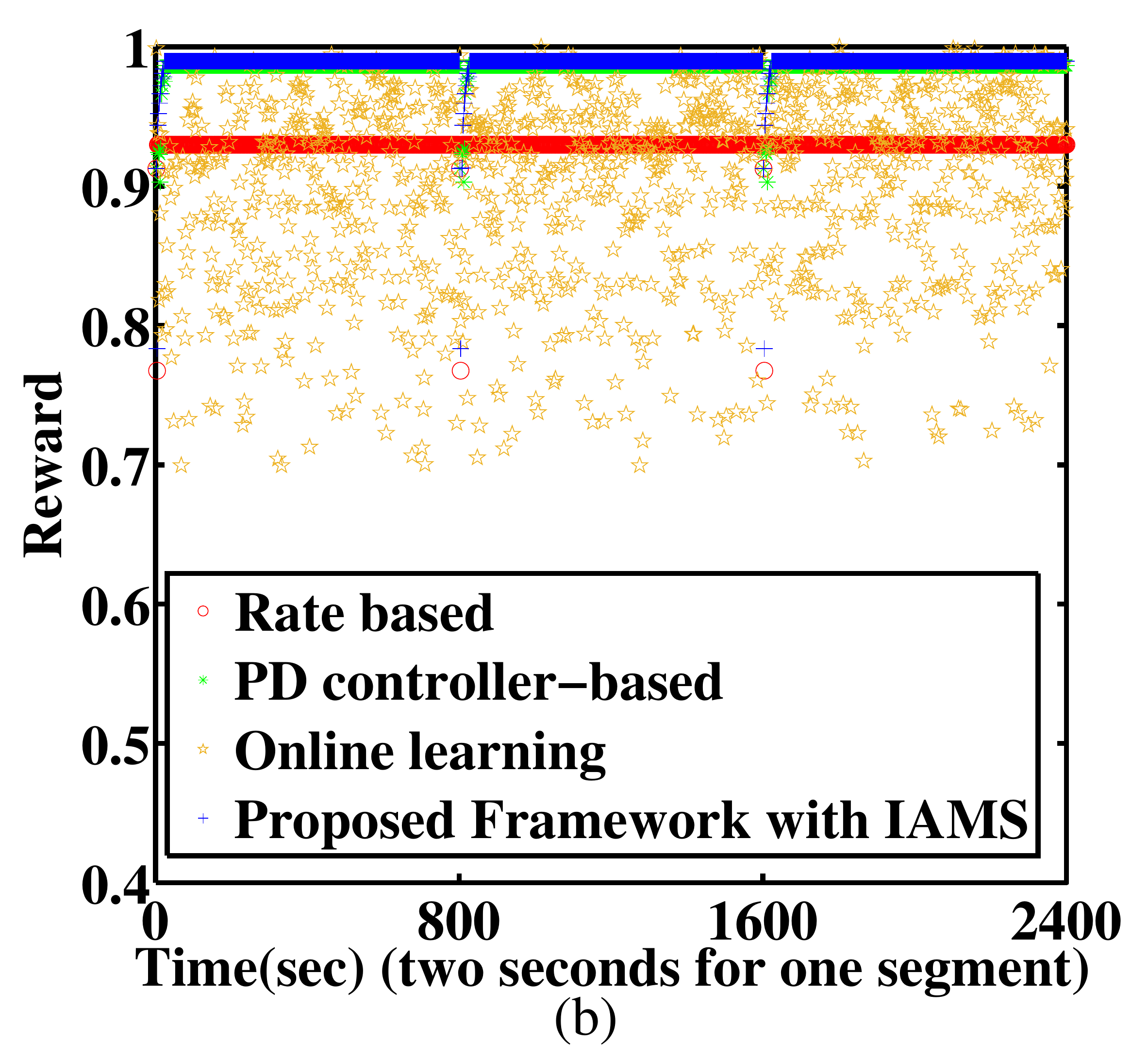}
\includegraphics[width=4.2cm]{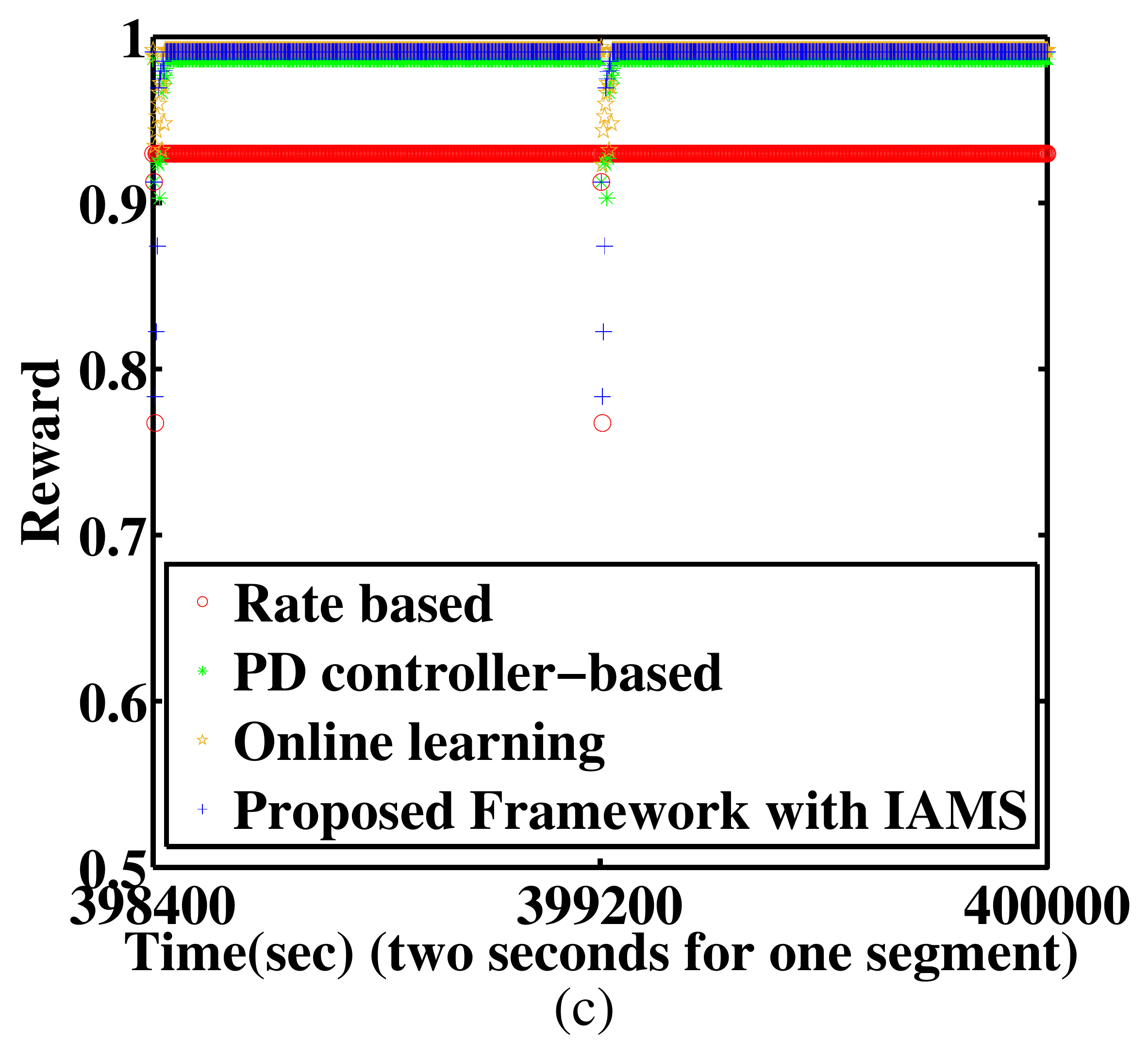}
\caption{Rewards and \emph{LT-QoE}s comparison under the constant
channel.} \label{fig6}
\end{figure}

\begin{figure}
\centering
\includegraphics[width=4.2cm]{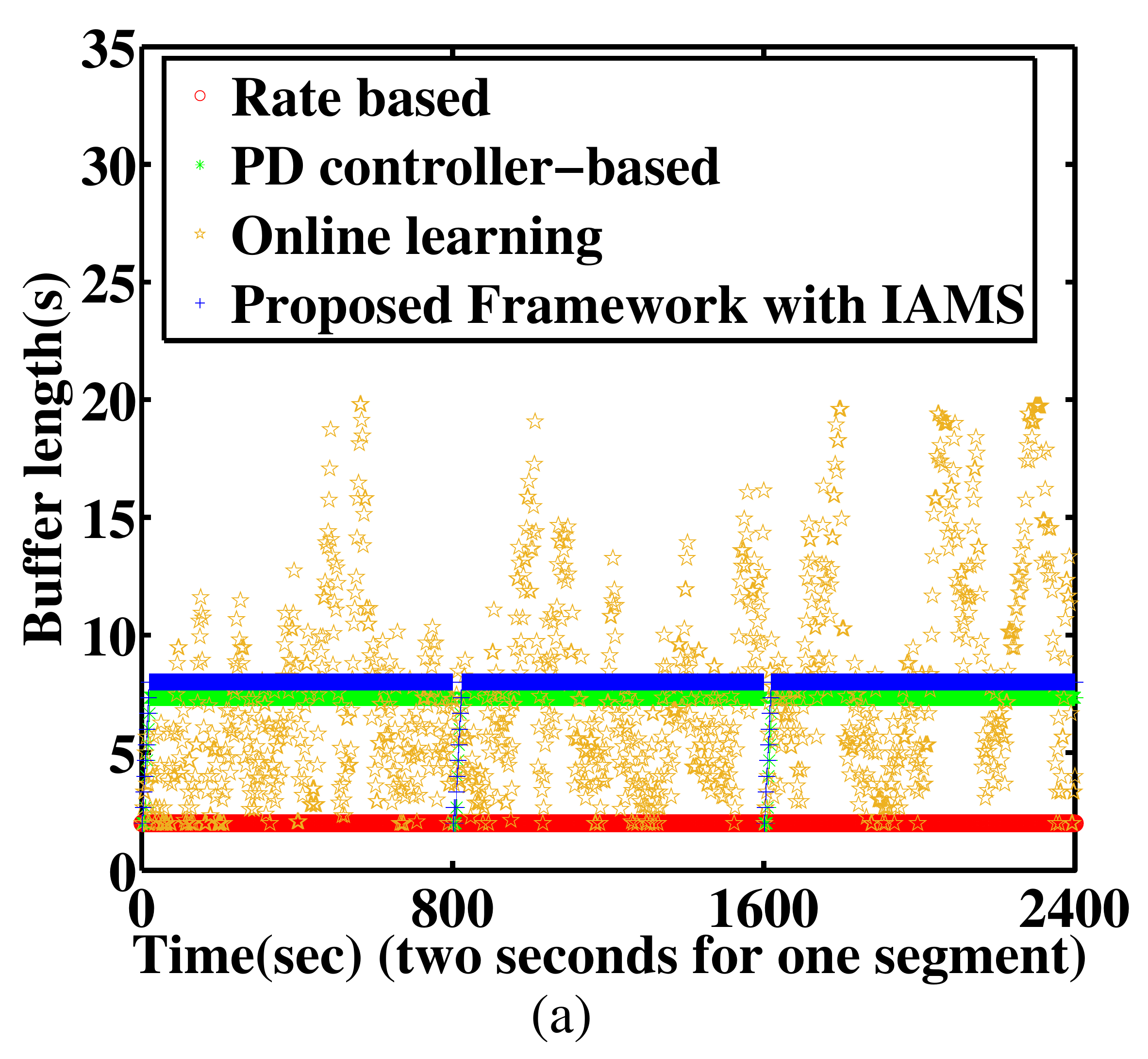}
\includegraphics[width=4.2cm]{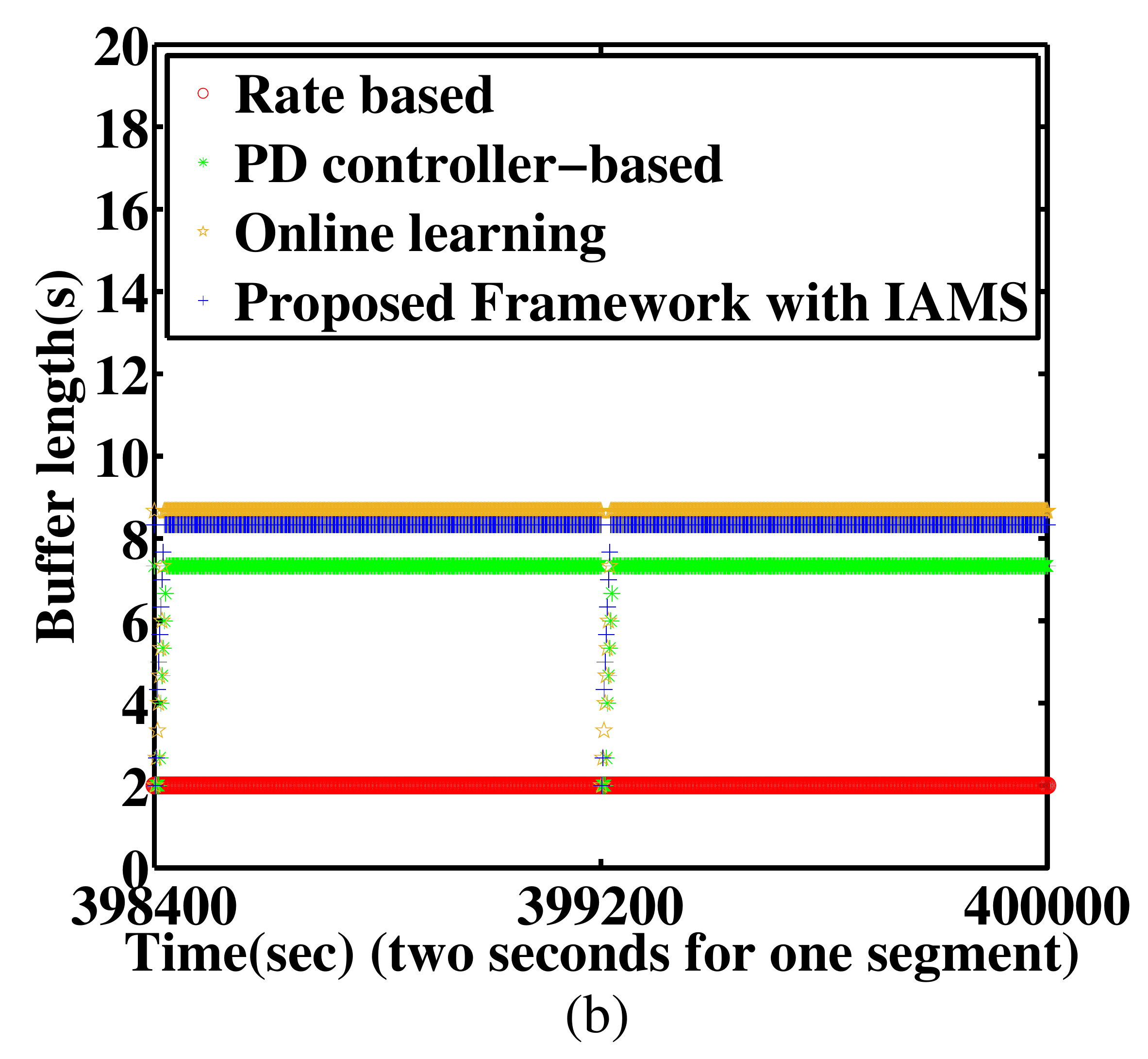}
\caption{Buffer length comparison under the constant channel.}
\label{fig7}
\end{figure}

\begin{figure}
\centering
\includegraphics[width=8.76cm]{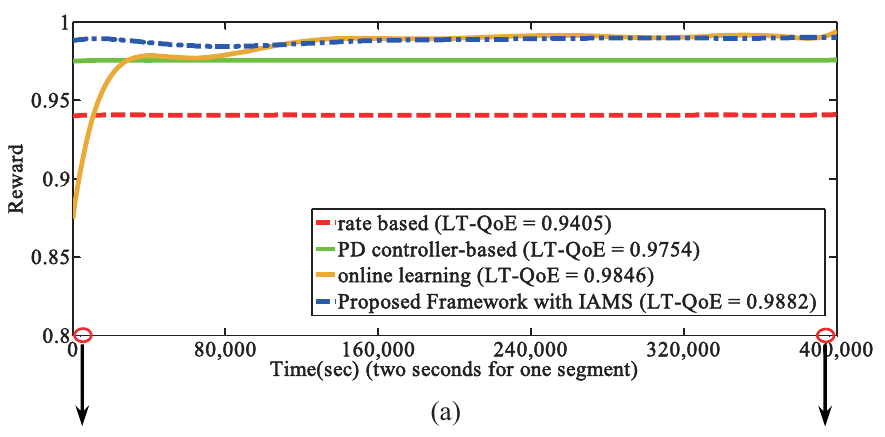}
\includegraphics[width=4.2cm]{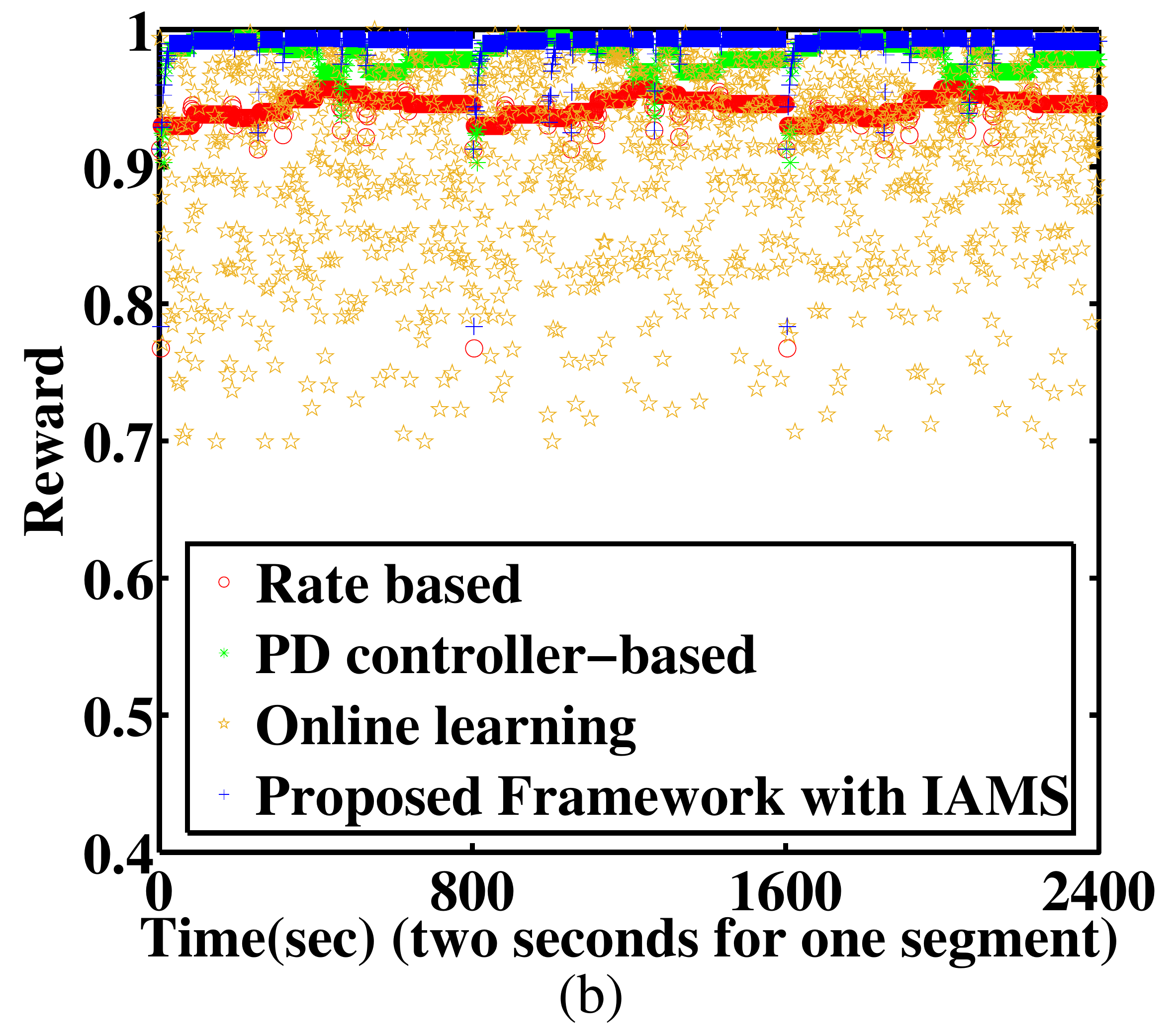}
\includegraphics[width=4.2cm]{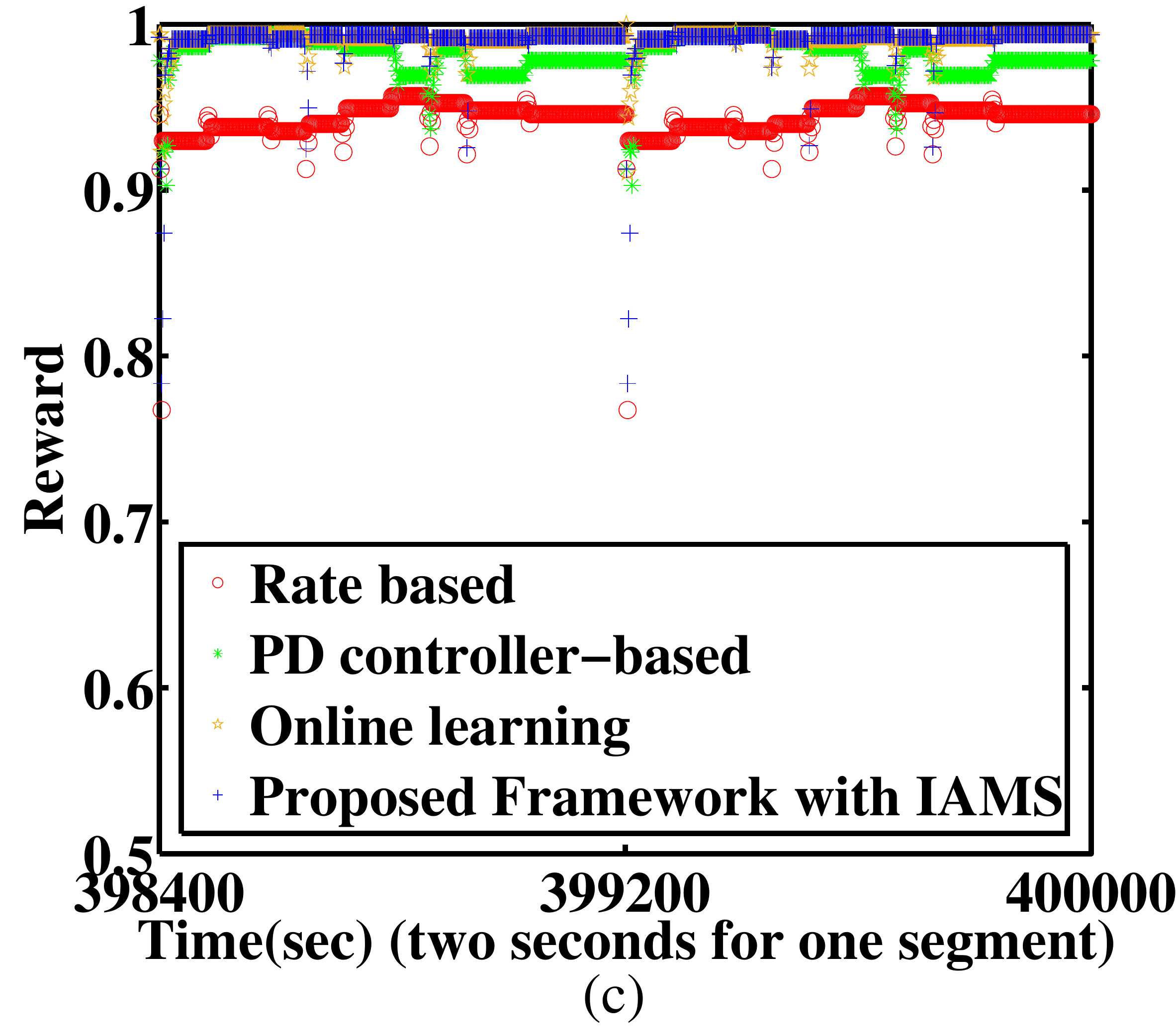}
\caption{Rewards and \emph{LT-QoE}s comparison under the short-term
fluctuating channel.} \label{fig8}
\end{figure}
\begin{figure}
\centering
\includegraphics[width=4.2cm]{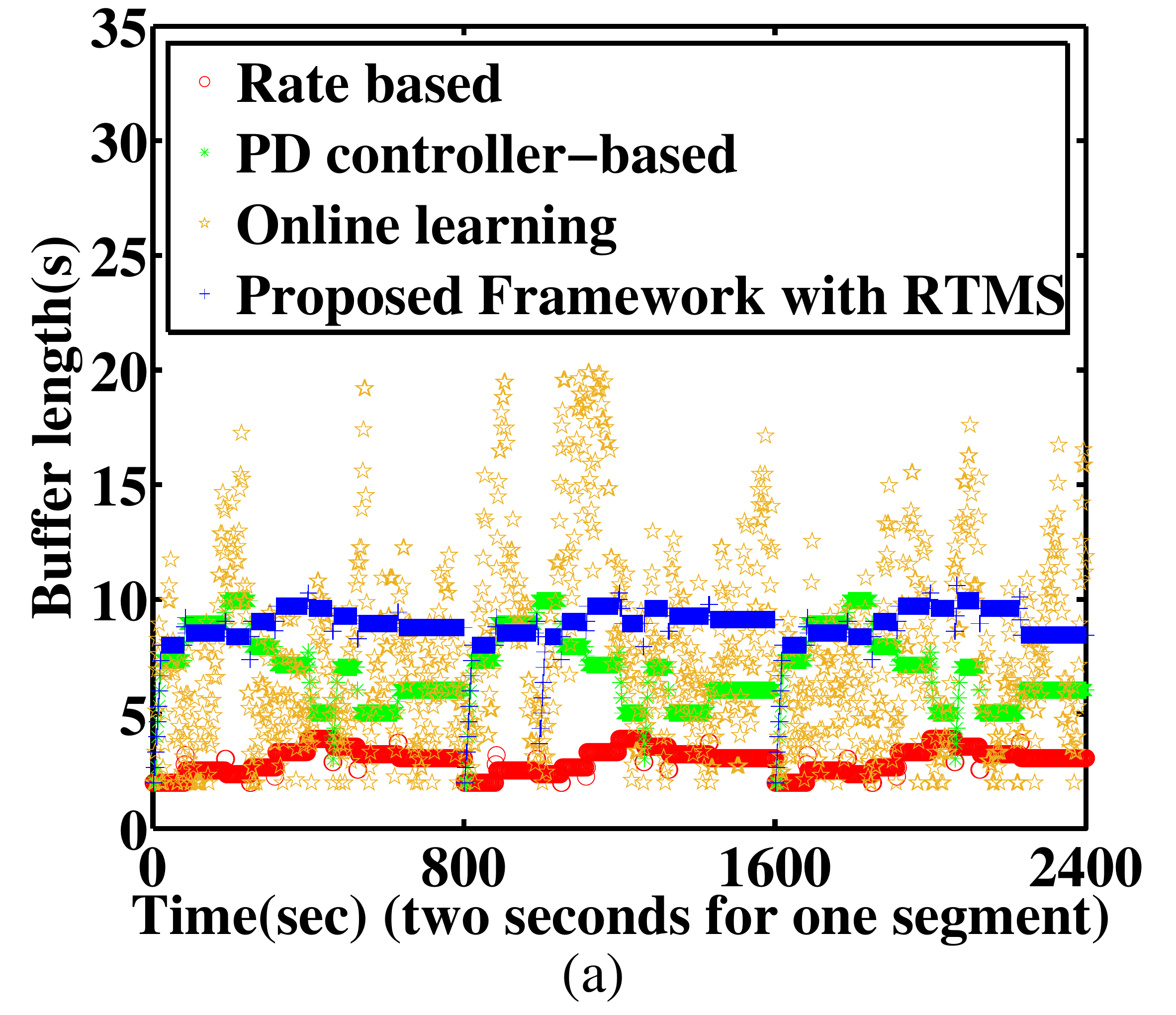}
\includegraphics[width=4.2cm]{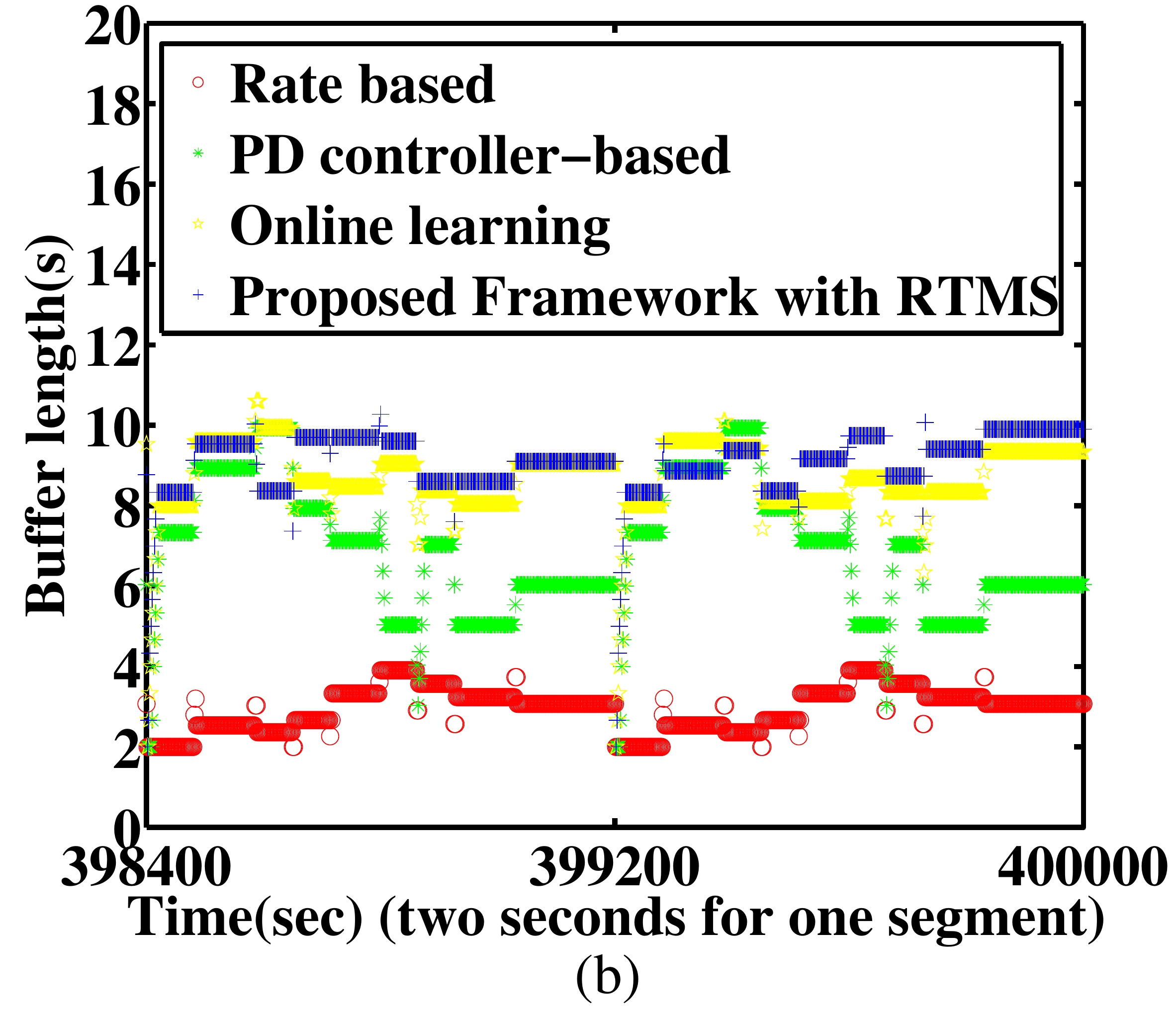}
\caption{Buffer length comparison under the short-term fluctuating
channel.} \label{fig9}
\end{figure}
\begin{figure}
\centering
\includegraphics[width=8.76cm]{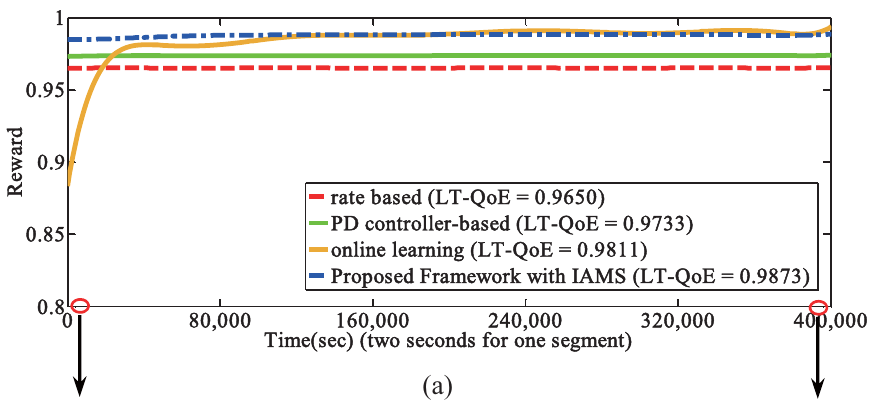}
\includegraphics[width=4.2cm]{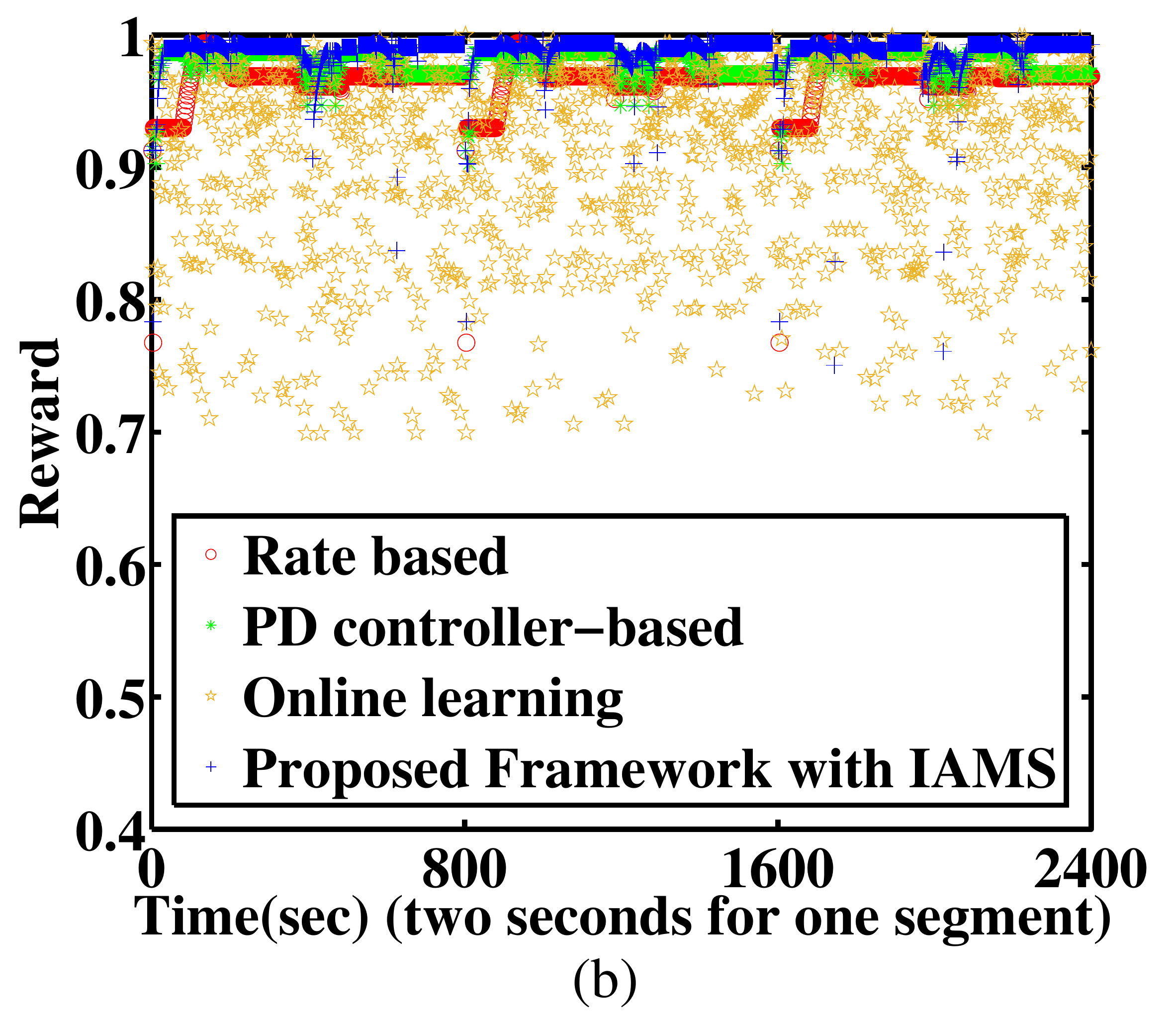}
\includegraphics[width=4.2cm]{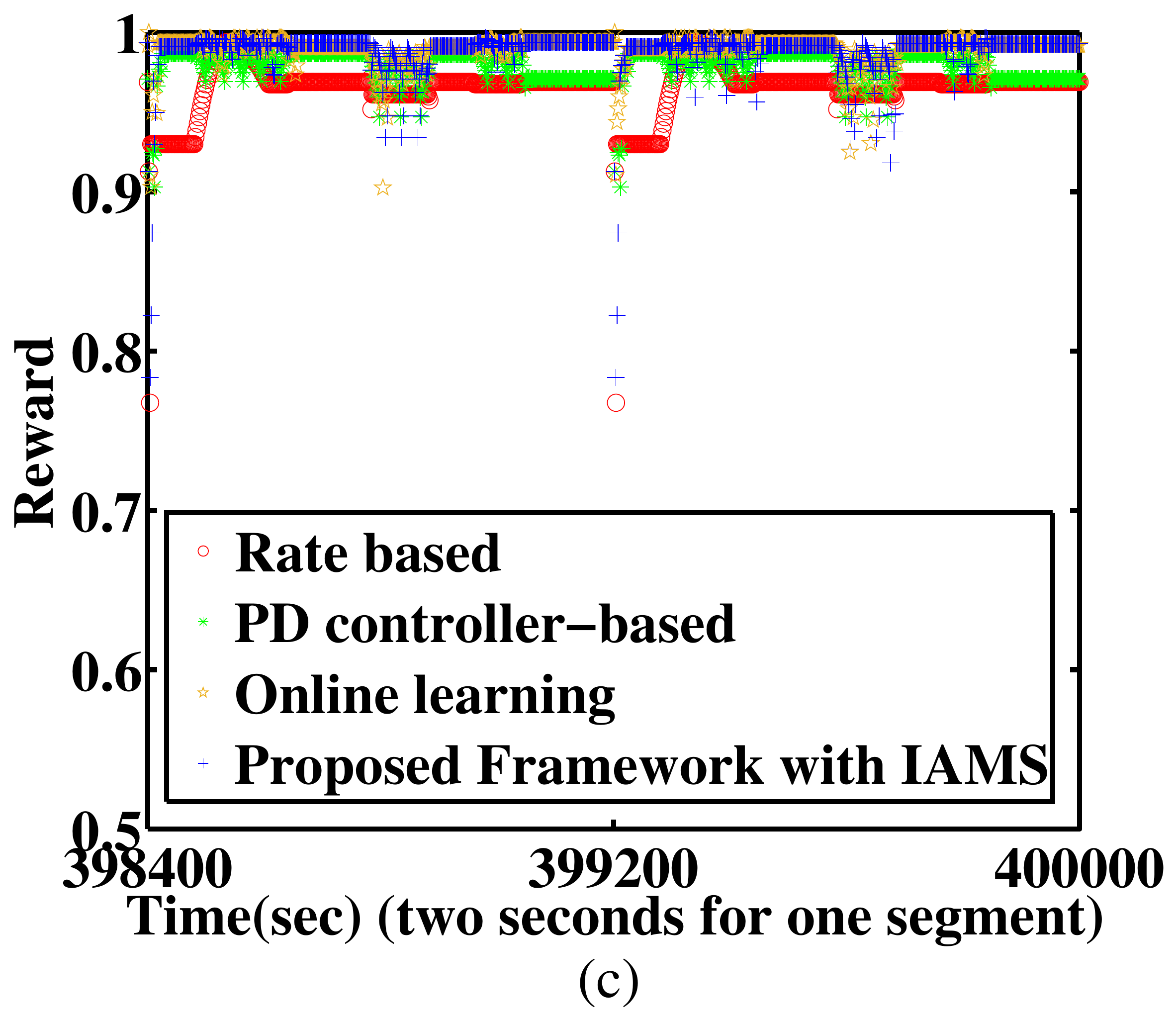}
\caption{Rewards and \emph{LT-QoE}s comparison under the long-term
fluctuating channel.} \label{fig10}
\end{figure}
\begin{figure}
\centering
\includegraphics[width=4.2cm]{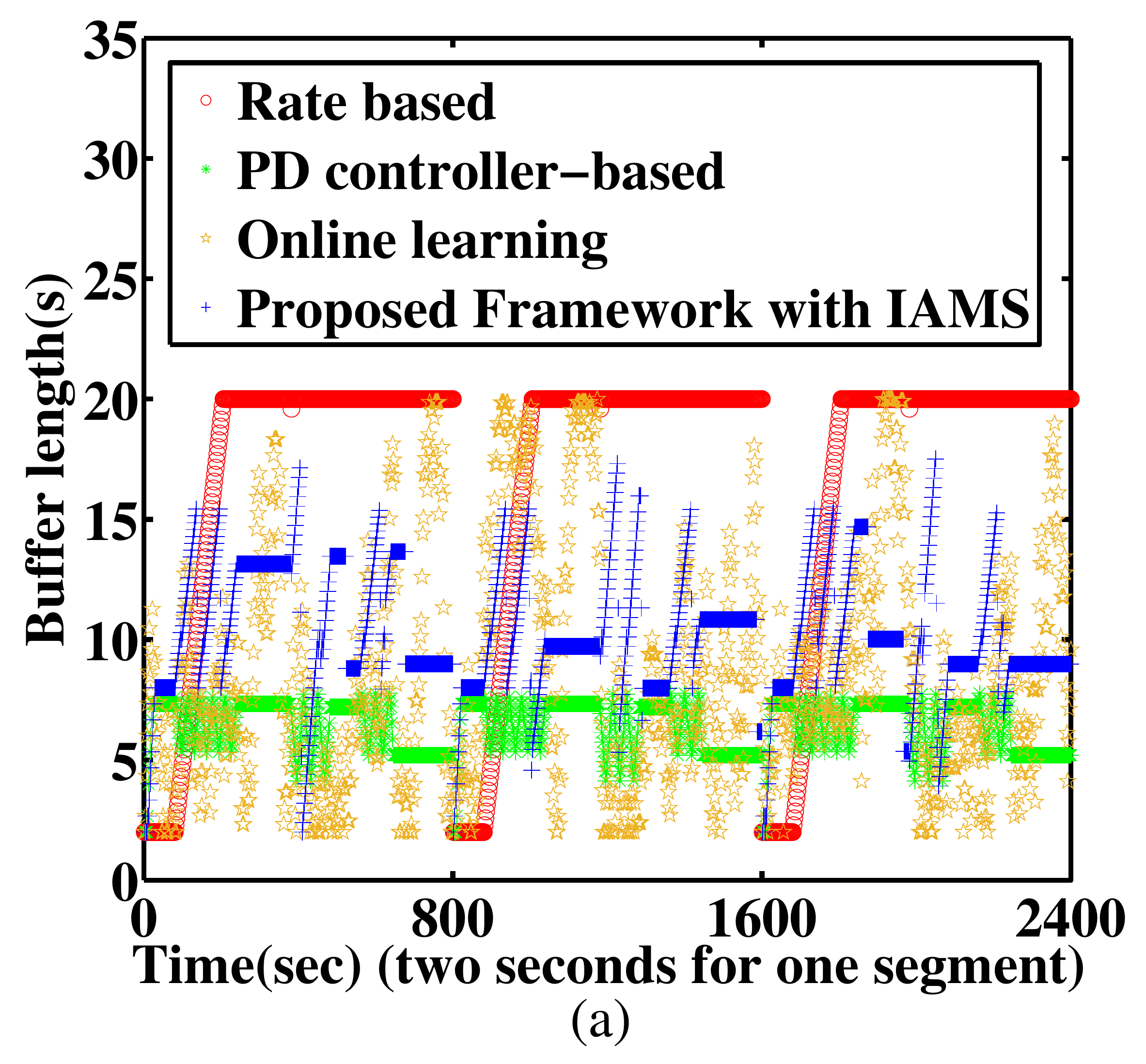}
\includegraphics[width=4.2cm]{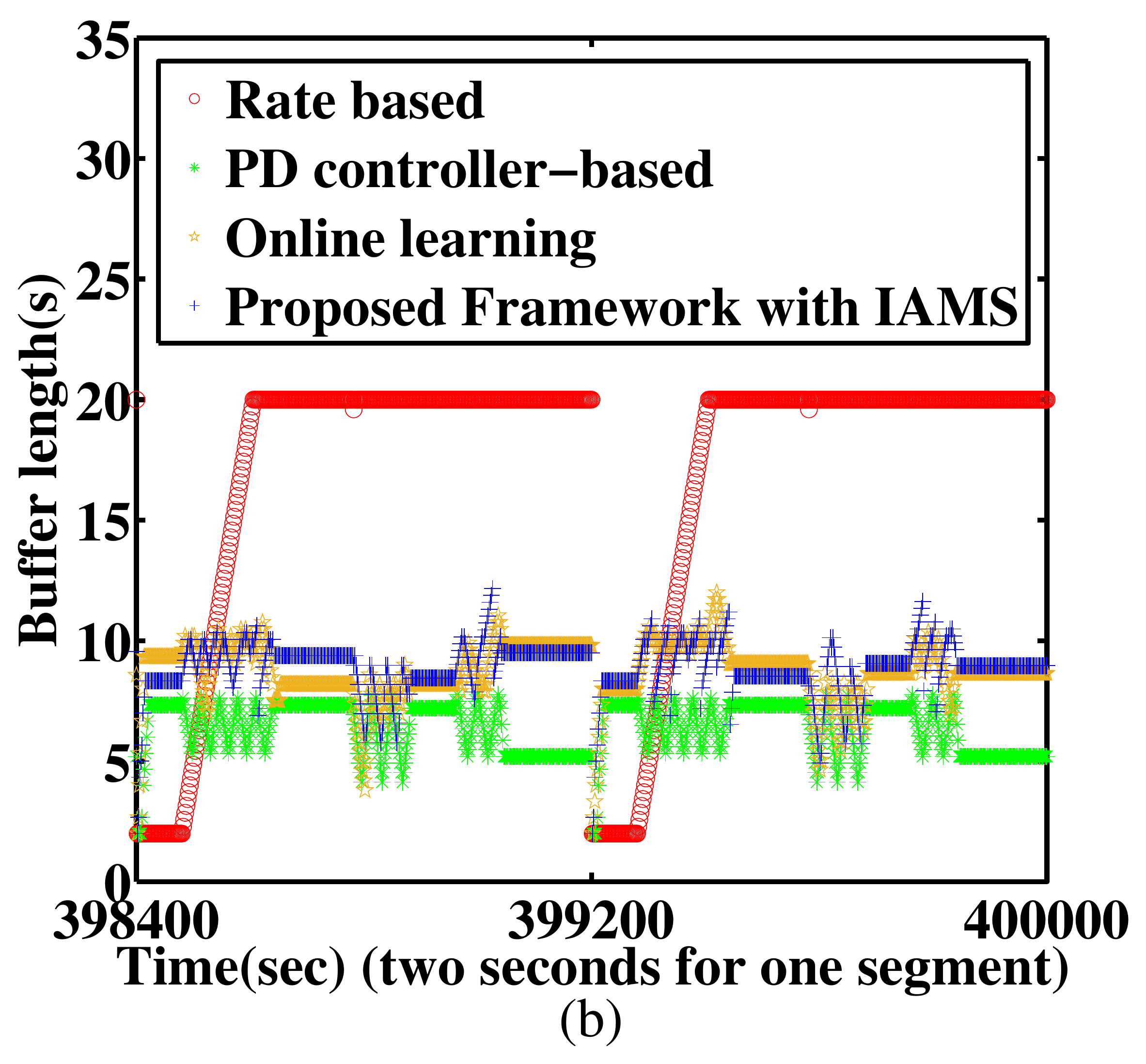}
\caption{Buffer length comparison under the long-term fluctuating
channel.} \label{fig11}
\end{figure}
\begin{figure}
\centering
\includegraphics[width=8.76cm]{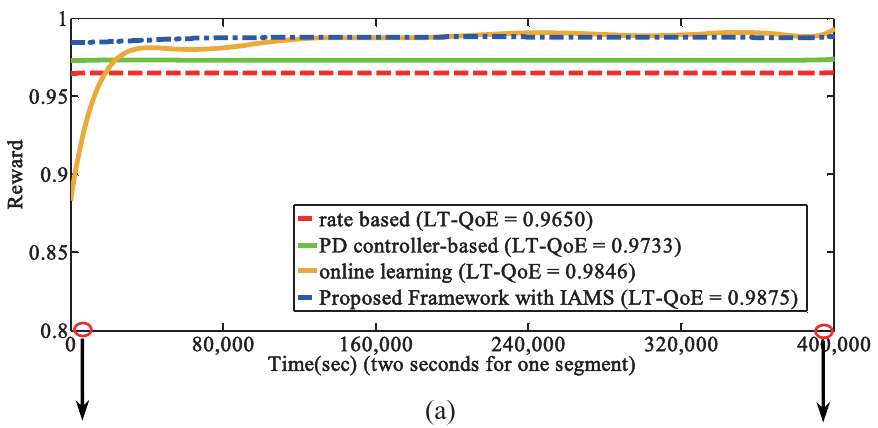}
\includegraphics[width=4.2cm]{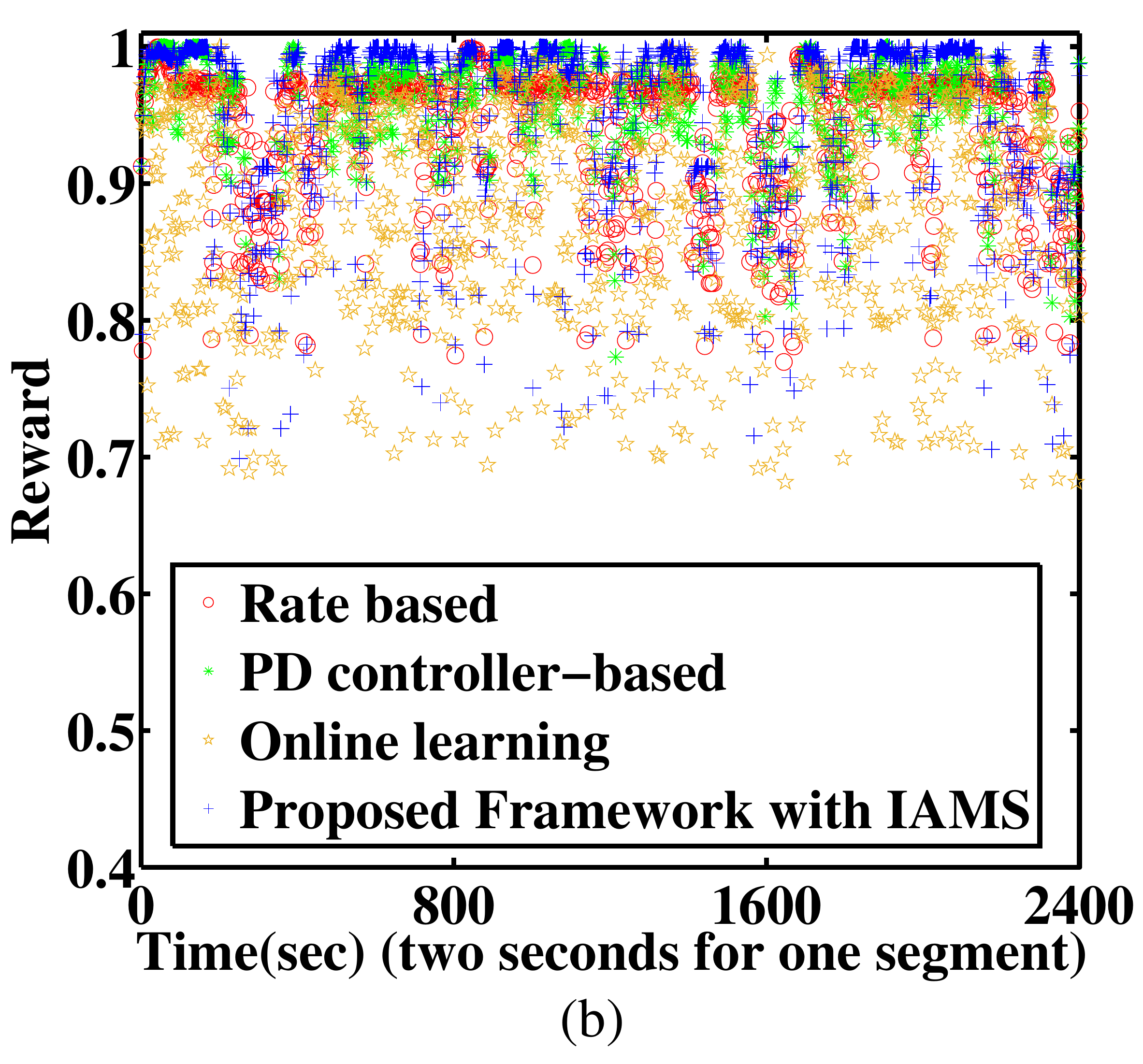}
\includegraphics[width=4.2cm]{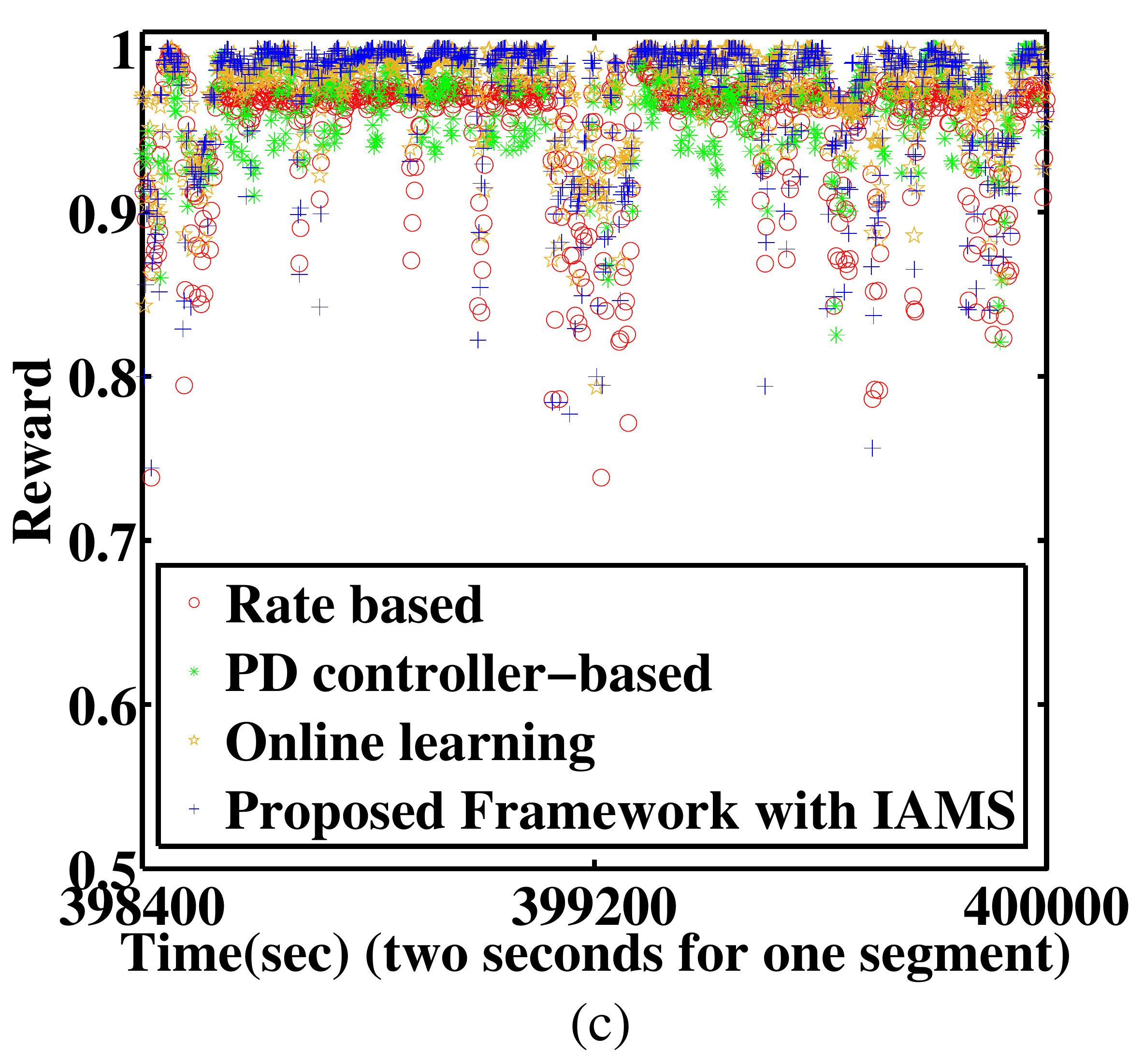}
\caption{Rewards and \emph{LT-QoE}s comparison under the Markov
channel $(p=0.5)$.} \label{fig12}
\end{figure}
\begin{figure}
\centering
\includegraphics[width=4.2cm]{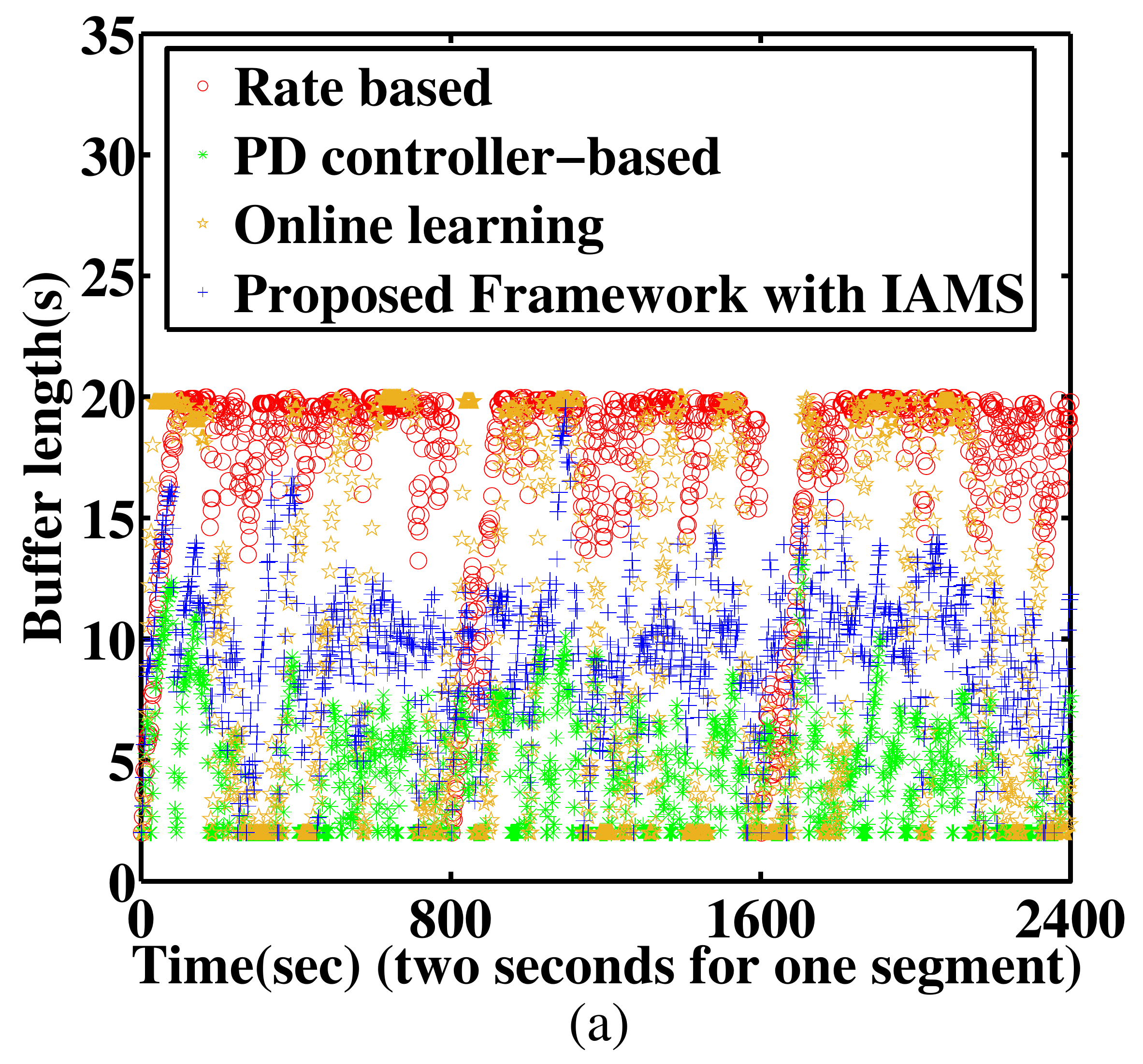}
\includegraphics[width=4.2cm]{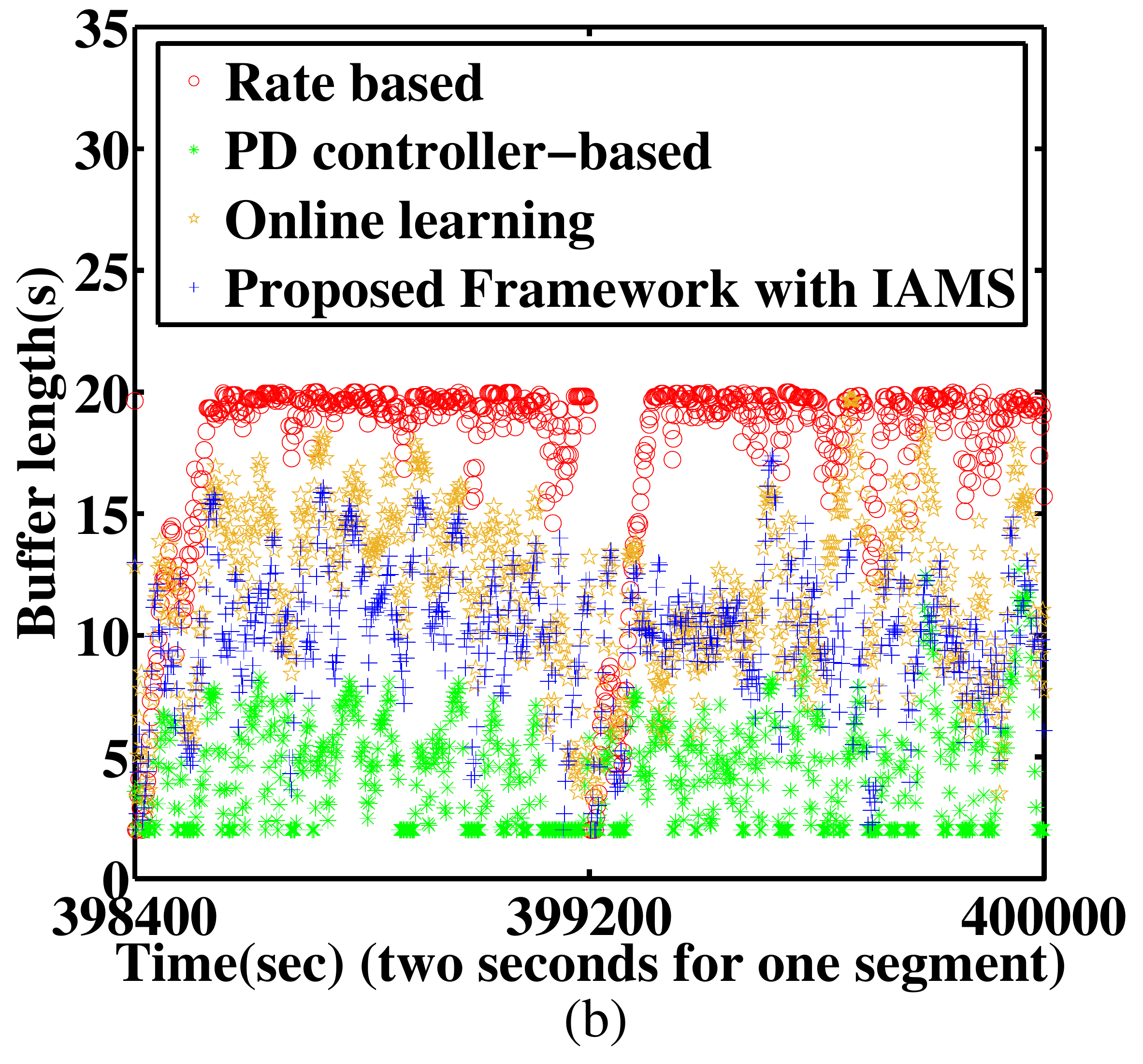}
\caption{Buffer length comparison under the Markov channel
$(p=0.5)$.} \label{fig13}
\end{figure}

For the constant channel, the reward and buffer length of each method are compared in Figs. 6 and 7, respectively. As the proposed framework will select a rate adaptation method at each decision time from the method pool, the instant reward of the proposed framework would be similar with the selected method, especially at the beginning of the request. Note that Fig. 6 (a) shows the nine order polynomial fitting curves of the rewards of the four methods to show the comparison results clearly. Therefore, there is a gap between the proposed framework and the other methods at the beginning of the request. However, from the detailed rewards shown in Fig. 6(b), we can see that the rewards of the proposed framework are similar to either the rate-based method or the PD controller based method at the beginning of the request. In the simulation, since the online learning-based method needs a very long time to converge, 500 episodes (each episode includes 400 segments) were requested to achieve a comprehensive comparison. To investigate the detailed reward variation of the four methods, Figs. 6 (b) and (c) compare the rewards of the first 2400s and the last 1600s. Since the channel bandwidth is constant ($\beta_{t}^{est}=3Mbps$), the reward of the \emph{rate-based method} is very stable, as shown in Fig. 6, whereas Fig. 7 shows that the buffer length of the \emph{rate-based method} is small, because it always chooses the bitrate that matches the capacity exactly. For the \emph{online learning-based method}, the reward is initially not convergent for the first 2400s in Fig. 6 (a). Accordingly, the buffer length also fluctuates greatly. When the \emph{online learning method} converges, the average rewards and buffer lengths become stable. For the \emph{PD controller-based method}, both the rewards and the buffer lengths are very stable, but the rewards are smaller than those of the converged \emph{online learning method}. The \textbf{\emph{Proposed Framework with IAMS}} can make use of the advantages of all the three methods. We can see that the rewards of the \textbf{\emph{Proposed Framework with IAMS}} is more stable than those of the \emph{online learning-based method}, especially at the initial requesting. This means that the proposed \textbf{\emph{Proposed Framework with IAMS}} is more suitable for today's popular short video applications than the \emph{online learning method}. From Fig. 7 (b), we can also observe that the buffer length of the \textbf{\emph{Proposed Framework with IAMS}} fluctuates slowly, and is closer to the reference buffer length ($b_{0}=8$s) than the other three methods. Besides, the average rewards (denoted as \emph{LT-QoE}s), i.e., $\frac{1}{{{t_{test}}}}\sum\nolimits_{t = {t_0}}^{{t_{test}}} {R_{wd}^t}$, of the four methods are also compared in Fig. 6. We can see that the \emph{LT-QoE} of the \textbf{\emph{Proposed Framework with IAMS}} is larger than each of the individual methods.

For the short-term fluctuating channel, long-term fluctuating channel, and the Markov channel ($p=0.5$), similar results can also be observed in Figs. 8 to 13. The \emph{LT-QoE}s of the \textbf{\emph{Proposed Framework with IAMS}} are always the largest.

\begin{figure}
\centering
\includegraphics[width=8.76cm]{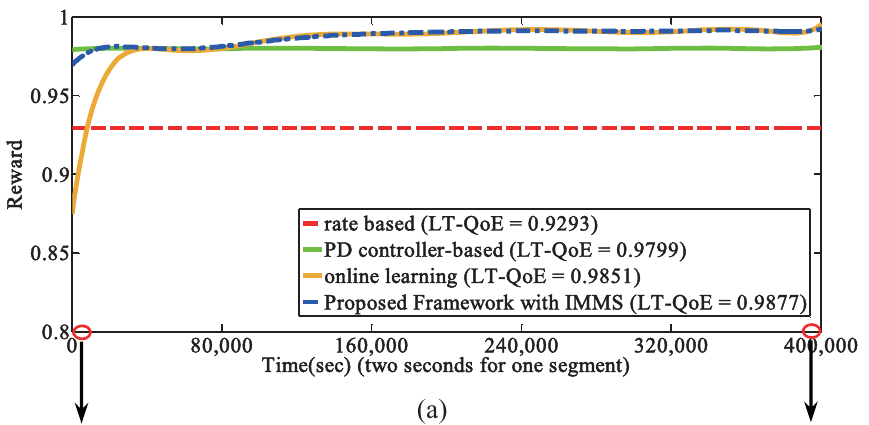}
\includegraphics[width=4.2cm]{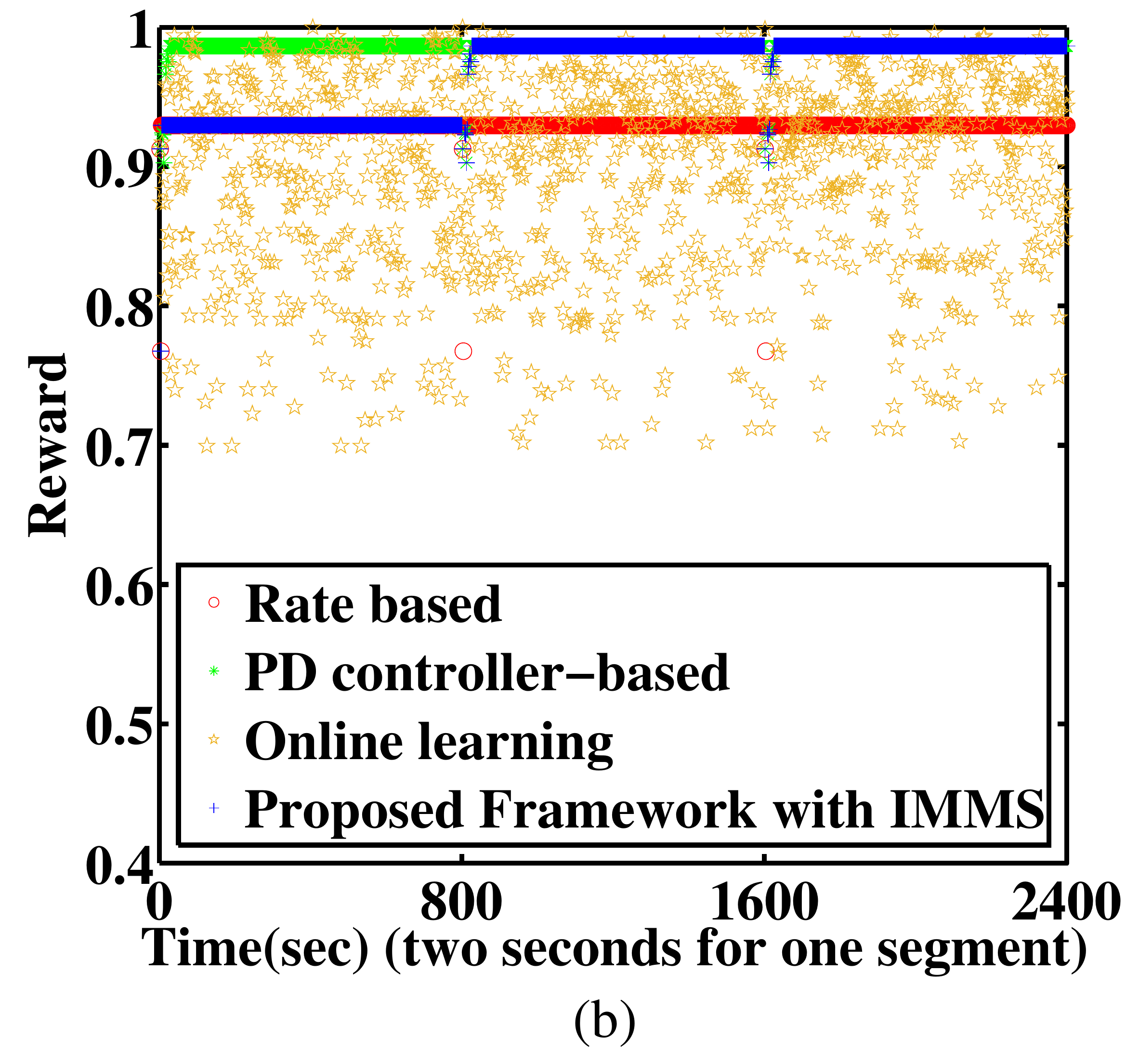}
\includegraphics[width=4.2cm]{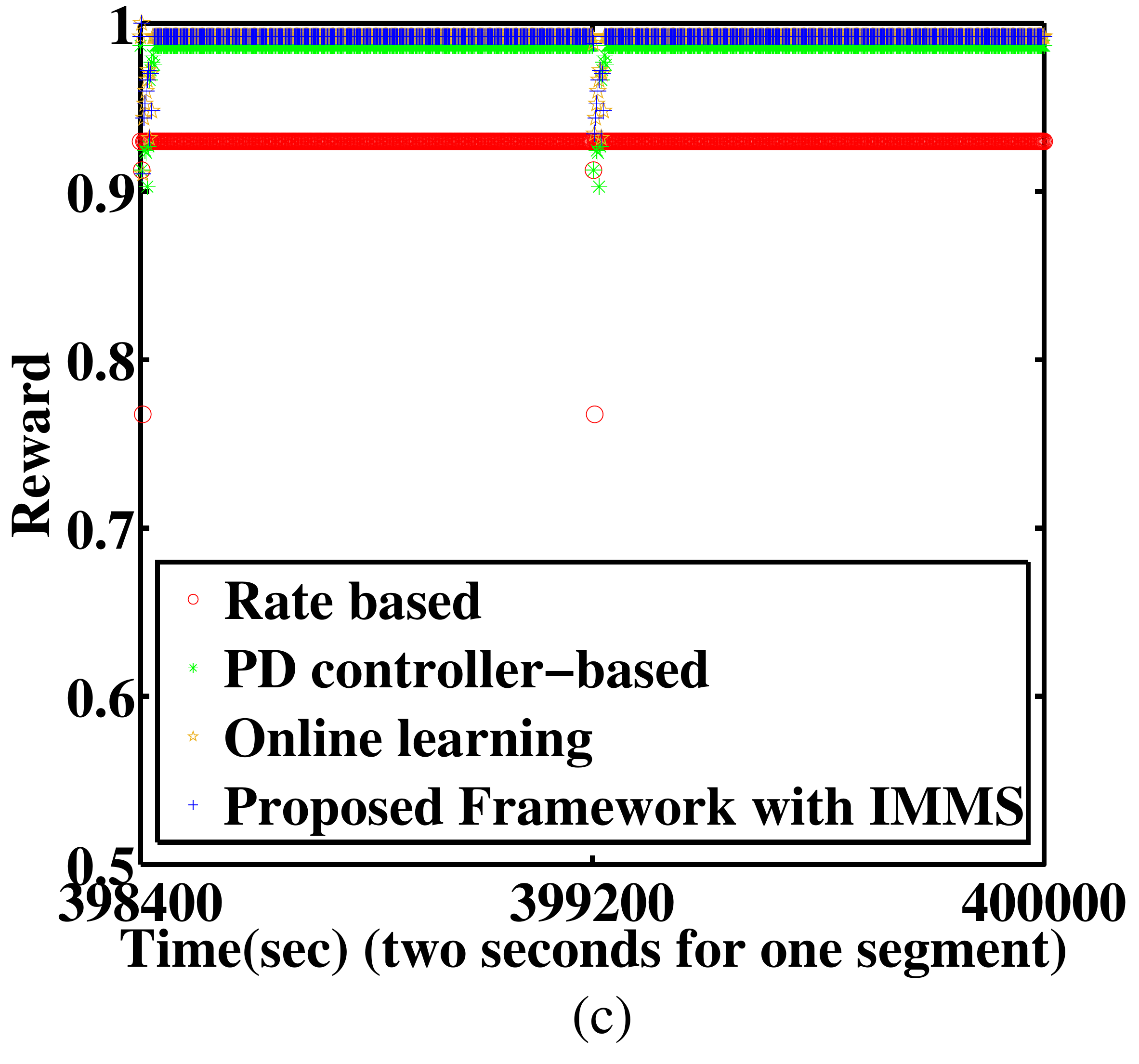}
\caption{Rewards and \emph{LT-QoE}s comparison under the constant
channel.} \label{fig14}
\end{figure}

\begin{figure}
\centering
\includegraphics[width=4.2cm]{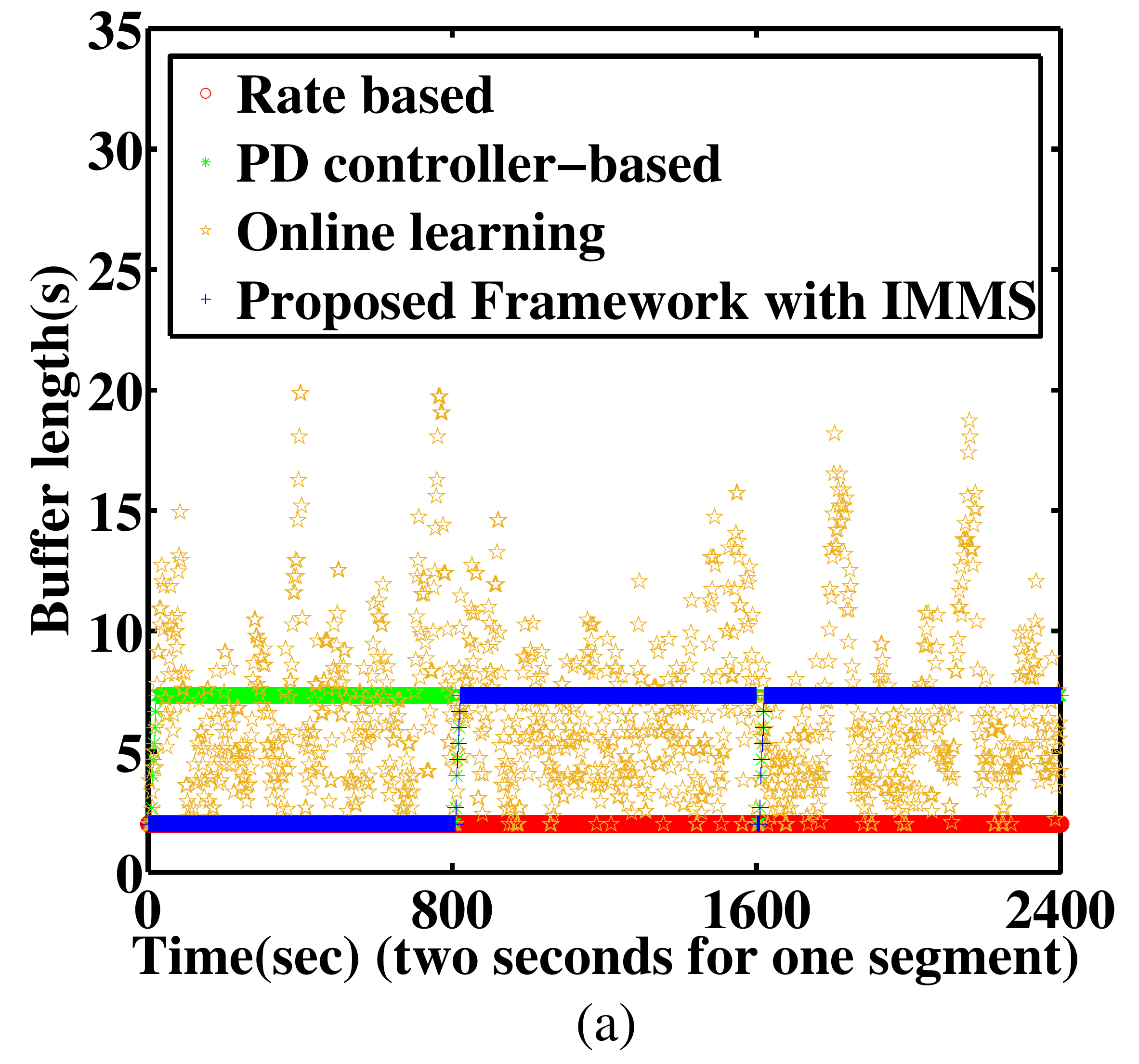}
\includegraphics[width=4.2cm]{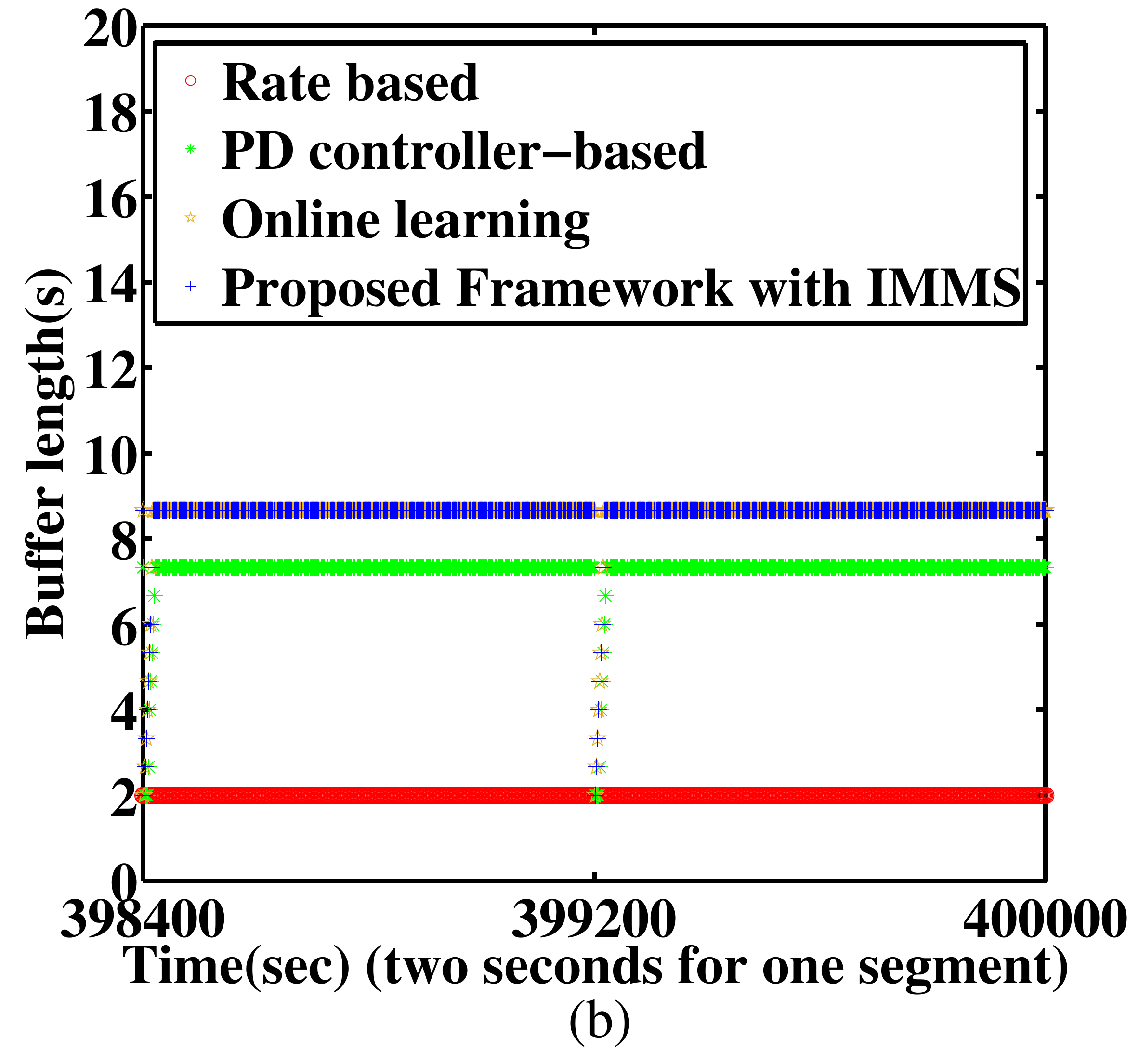}
\caption{Buffer length comparison under the constant channel.}
\label{fig15}
\end{figure}

\begin{figure}
\centering
\includegraphics[width=8.76cm]{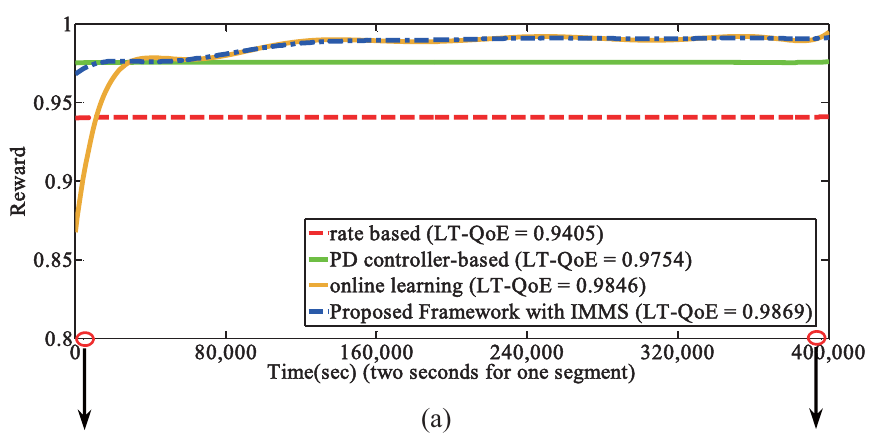}
\includegraphics[width=4.2cm]{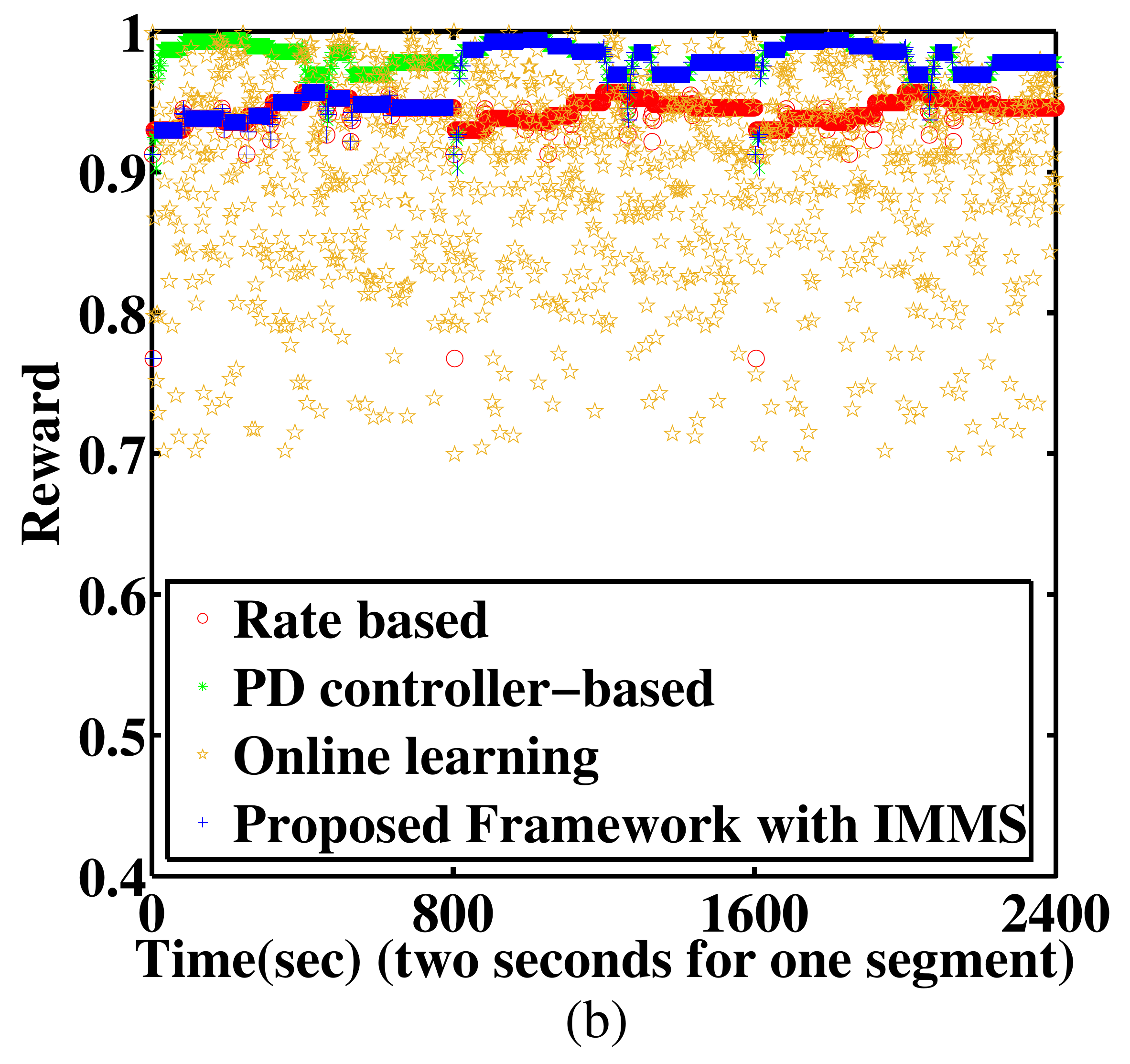}
\includegraphics[width=4.2cm]{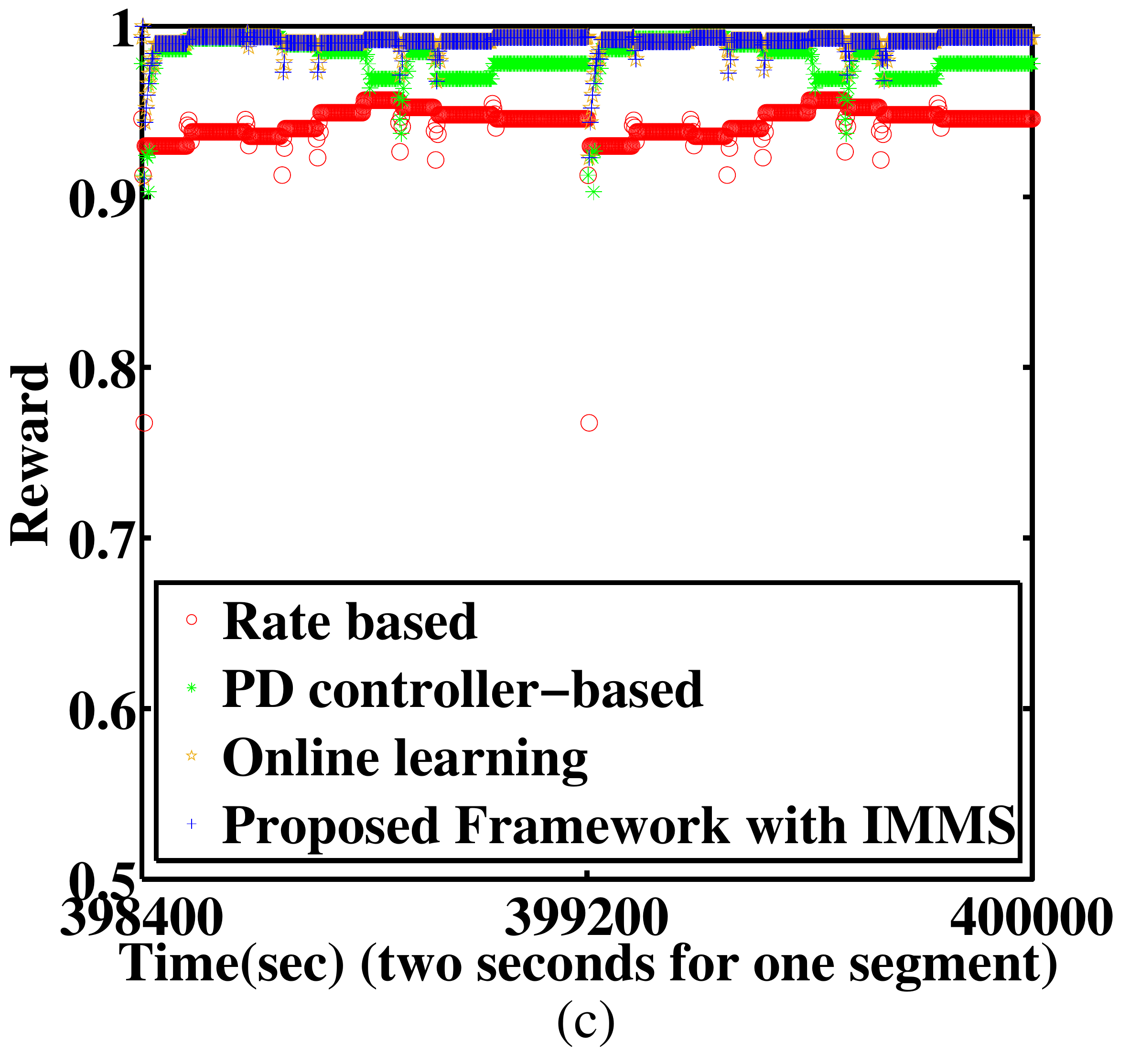}
\caption{Rewards and \emph{LT-QoE}s comparison under the short-term
fluctuating channel.} \label{fig16}
\end{figure}

\begin{itemize}
\item[(\emph{\romannumeral2})] \emph{Results of the proposed framework with \textbf{IMMS}}
\end{itemize}

Similar to part (\emph{\romannumeral1}), we also compared the
performance under the four kinds of channels (as shown in Fig. 5).
The video complexity level is also set to be 4. For the proposed
framework (denoted by \emph{\textbf{Proposed Framework with IMMS}}),
the method controller will check the product of the average reward
(denoted as ($\overline{R_{wd}}$) of each method and the ratio that
the reward of a method achieves the maximum in the previous
$N_{IMMS}$ segments ($N_{IMMS}=400$, corresponding to an episode).
The comparison results are shown in Figs. 14 to 21 for different
kinds of channels. Since the \emph{rate-based method} will be
selected at the beginning of \emph{\textbf{Proposed Framework with
IMMS}}, the rewards and the buffer lengths of the
\emph{\textbf{Proposed Framework with IMMS}} are the same with those
of the rate-based method for the first episode. From those figures,
we can observe that the \emph{LT-QoE}s of the \emph{\textbf{Proposed
Framework with IMMS}} are still the largest.

\begin{figure}
\centering
\includegraphics[width=4.2cm]{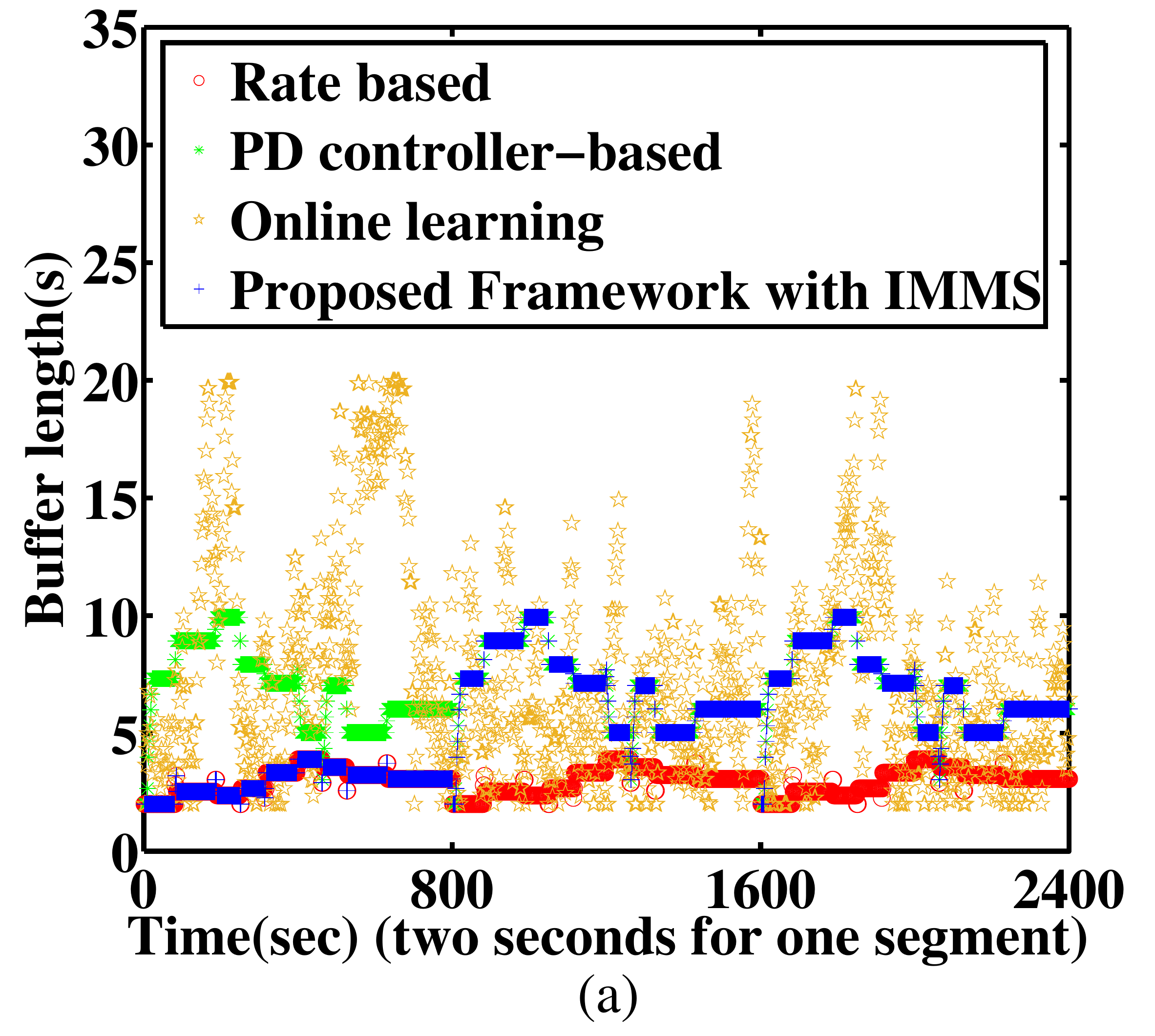}
\includegraphics[width=4.2cm]{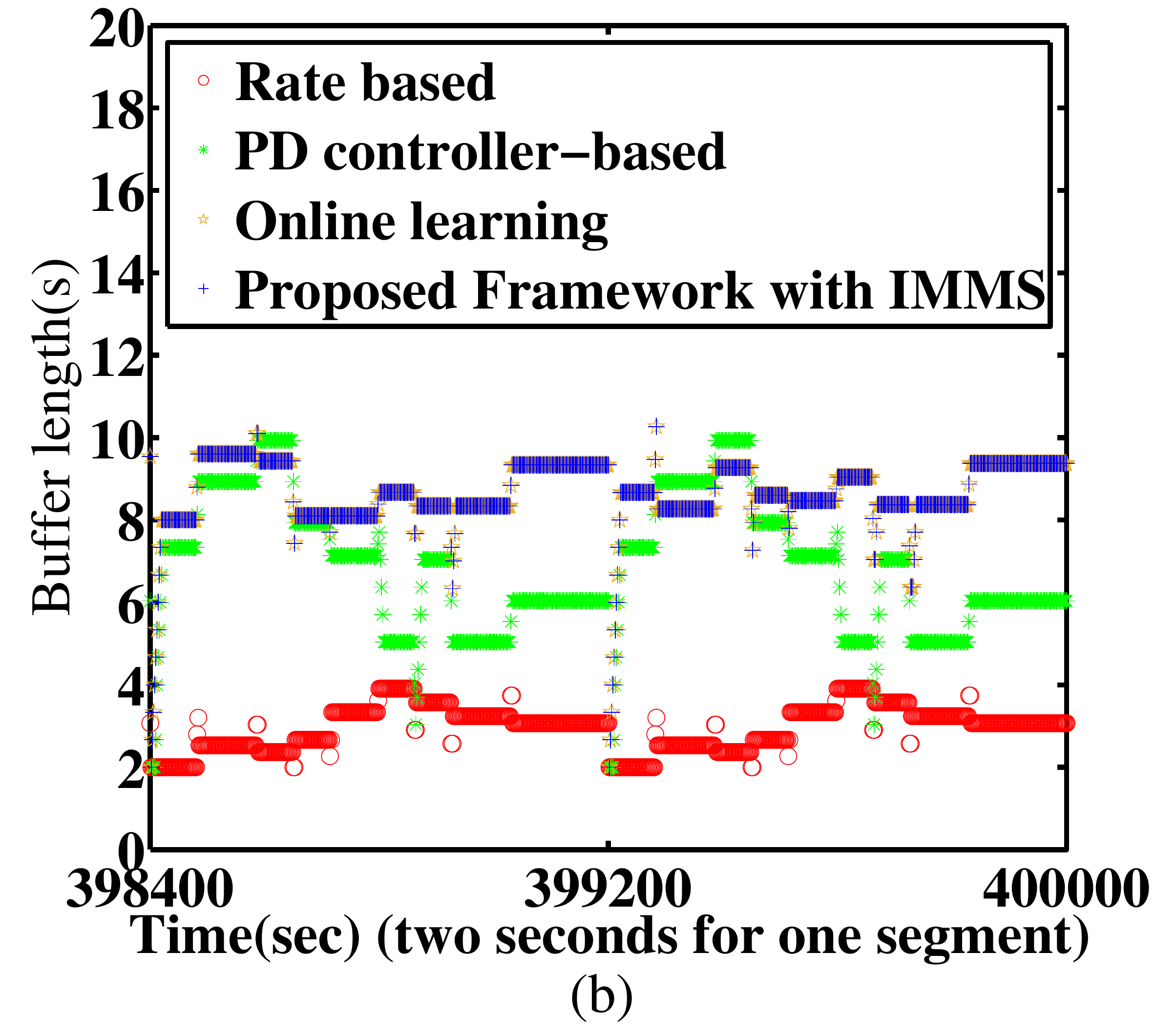}
\caption{Buffer length comparison under the short-term fluctuating
channel.} \label{fig17}
\end{figure}

\begin{figure}
\centering
\includegraphics[width=8.76cm]{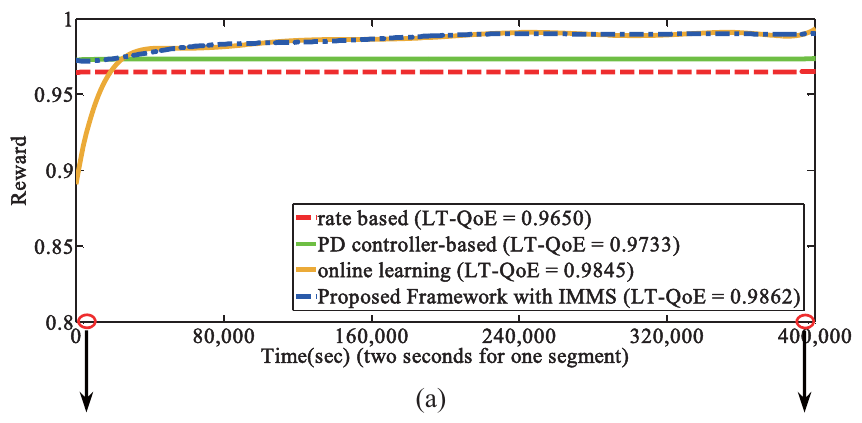}
\includegraphics[width=4.2cm]{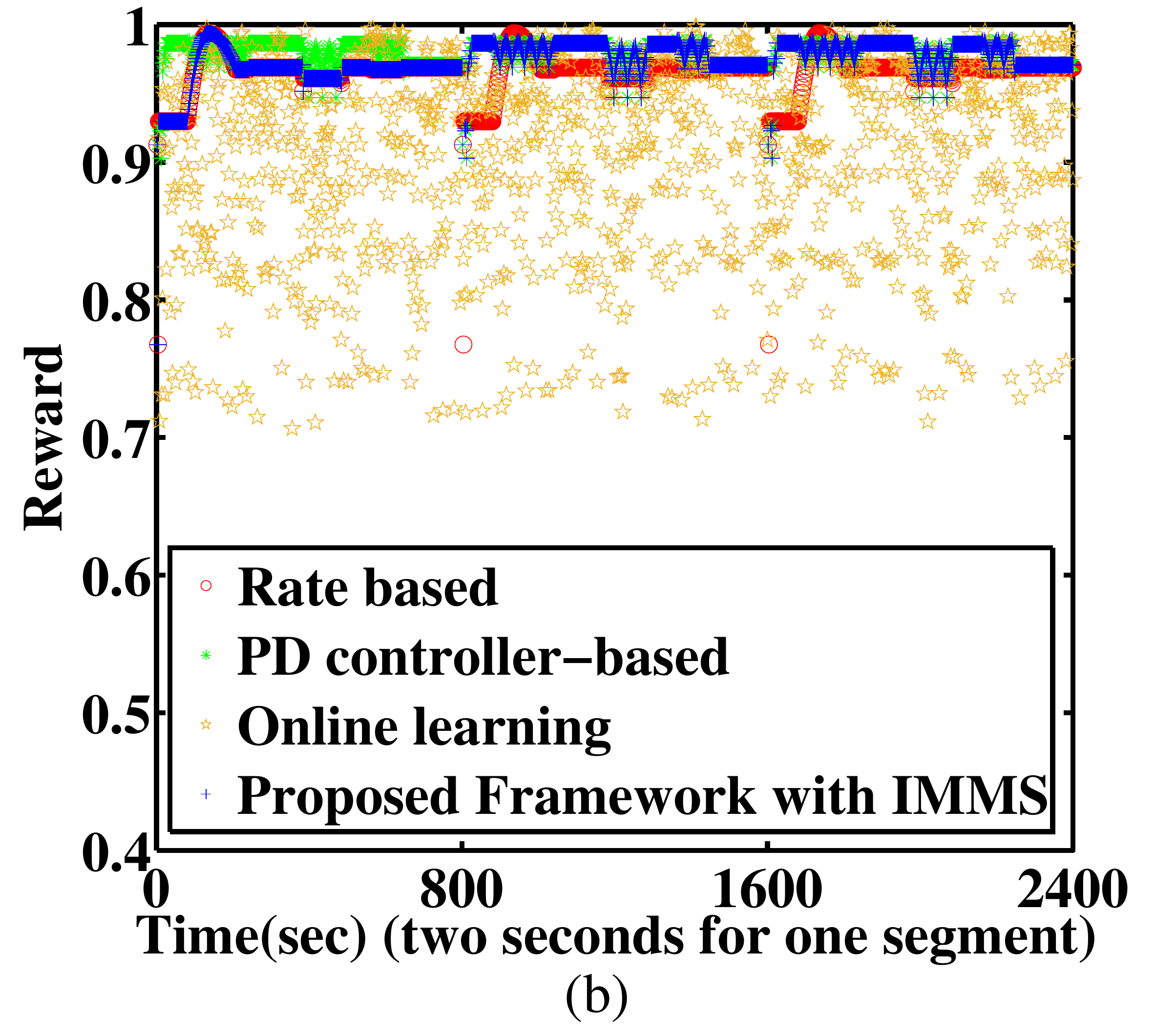}
\includegraphics[width=4.2cm]{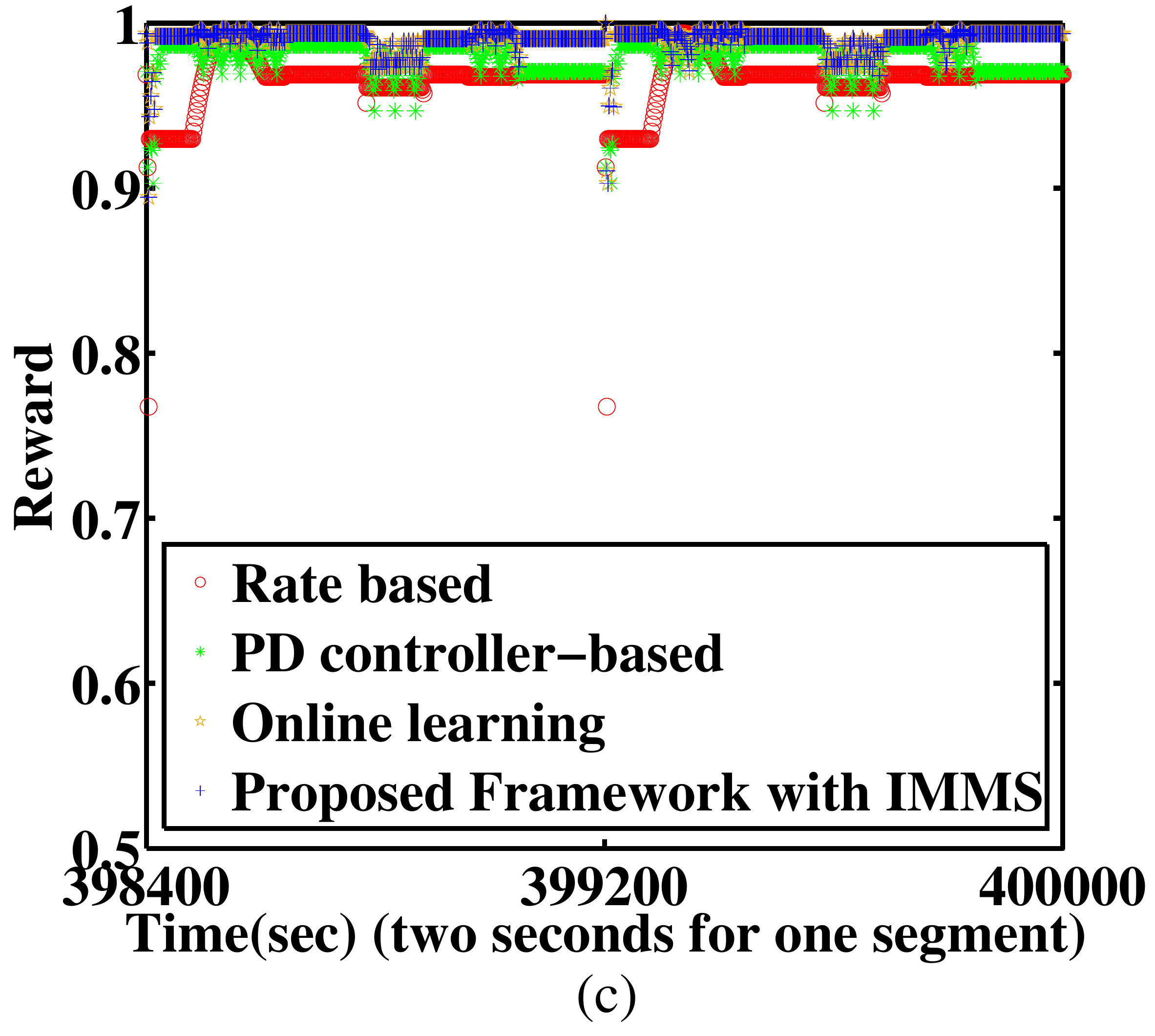}
\caption{Rewards and \emph{LT-QoE}s comparison under the long-term
fluctuating channel.} \label{fig18}
\end{figure}

\begin{figure}
\centering
\includegraphics[width=4.2cm]{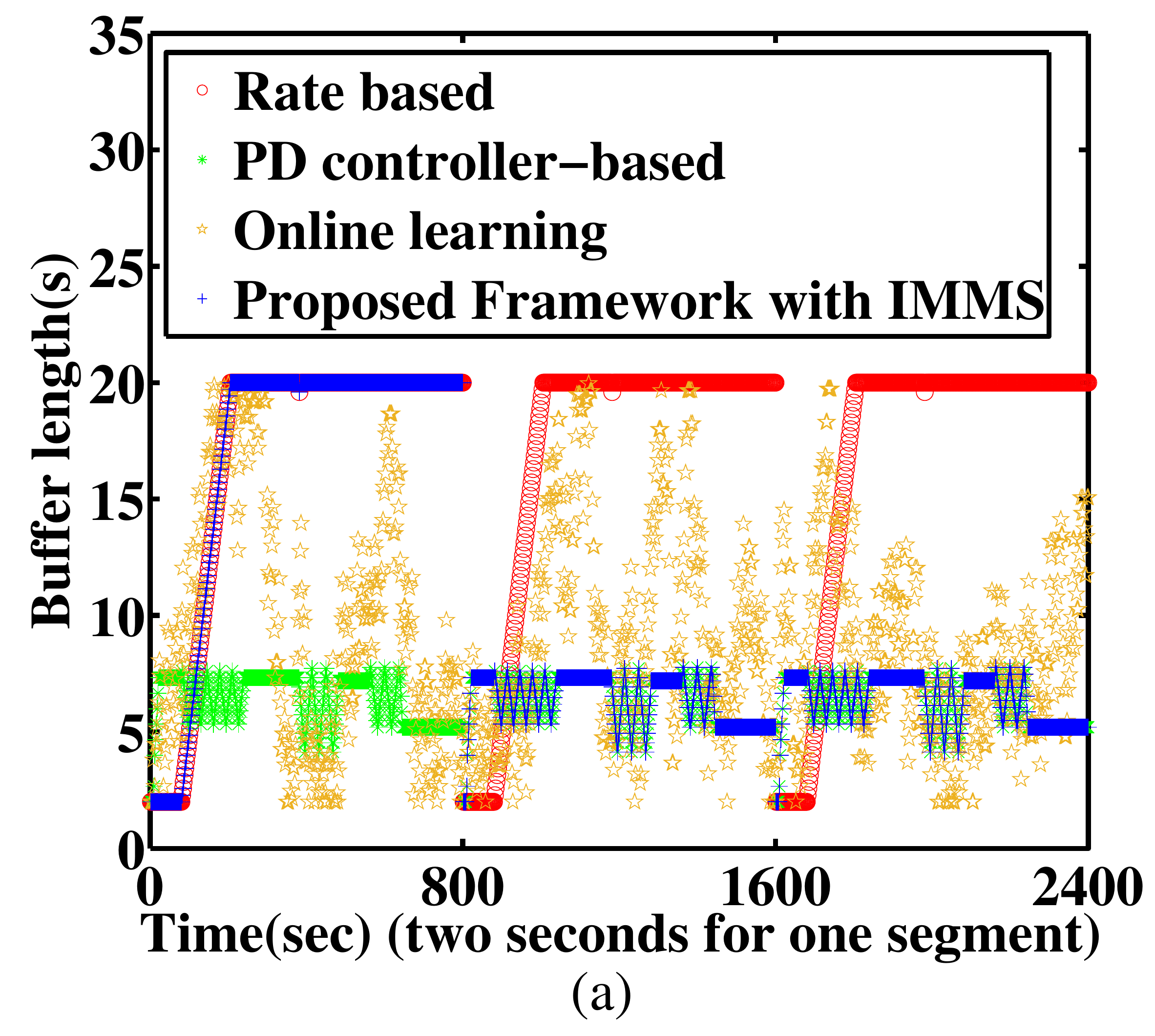}
\includegraphics[width=4.2cm]{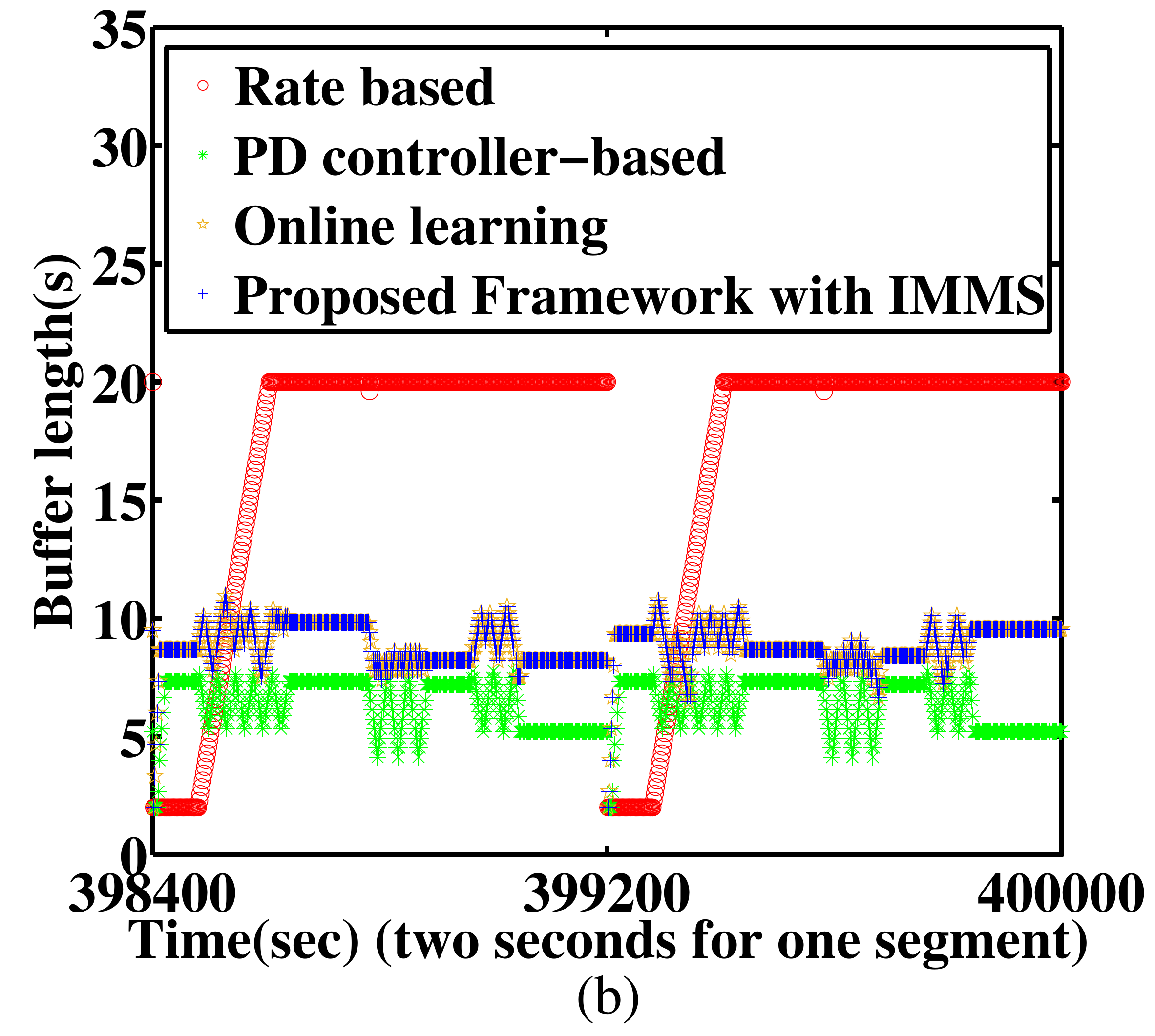}
\caption{Buffer length comparison under the long-term fluctuating
channel.} \label{fig19}
\end{figure}

\begin{figure}
\centering
\includegraphics[width=8.76cm]{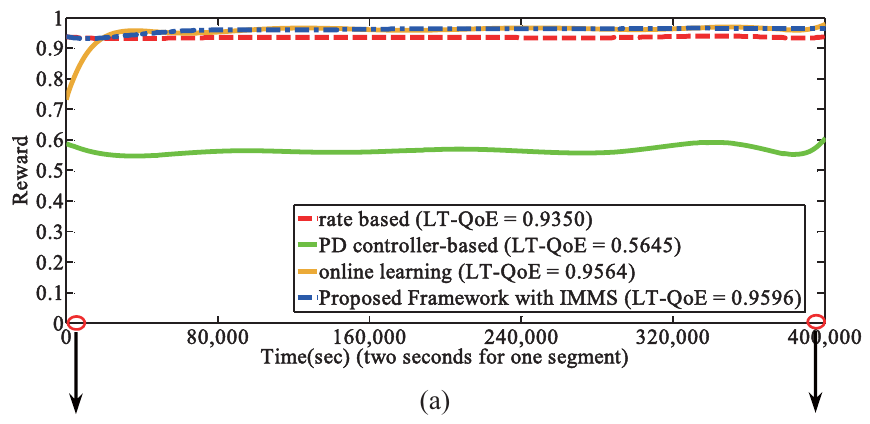}
\includegraphics[width=4.2cm]{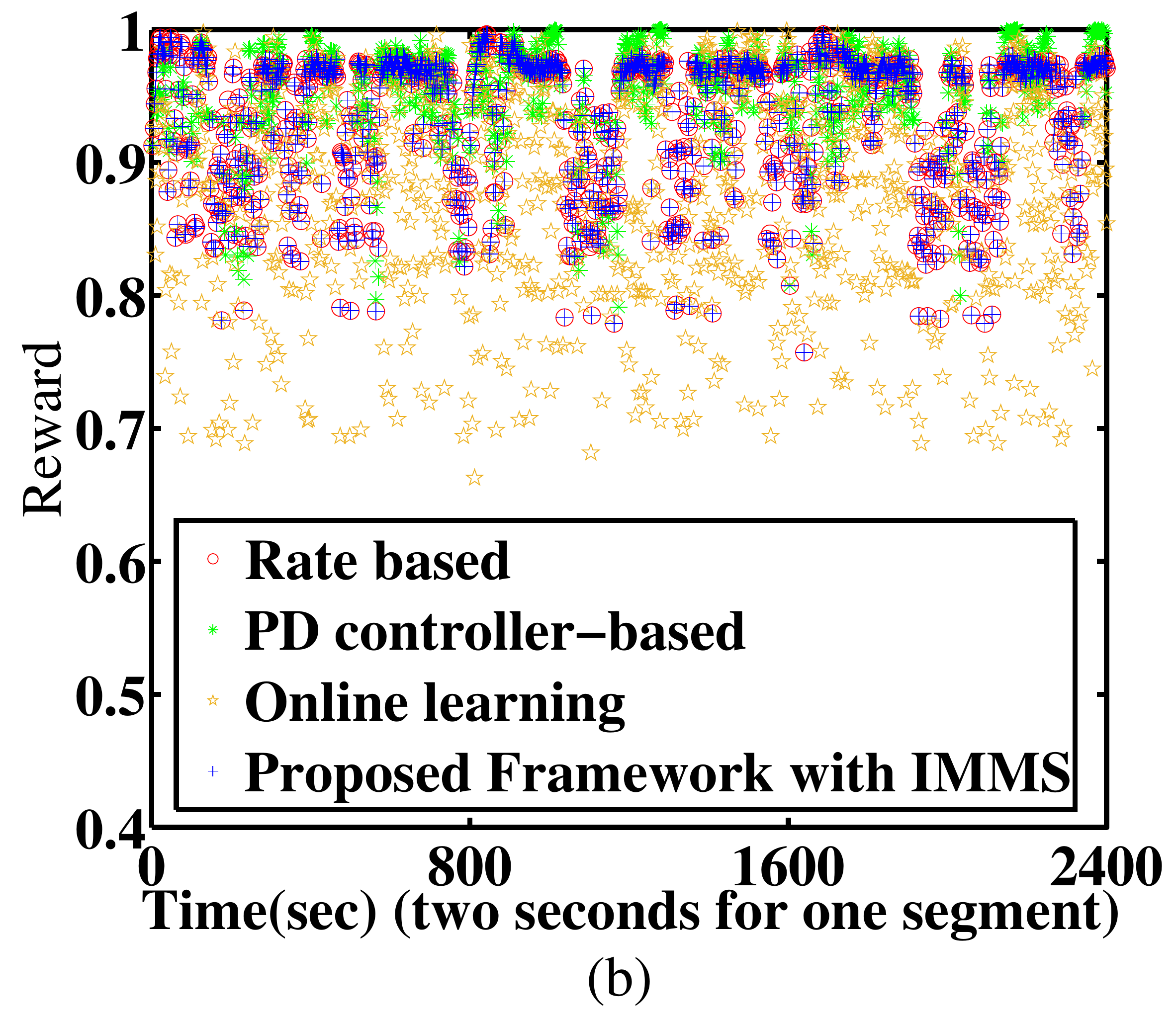}
\includegraphics[width=4.2cm]{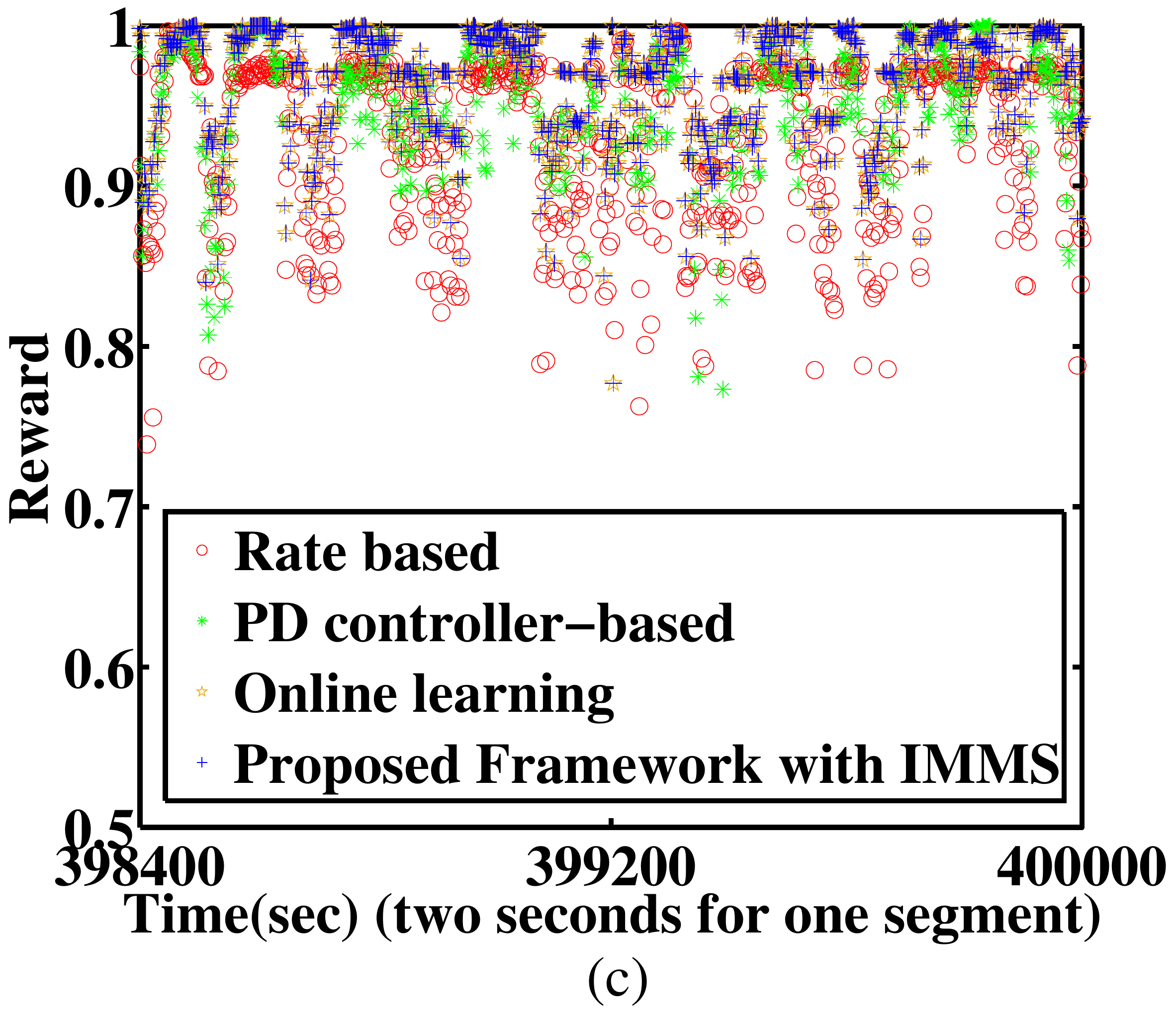}
\caption{Rewards and \emph{LT-QoE}s comparison under the Markov
channel $(p=0.5)$.} \label{fig20}
\end{figure}

\begin{figure}
\centering
\includegraphics[width=4.2cm]{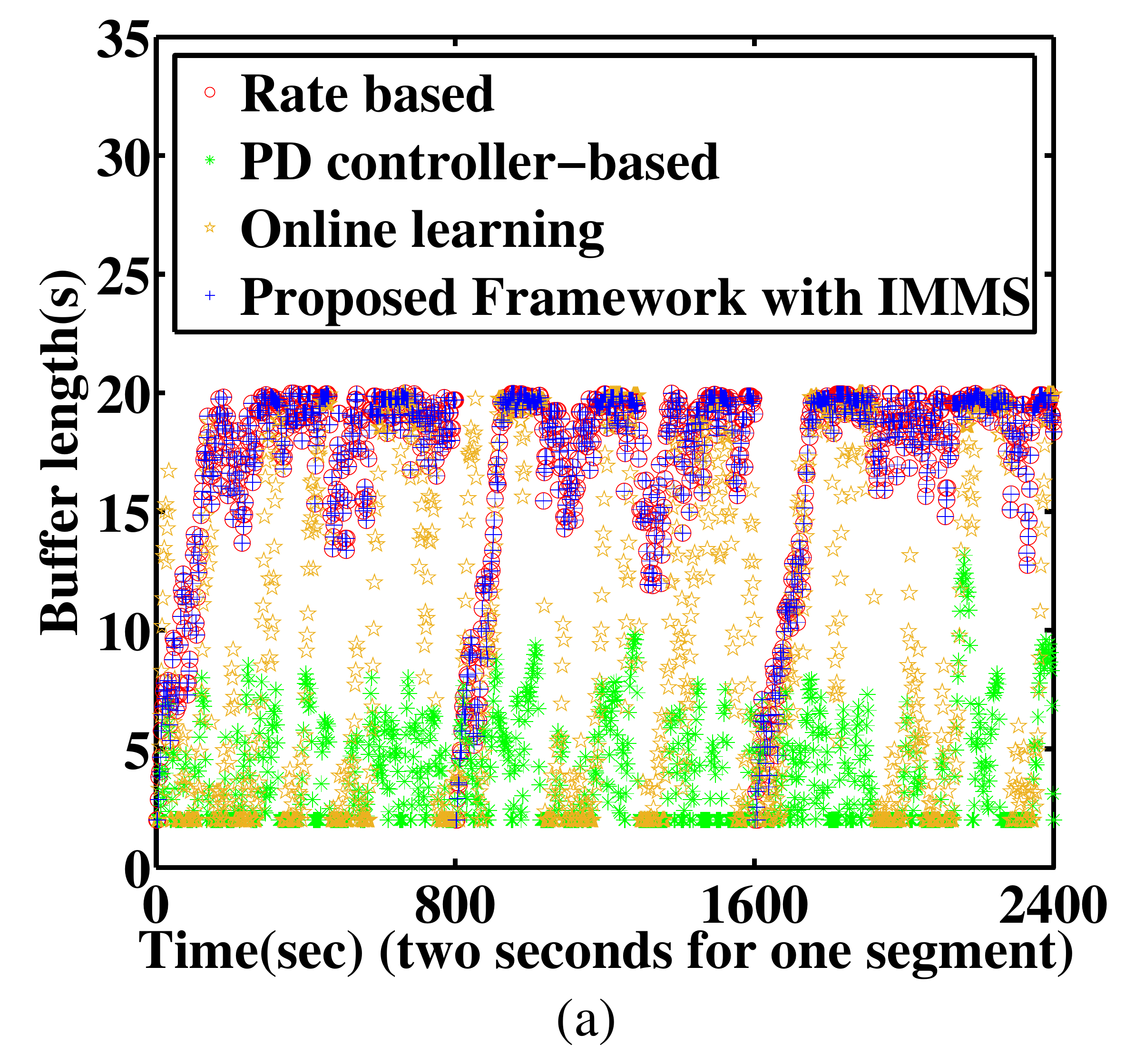}
\includegraphics[width=4.2cm]{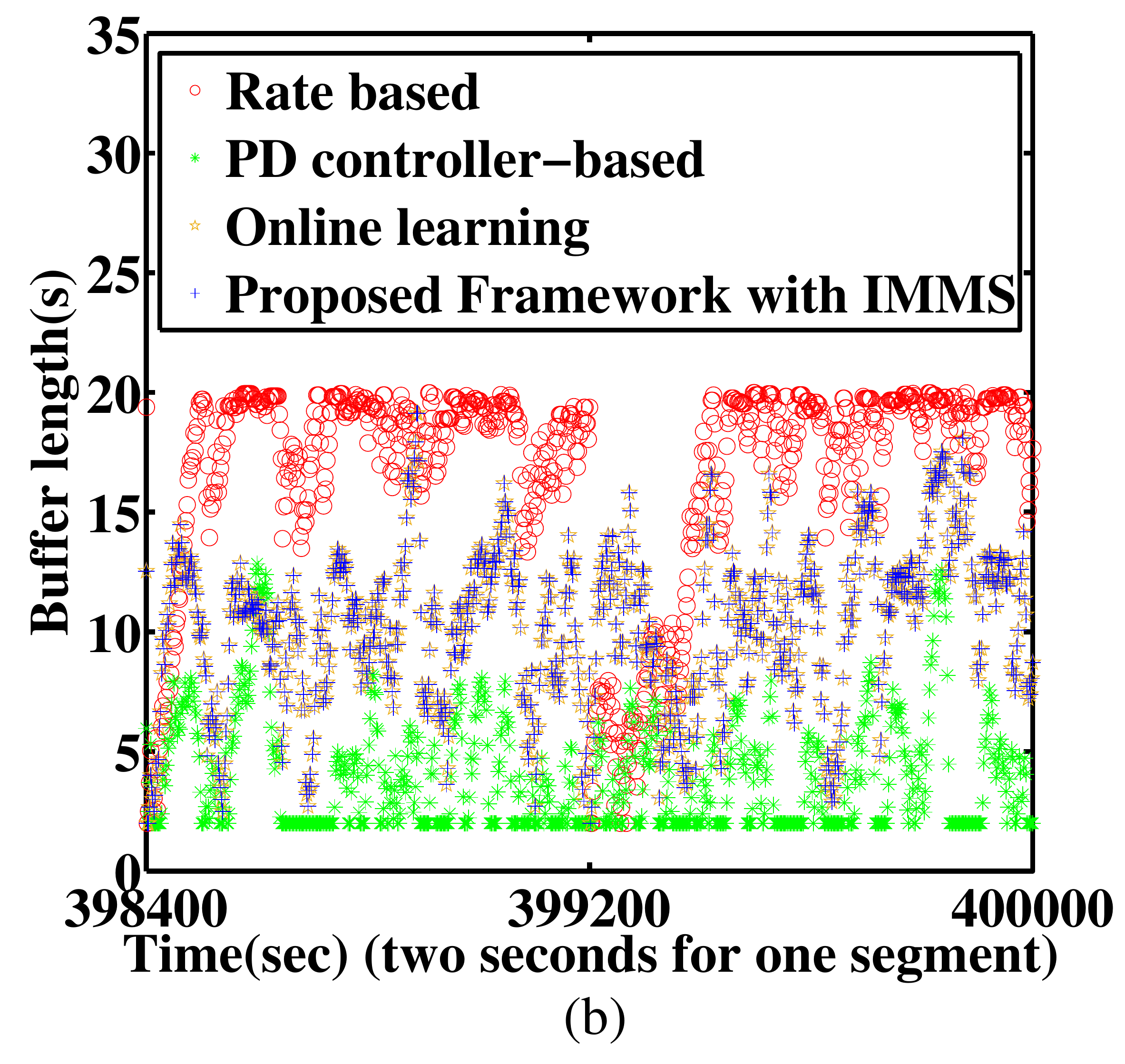}
\caption{Buffer length comparison under the Markov channel
$(p=0.5)$.} \label{fig21}
\end{figure}

\begin{itemize}
\item[(\emph{\romannumeral3})] \emph{Evaluation of the adaptation ability}
\end{itemize}

\begin{figure}
\centering
\includegraphics[width=8.76cm]{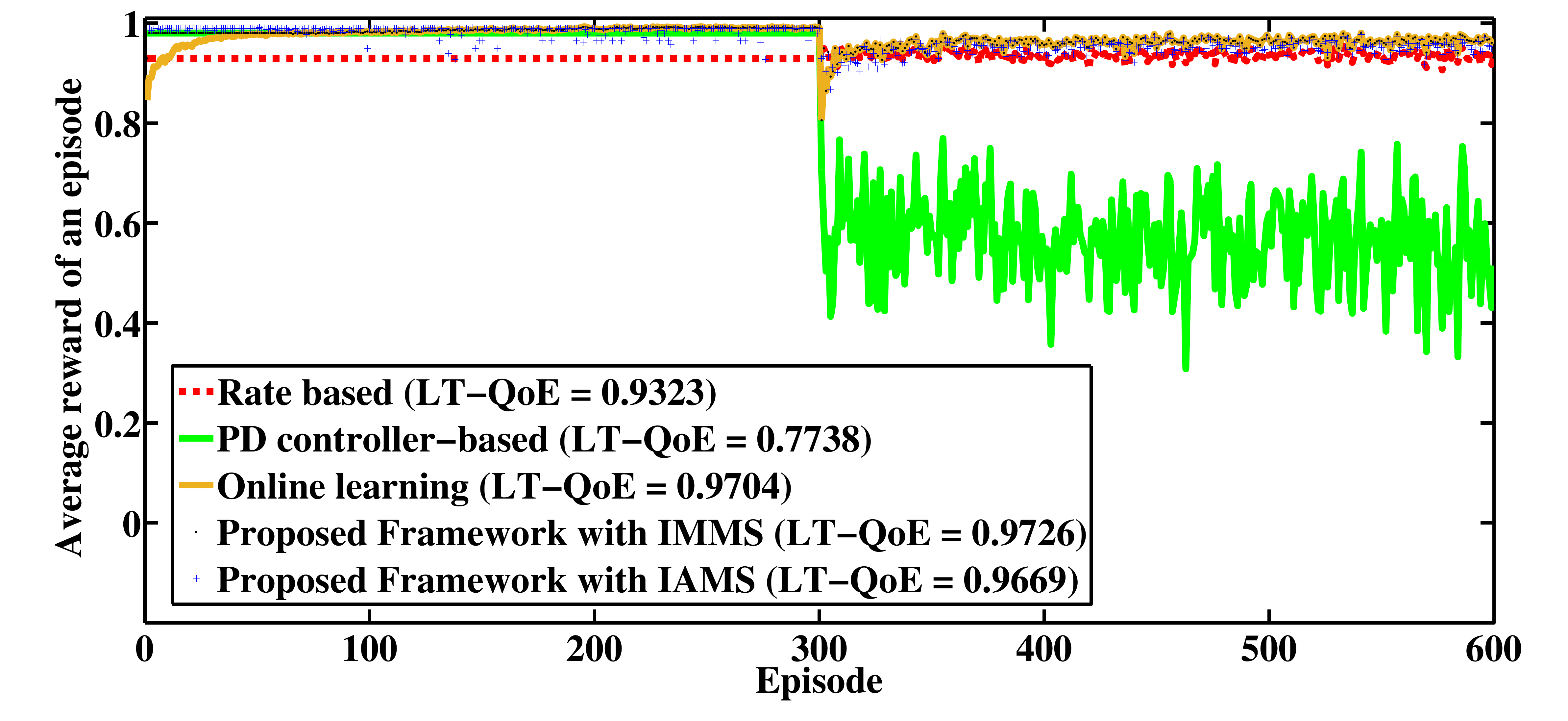}
\caption{Rewards comparison for 600 episodes with abrupt channel
changes at the 300-$th$ episode.} \label{fig22}
\end{figure}

\begin{figure}
\centering
\includegraphics[width=8.76cm]{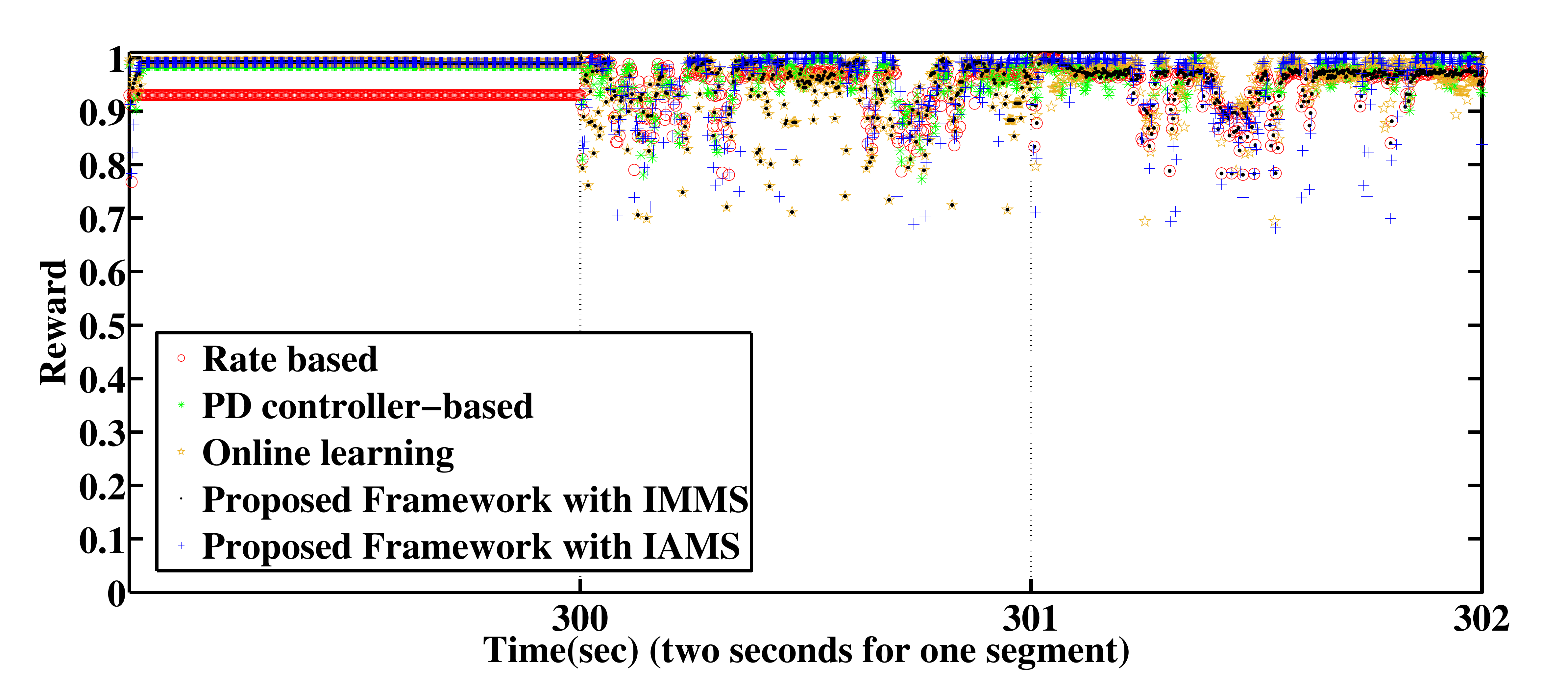}
\caption{Rewards comparison of the 300-$th$, 301-$th$, and the
302-$th$ episodes.} \label{fig23}
\end{figure}

In order to further verify the adaptation ability of the proposed frameworks (both \emph{\textbf{IAMS}} and \emph{\textbf{IMMS}}), we first considered the case that the channel environment changes abruptly. In this simulation, there are 24000 video segments (corresponding to 600 episodes) to be requested. For the first 300 episodes (video complexity level is 4), a stable channel with a bandwidth of 3000kbps is simulated, while for the last 300 episodes, the channel environment changes abruptly to a Markov channel with $p=0.5$. Fig. 22 shows the fitted (also by a nine order polynomial fitting method) curves of the rewards for different rate adaptation methods. We can see that the rewards of the \emph{PD controller-based method} decline more seriously than the other methods. This is because the model parameters of the PD controller-based method cannot adapt the suddenly changed channel environment. We can also observe that the rewards of the \emph{online learning-based method} and the \emph{\textbf{proposed frameworks}} also suffer from sharp changes. But the \emph{LT-QoE} of the \emph{\textbf{proposed framework with IMMS}} is a little larger than that of the \emph{online learning-based method}. To show the instantaneous rewards clearly at the changing time, Fig. 23 compares the detailed instantaneous rewards of different rate adaptation methods at the changing time.

Besides, the adaptation ability is also investigated on the video
complexity variation. In this simulation, there are also 600
episodes to be requested. A Markov channel with $p=0.5$ is used. For
the first 300 episodes, the complexity is 5, while for the last
300-$th$ episode, the video complexity level randomly changes from 5
to 1. Fig. 24 shows the rewards of each episode for different
methods. We see that the performance of the rate-based method is the
worst since it cannot adapt to the Markov channel efficiently. Since
the model parameters of the \emph{PD controller-based method} are
not suitable, the performance of this method is also not very good.
We can also see that the performances of the proposed frameworks are
a little better than the online learning-based method.

\begin{figure}
\centering
\includegraphics[width=8.76cm]{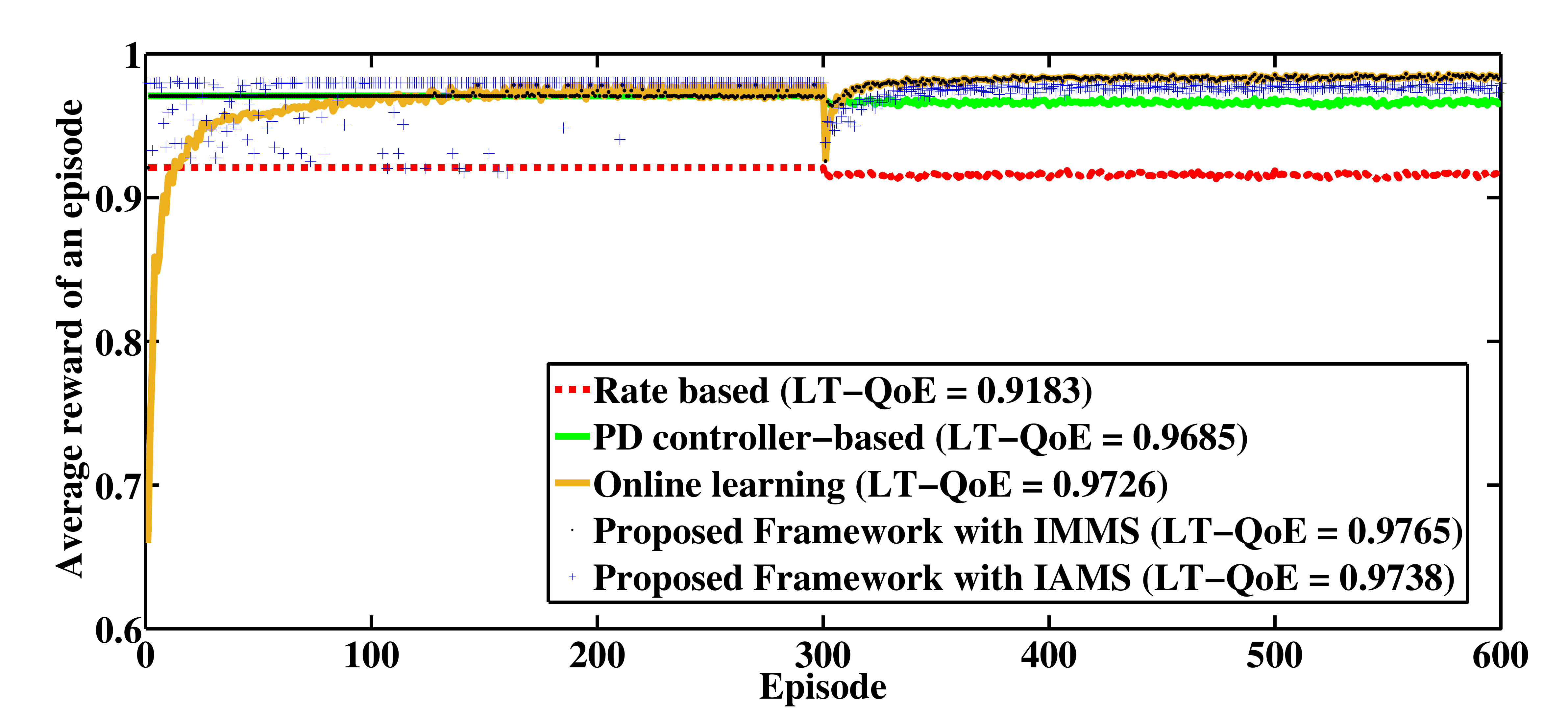}
\caption{Rewards comparison under the condition that the video
complexity changes from a constant one ($c=5$) to random values
after the 300-$th$ episode.} \label{fig24}
\end{figure}

\begin{figure}
\centering
\includegraphics[width=8.76cm]{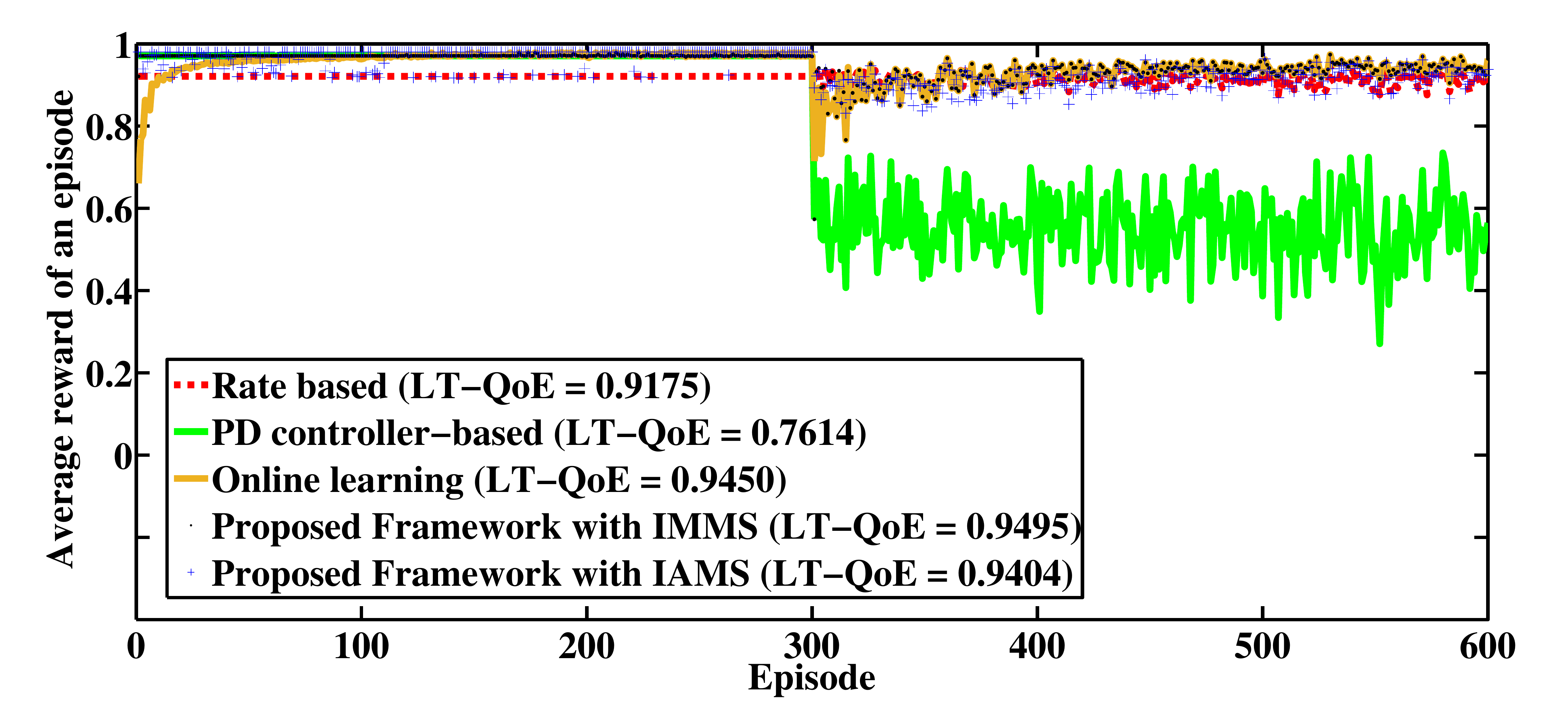}
\caption{Rewards comparison under the condition that both the
channel state and the video complexity changes at the 300-$th$
episode} \label{fig25}
\end{figure}

Moreover, we also tested the dynamic case that both the channel
environment and the video complexity level are changed. In this
case, the channel environment is changed from a static channel to a
Markov channel with parameter $p=0.5$ at the 300-$th$ episode, while
the video complexity is randomly changed for each segment after the
300-$th$ episode, as shown in Fig. 25. We can see that the
performance of the rate-based method is the worst. The performance
of the \emph{PD controller-based method} is also not good enough
since it cannot adapt to the frequent fluctuations of channel
environments. The \emph{online learning-based method} can adapt to
the complexity fluctuations to some extent, but the
\textbf{\emph{proposed frameworks}} (\textbf{\emph{IMMS}} and
\textbf{\emph{IAMS}}) perform better. Also we should note that the
\emph{LT-QoE} of the online learning-based method is similar to
that of the proposed frameworks as shown in Fig. 25. The reason is
that the long term simulation time is propitious to the online
learning-based method.

\begin{itemize}
\item[(\emph{\romannumeral4})] \emph{Comparison with state-of-the-art methods}
\end{itemize}

\begin{figure}
\centering
\includegraphics[width=8.76cm]{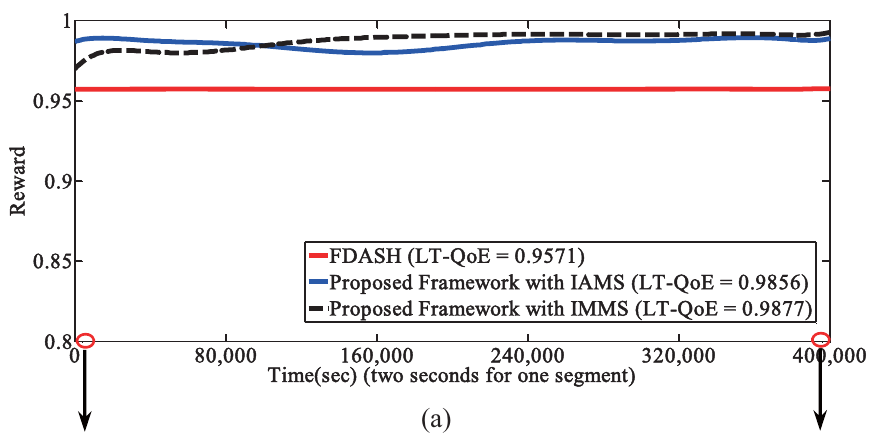}
\includegraphics[width=4.2cm]{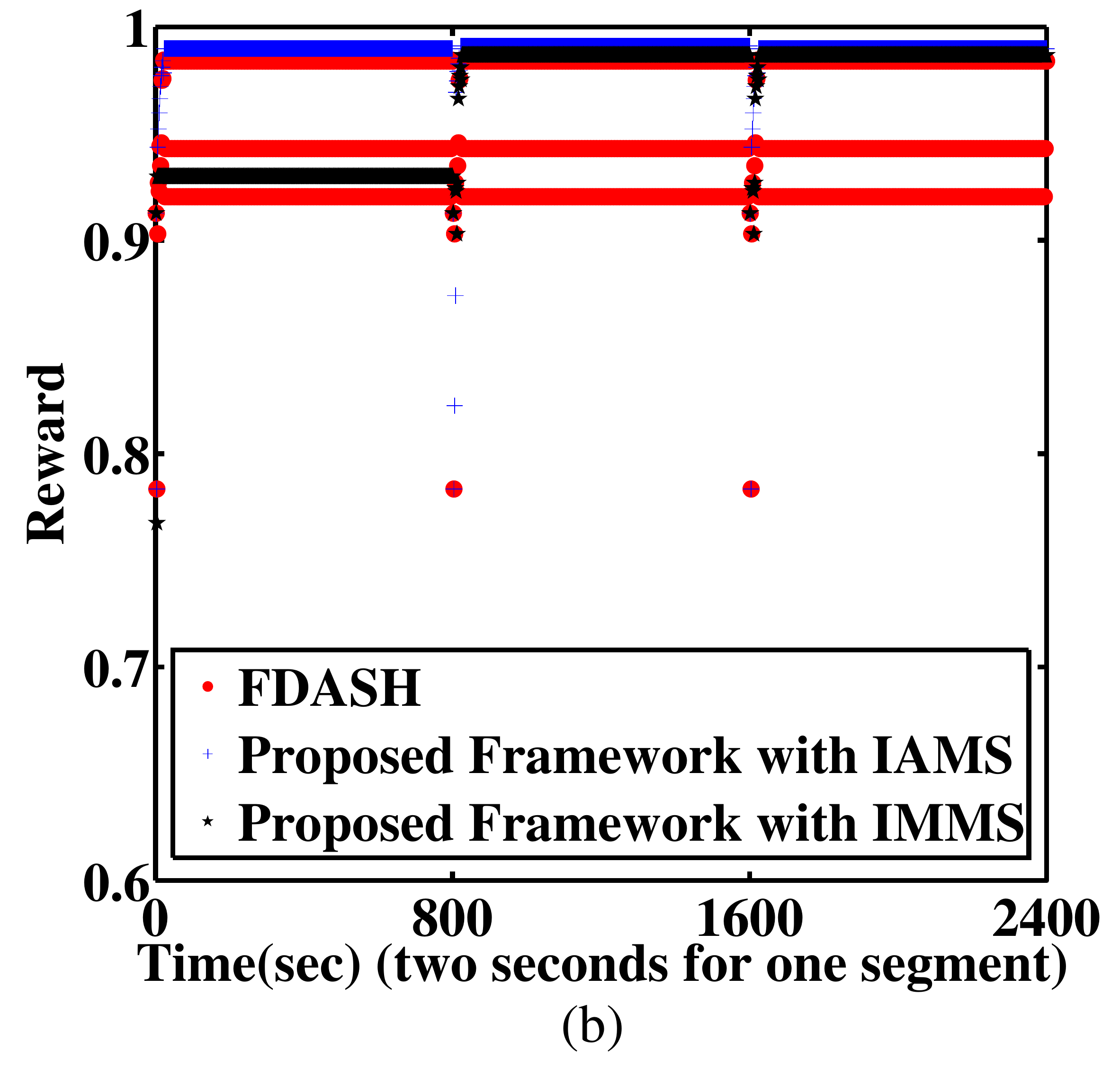}
\includegraphics[width=4.2cm]{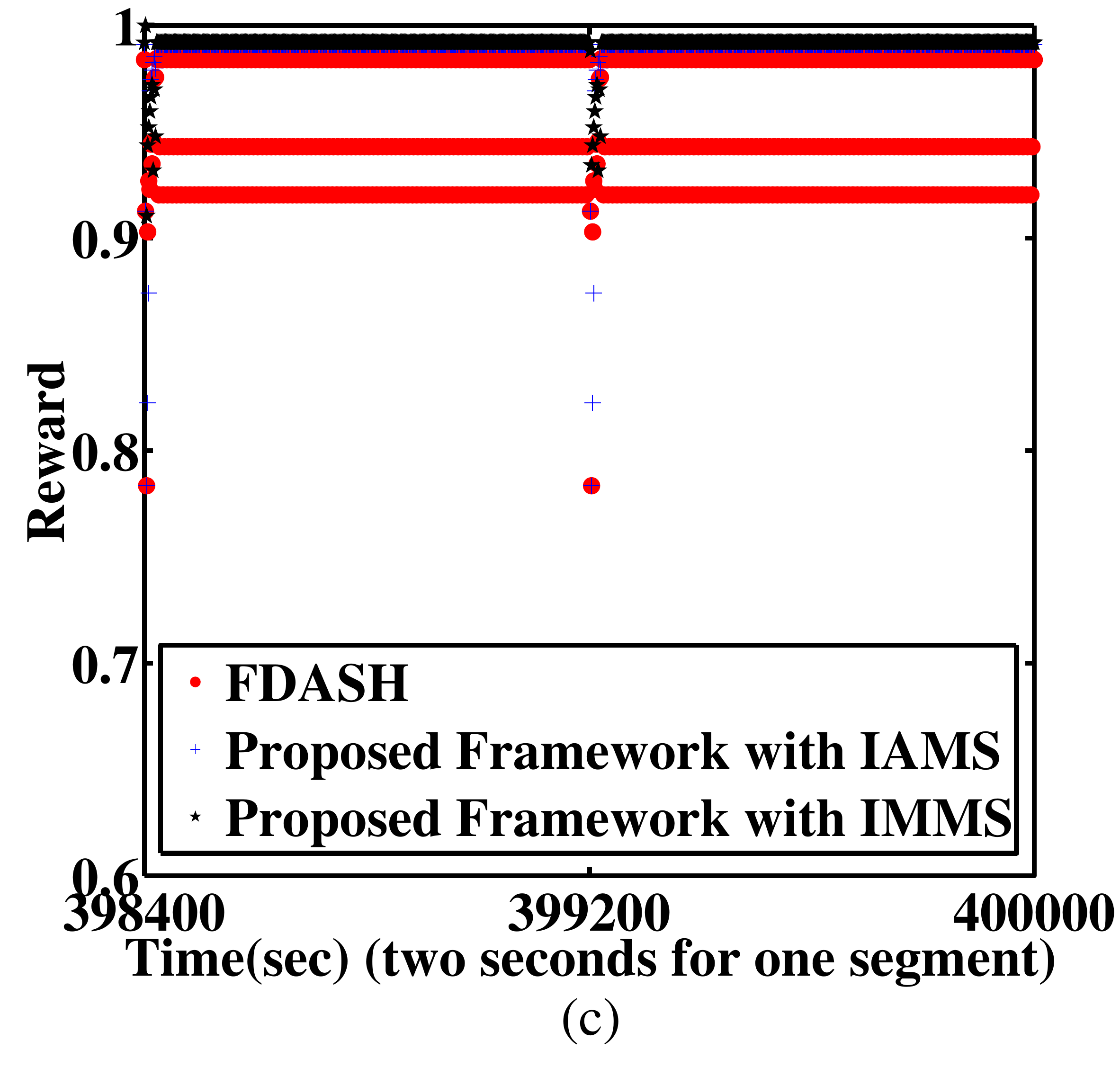}
\caption{Rewards and \emph{LT-QoE}s comparisons under the constant
channel.} \label{fig26}
\end{figure}
\begin{figure}
\centering
\includegraphics[width=4.2cm]{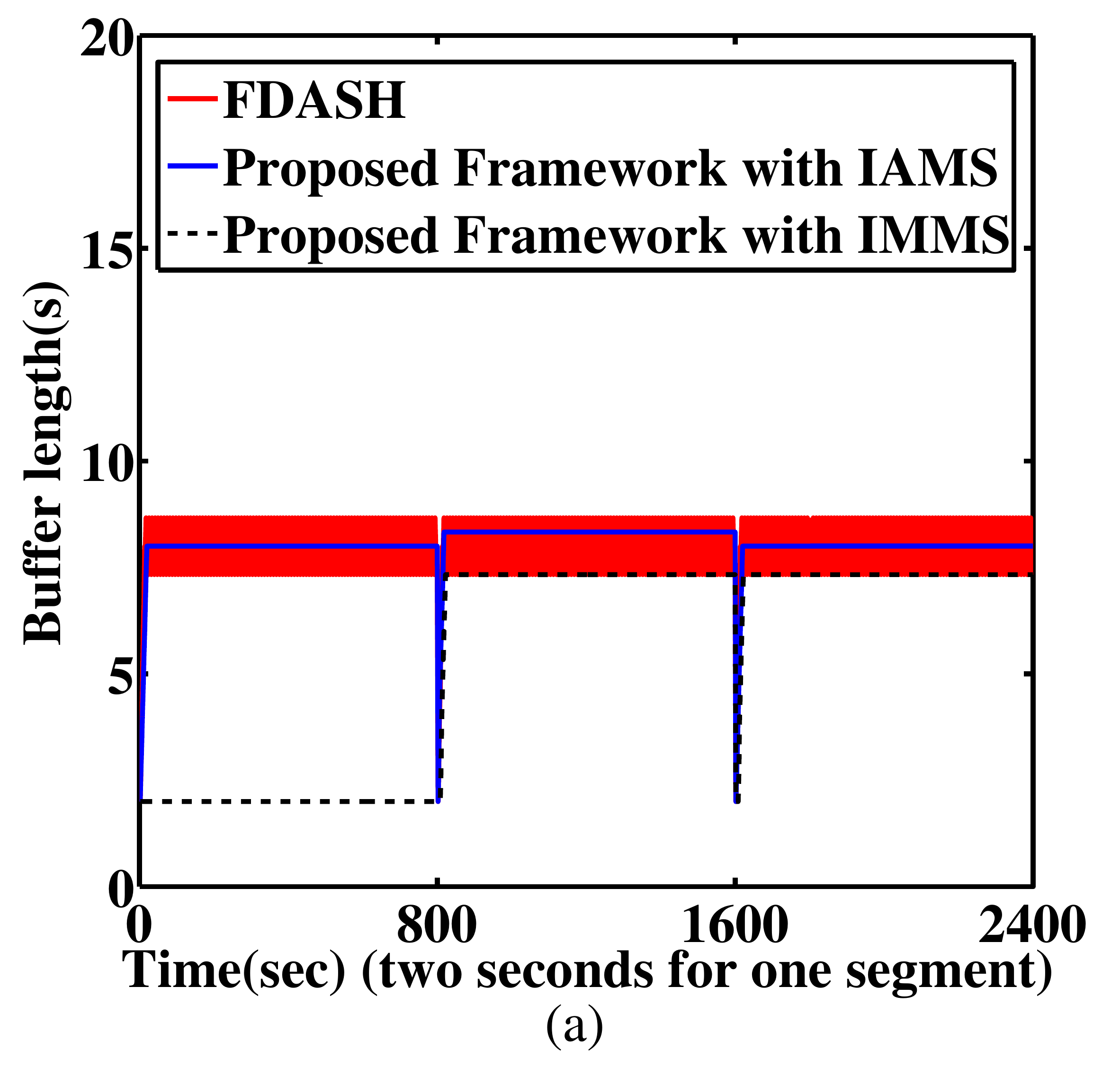}
\includegraphics[width=4.2cm]{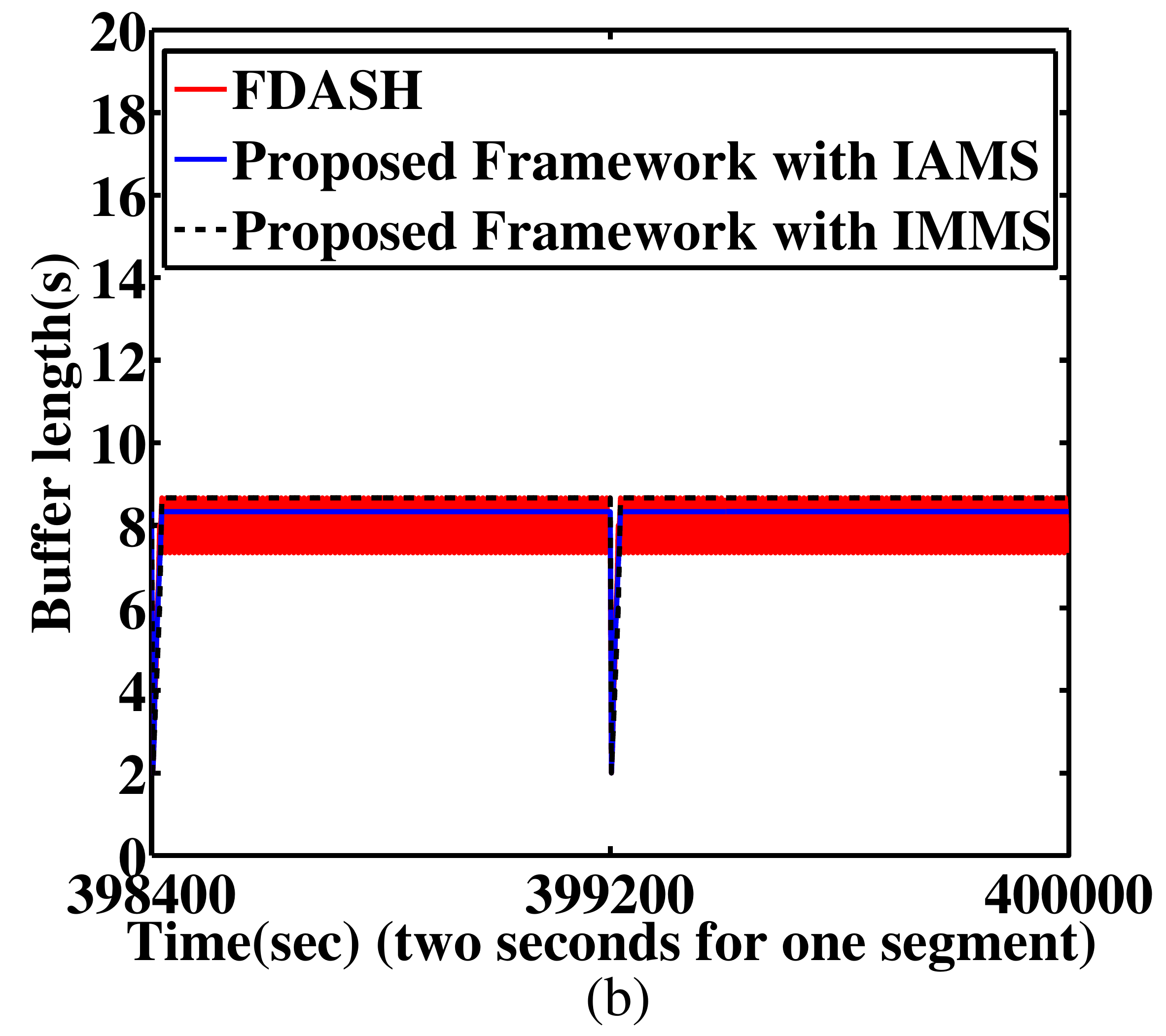}
\caption{Buffer length comparisons under the constant channel.}
\label{fig27}
\end{figure}

In order to further validate the performance of the proposed
frameworks (\textbf{\emph{IMMS}} and \textbf{\emph{IAMS}}), we have
also compared them with a fuzzy logic-based rate adaptation method
(denoted by \textbf{\emph{FDASH}}) [53] under different channels.
Although the \textbf{\emph{FDASH}} method can also be integrated
into the proposed frameworks, we should note that we only integrated
the rate-based method, the PD controller-based method, and the
online learning-based method into the \textbf{\emph{IMMS}} and
\textbf{\emph{IAMS}} frameworks. For the constant channel, the
rewards and \emph{LT-QoE}s are compared in Fig. 26, while the
corresponding buffer lengths are compared in Fig. 27. We can see
that the buffer lengths of the \textbf{\emph{FDASH}} are more stable
than those of the proposed frameworks, but both the rewards and the
\emph{LT-QoE} of each of the proposed frameworks are larger than
those of \textbf{\emph{FDASH}}. Similar results can also be observed
from Figs. 28-32, for the other channels.

\begin{itemize}
\item[(\emph{\romannumeral5})] \emph{Evaluation with commonly used QoE models}
\end{itemize}

\begin{figure}
\centering
\includegraphics[width=8.76cm]{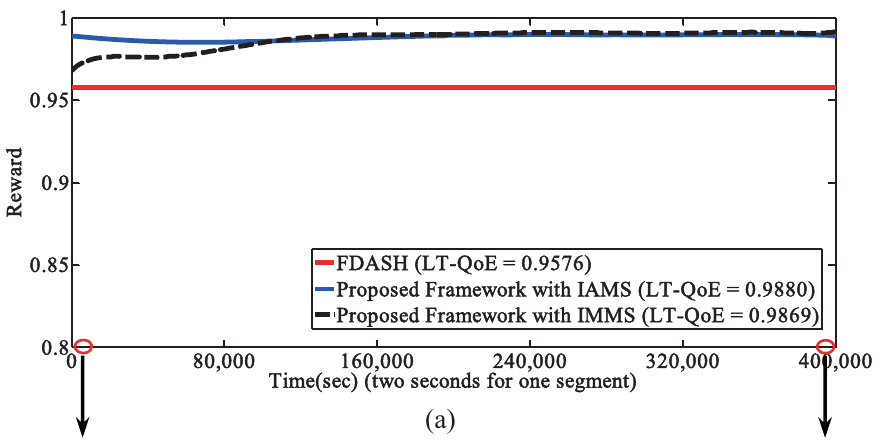}
\includegraphics[width=4.2cm]{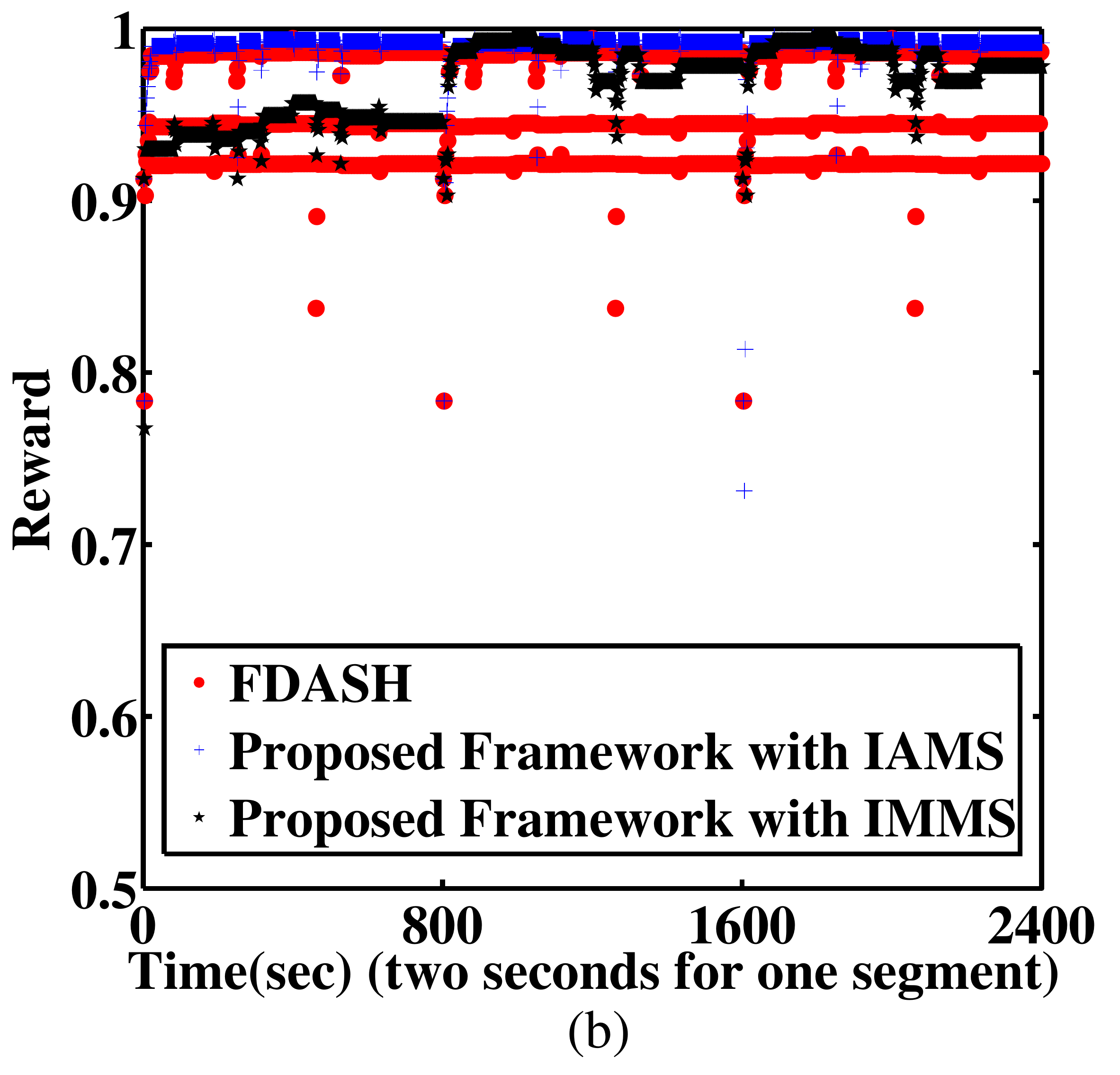}
\includegraphics[width=4.2cm]{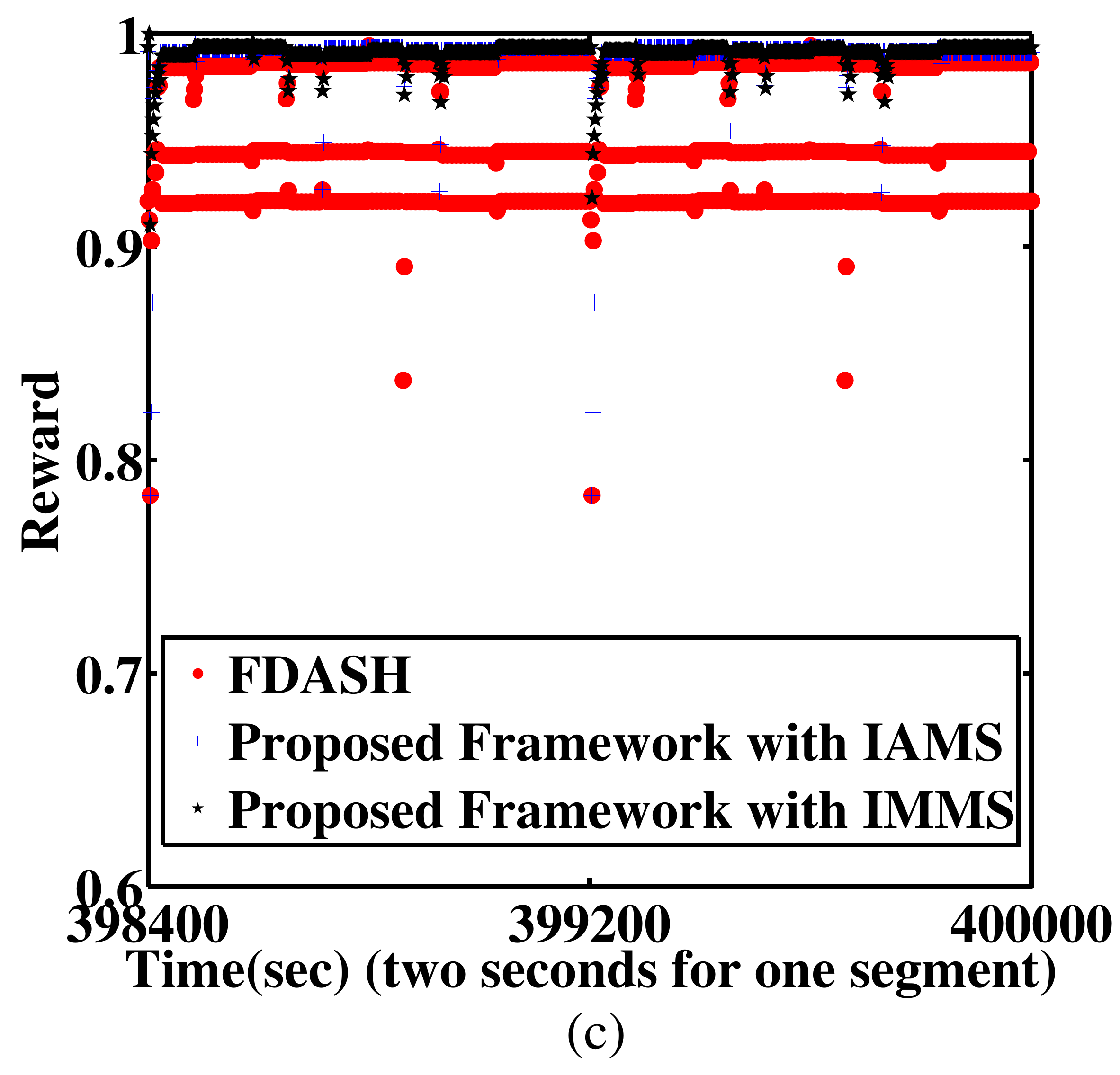}
\caption{Rewards and \emph{LT-QoE}s comparisons under the short-term
fluctuating channel.} \label{fig28}
\end{figure}
\begin{figure}
\centering
\includegraphics[width=4.2cm]{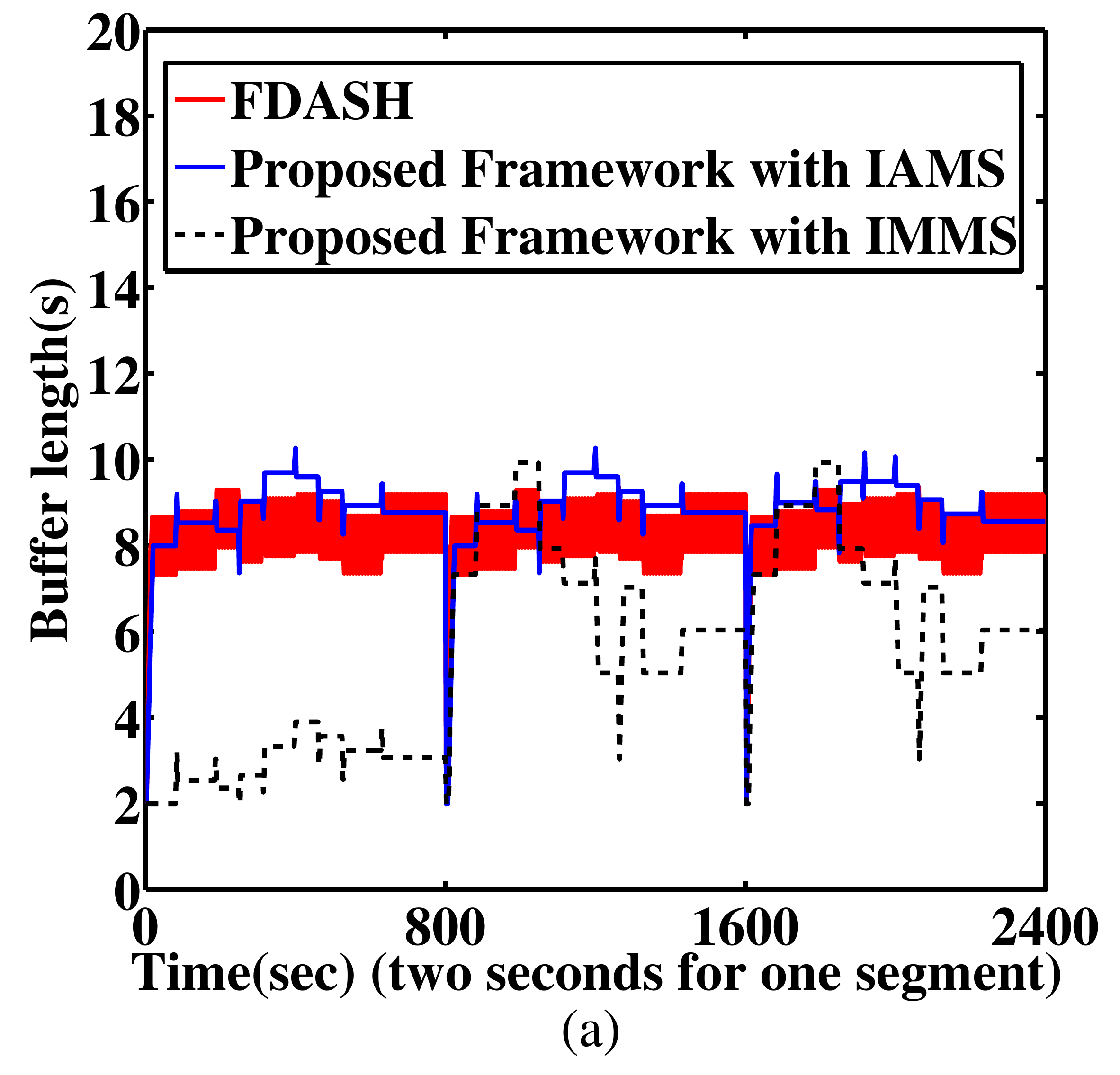}
\includegraphics[width=4.2cm]{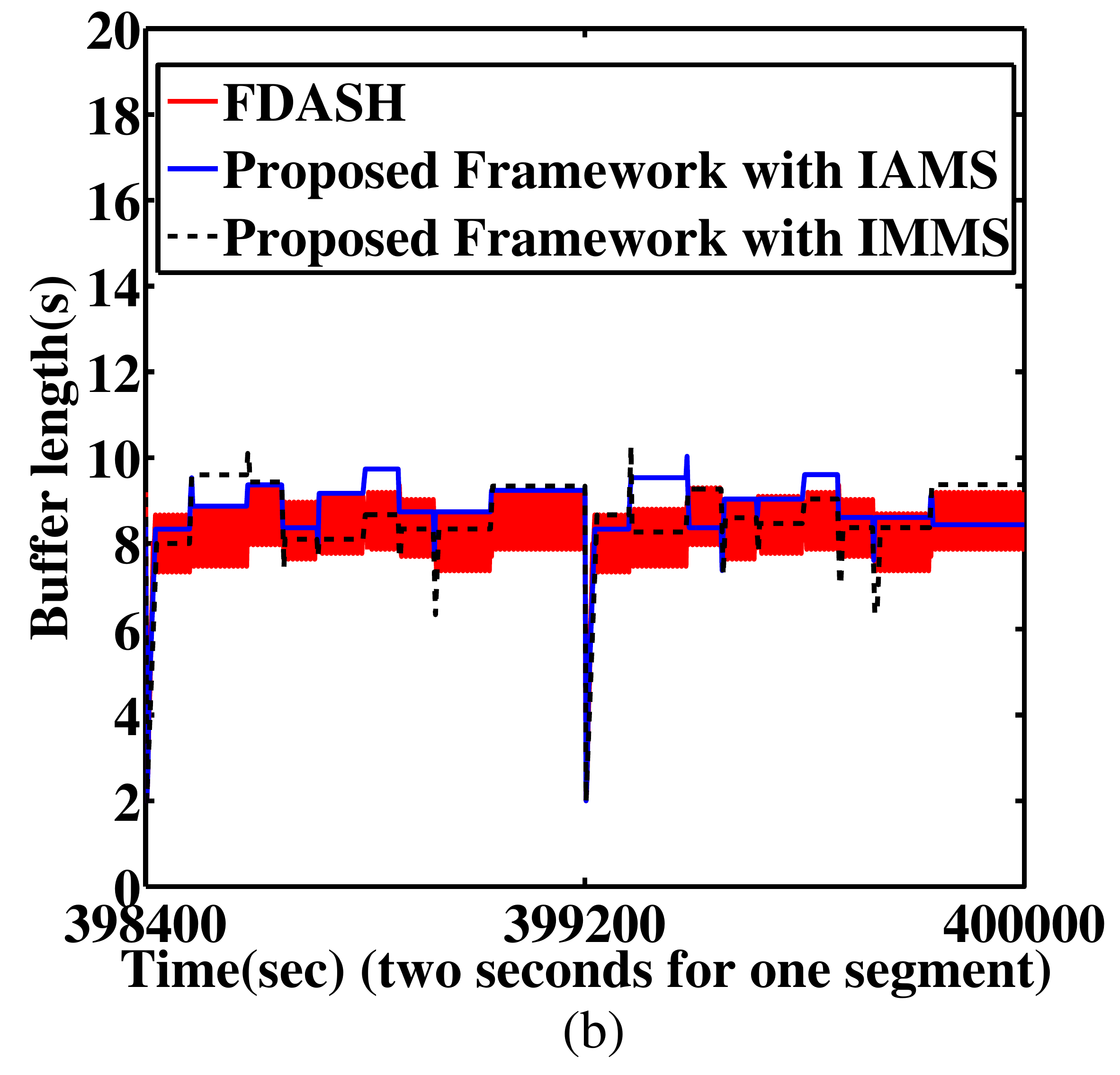}
\caption{Buffer length comparisons under the short-term fluctuating
channel.} \label{fig29}
\end{figure}

\begin{figure}
\centering
\includegraphics[width=8.76cm]{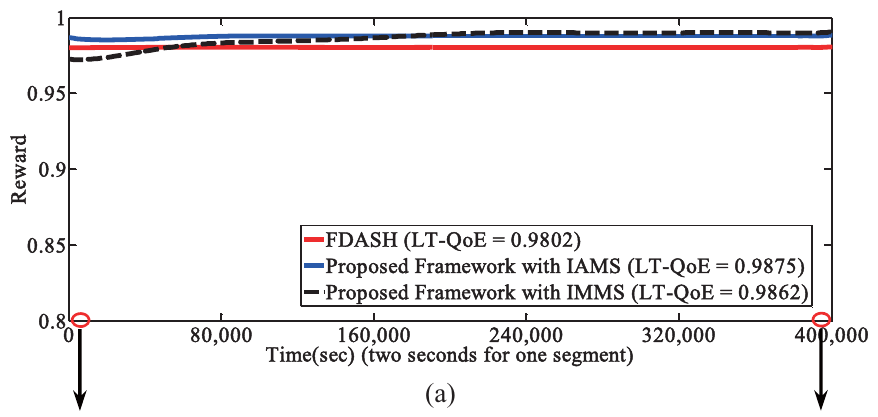}
\includegraphics[width=4.2cm]{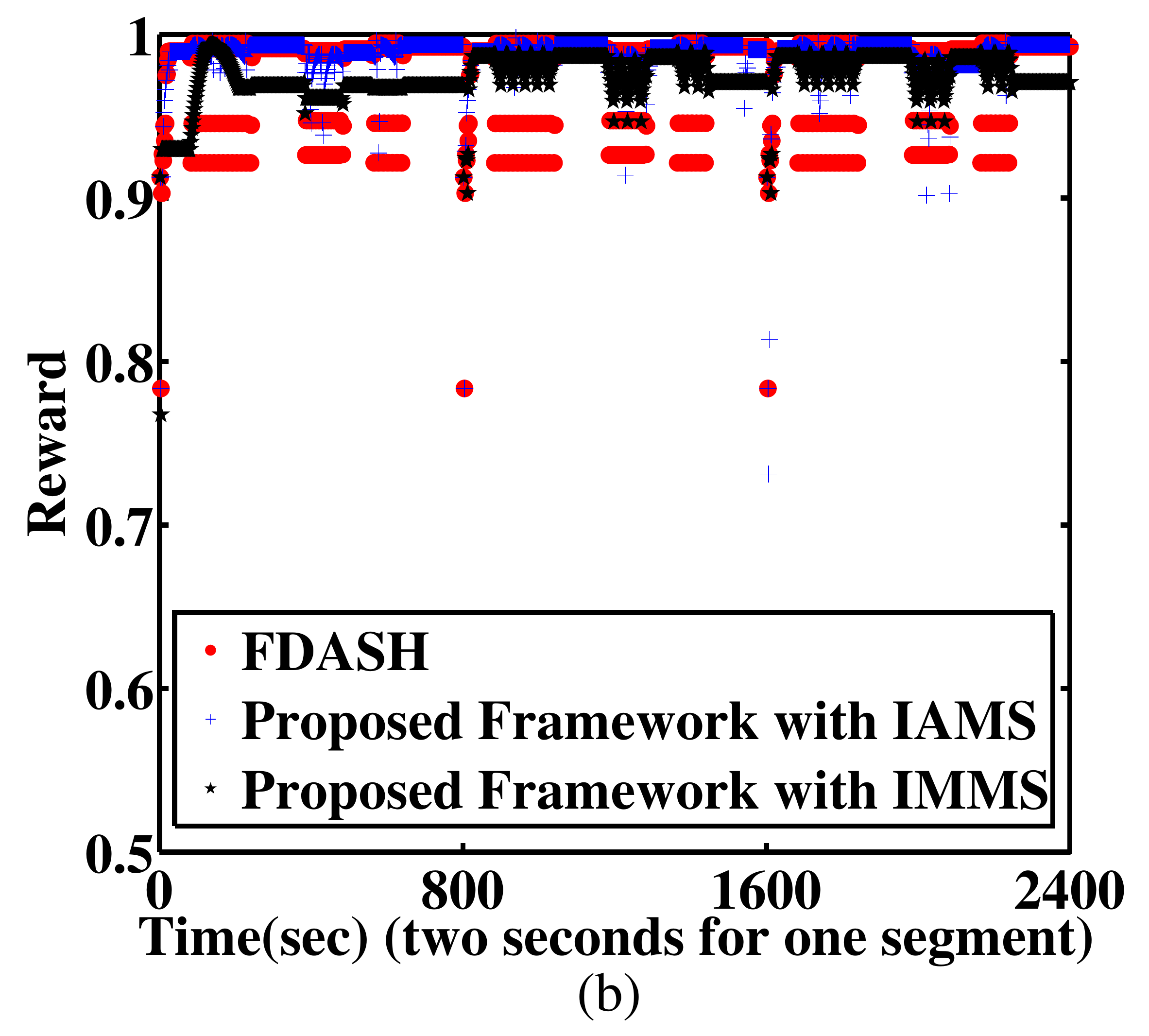}
\includegraphics[width=4.2cm]{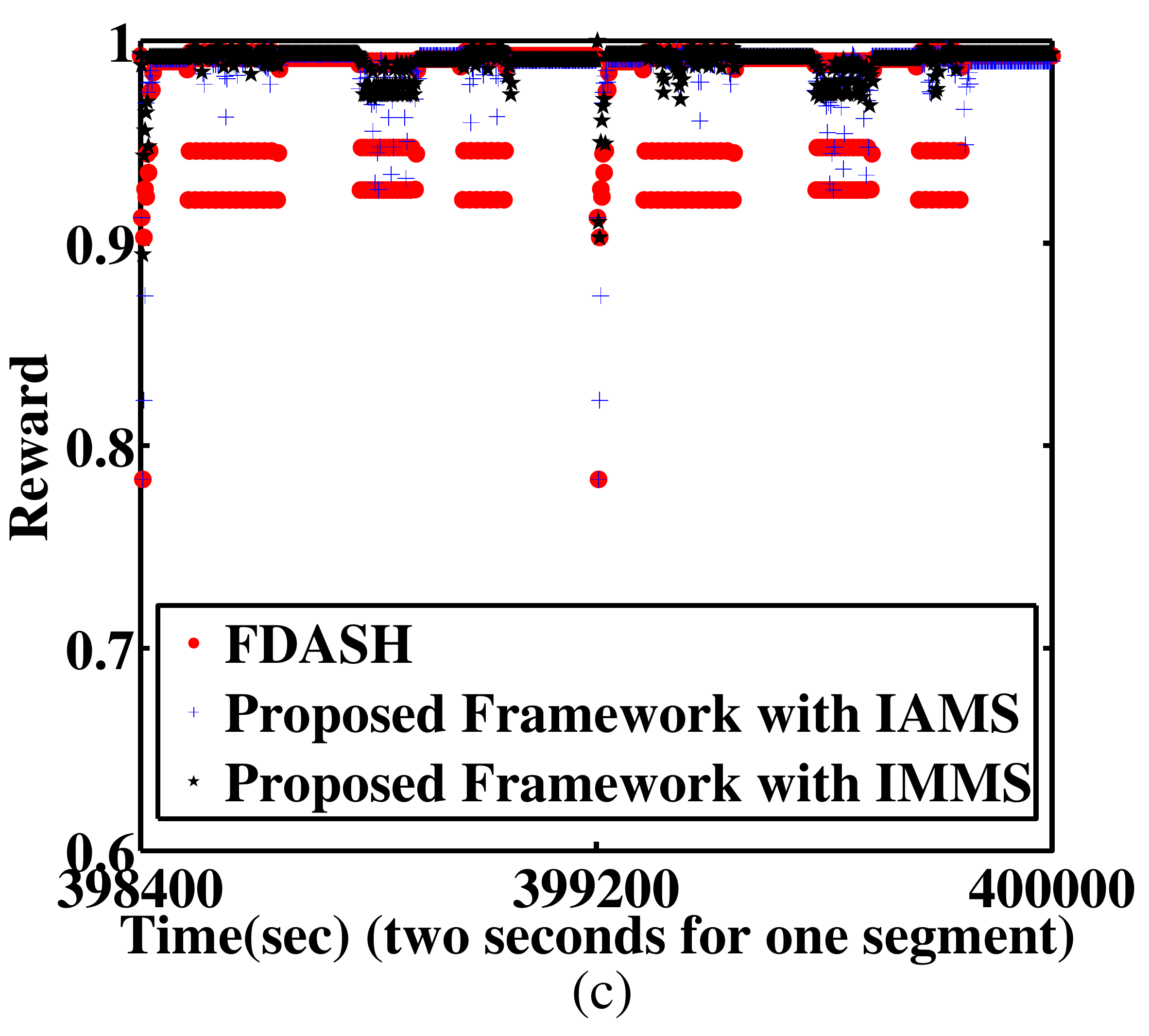}
\caption{Rewards and \emph{LT-QoE}s comparisons under the long-term
fluctuating channel.} \label{fig30}
\end{figure}

\begin{figure}
\centering
\includegraphics[width=4.2cm]{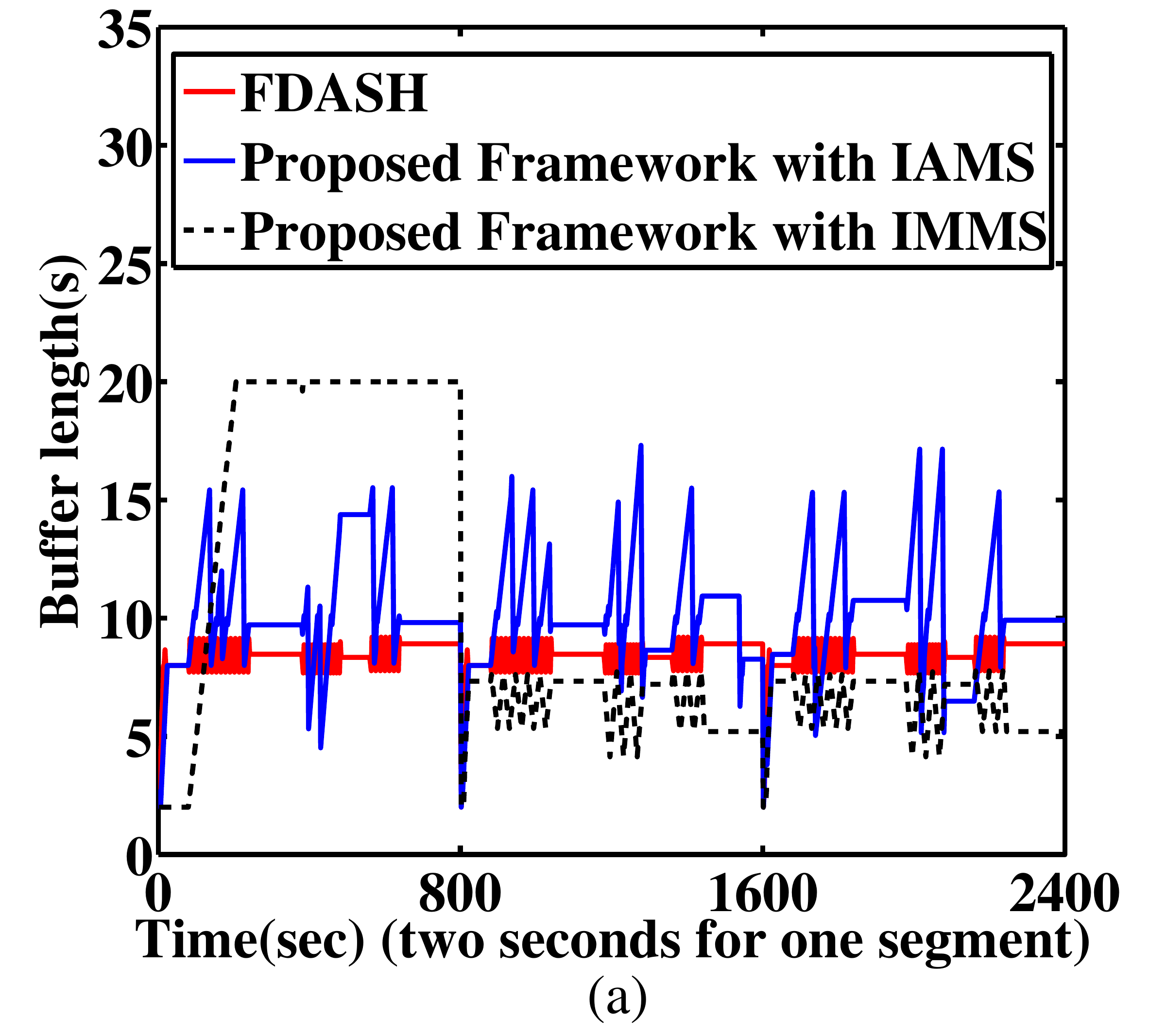}
\includegraphics[width=4.2cm]{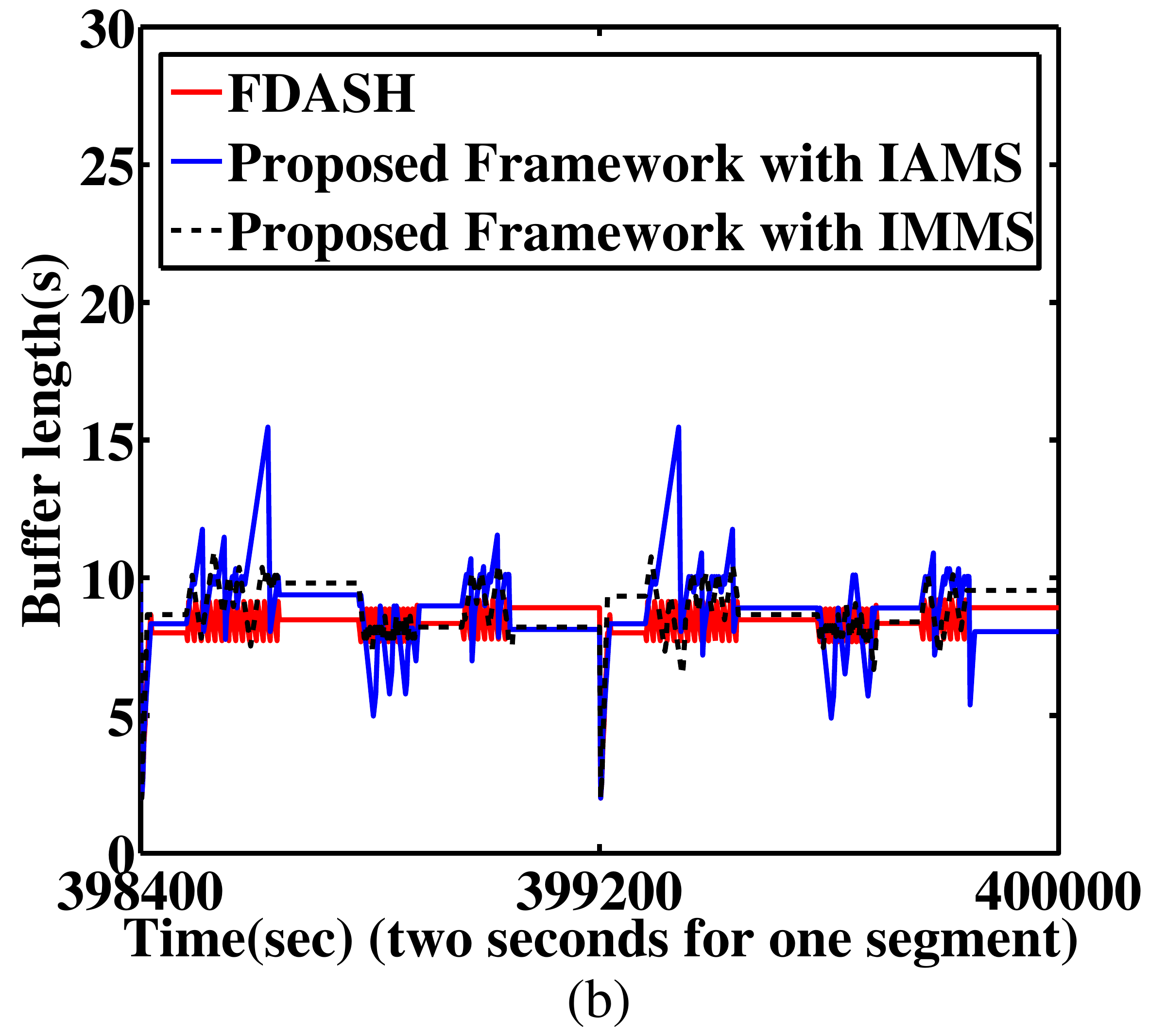}
\caption{Buffer length comparisons under the long-term fluctuating
channel.} \label{fig31}
\end{figure}

Lastly, we evaluate the performance of the rate-based method, the PD controller-based method, the online learning-based method, the \emph{\textbf{FDASH}}, the proposed framework with \textbf{\emph{IAMS}}, and the proposed framework with \textbf{\emph{IMMS}}, by using two extensively used \emph{QoE} models [43] and [45], as shown in TABLE \uppercase\expandafter{\romannumeral3}.

The \emph{QoE} model proposed in [43] can be written as
\begin{equation}\label{E13}
\begin{split}
QoE^{M}=\sum_{m=1}^{M}q(r_{m})-\lambda\sum_{m=1}^{M}|q(r_{m+1})-q(r_{m})|
\\-\mu\sum_{m=1}^{M}(T_{m}^{down}-B_{m}),
\end{split}
\end{equation}
where $\lambda=1$ and $\mu=6$ are model parameters that are
empirically defined in [43], $M$ is the number of received segments,
$r_{m}$ is the bitrate of the  $m$-$th$ requested video segment,
$q(r_{m})$ is the corresponding SSIM value, $T_{m}^{down}$ is the
download time of the $m$-$th$ segment, and $B_{m}$ is the buffer
length at the end time of the $m$-$th$ segment.

\begin{figure}
\centering
\includegraphics[width=8.76cm]{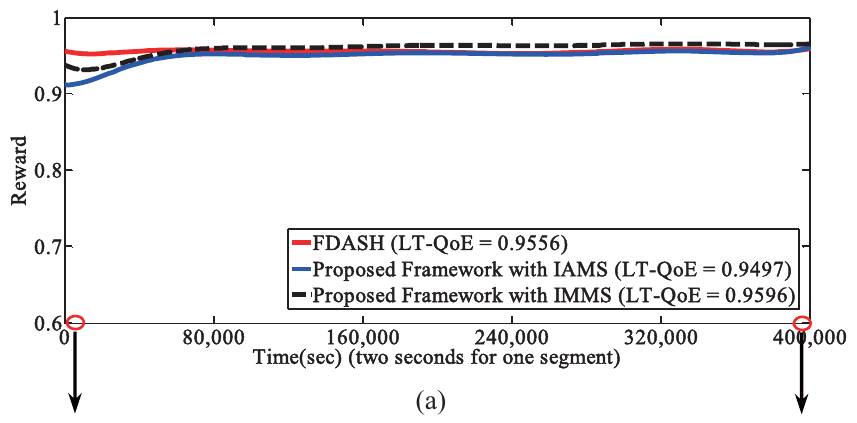}
\includegraphics[width=4.2cm]{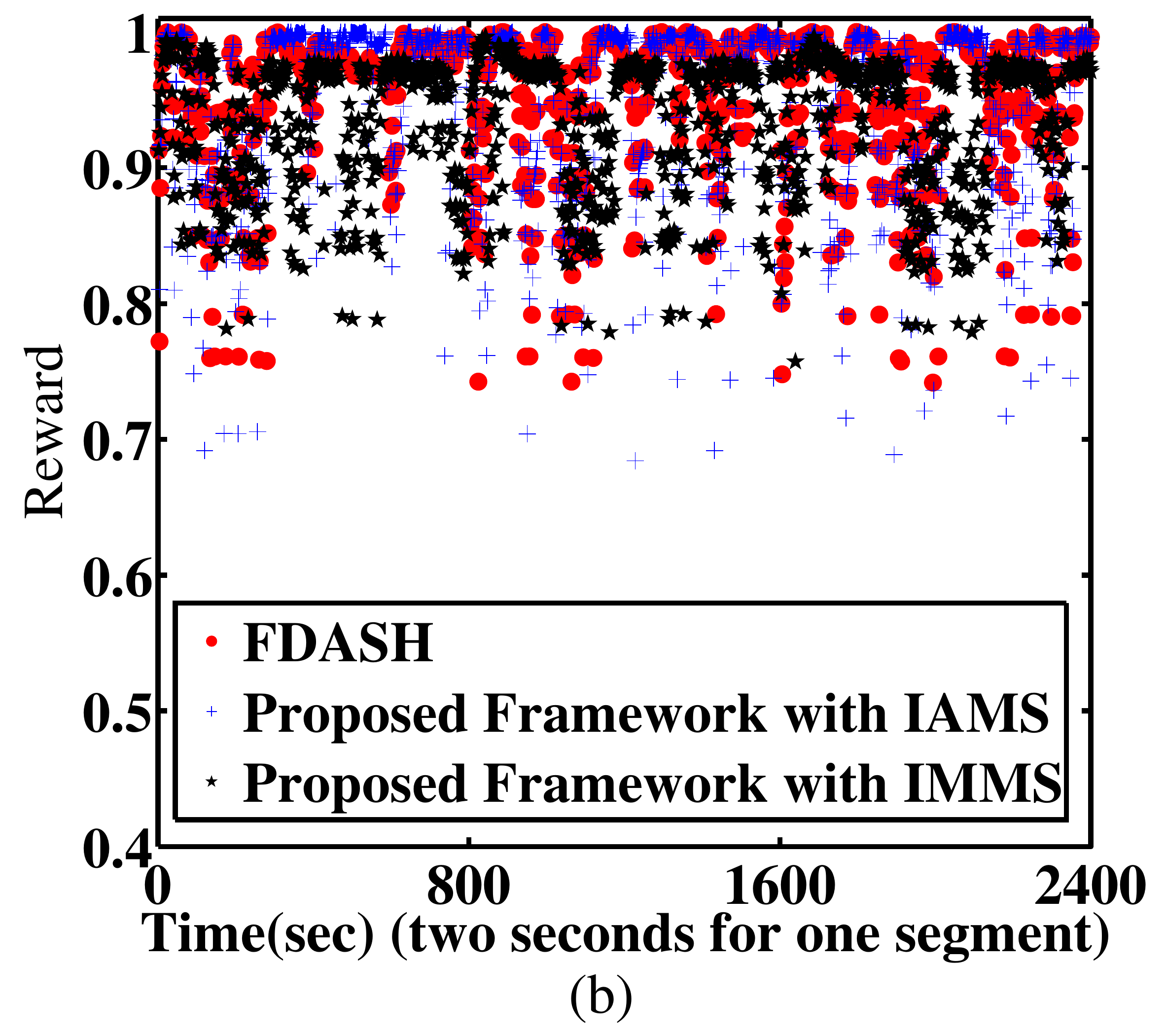}
\includegraphics[width=4.2cm]{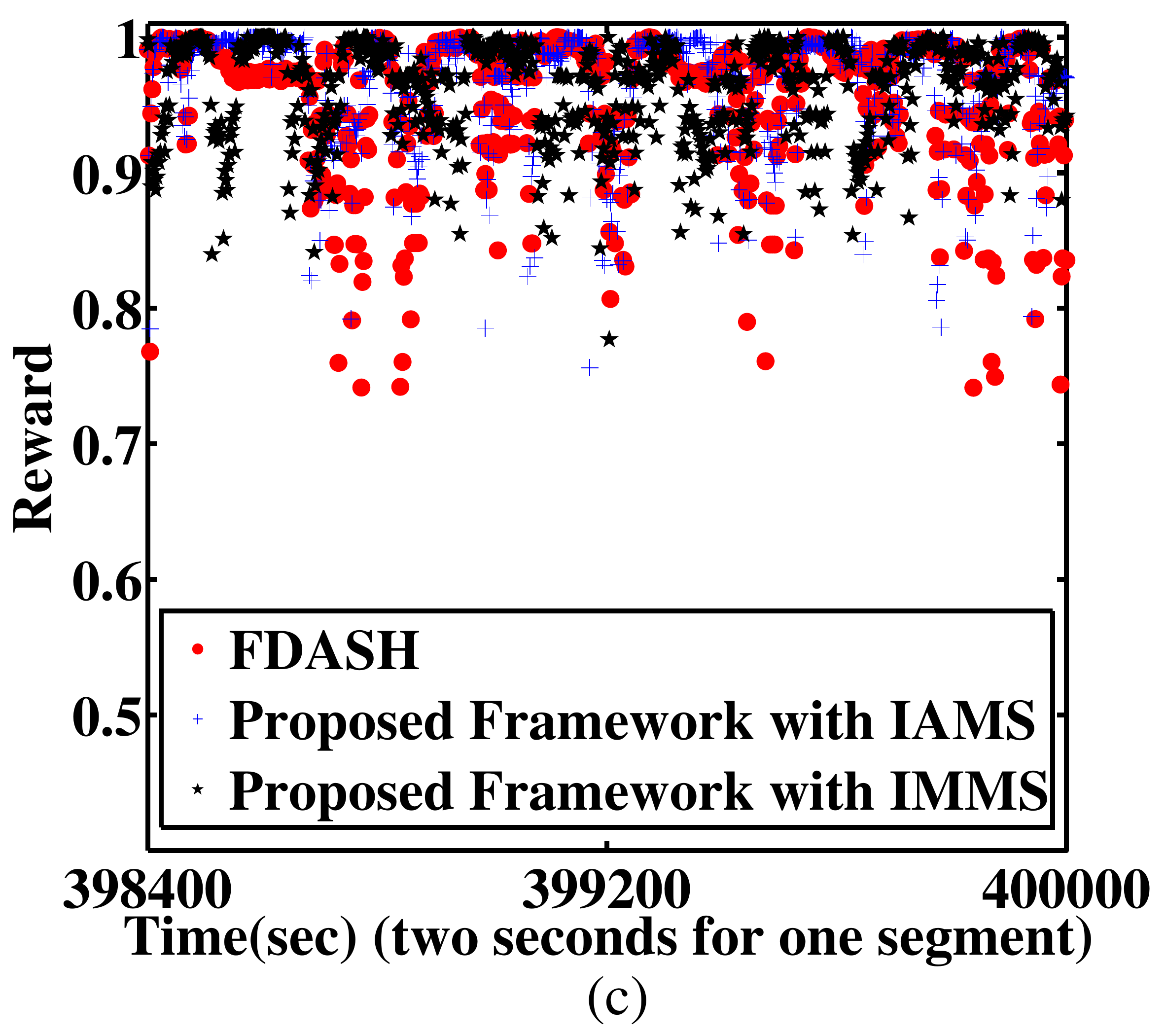}
\caption{Rewards and \emph{LT-QoE}s comparisons under the Markov
channel $(p=0.5)$.} \label{fig32}
\end{figure}

\begin{figure}
\centering
\includegraphics[width=4.2cm]{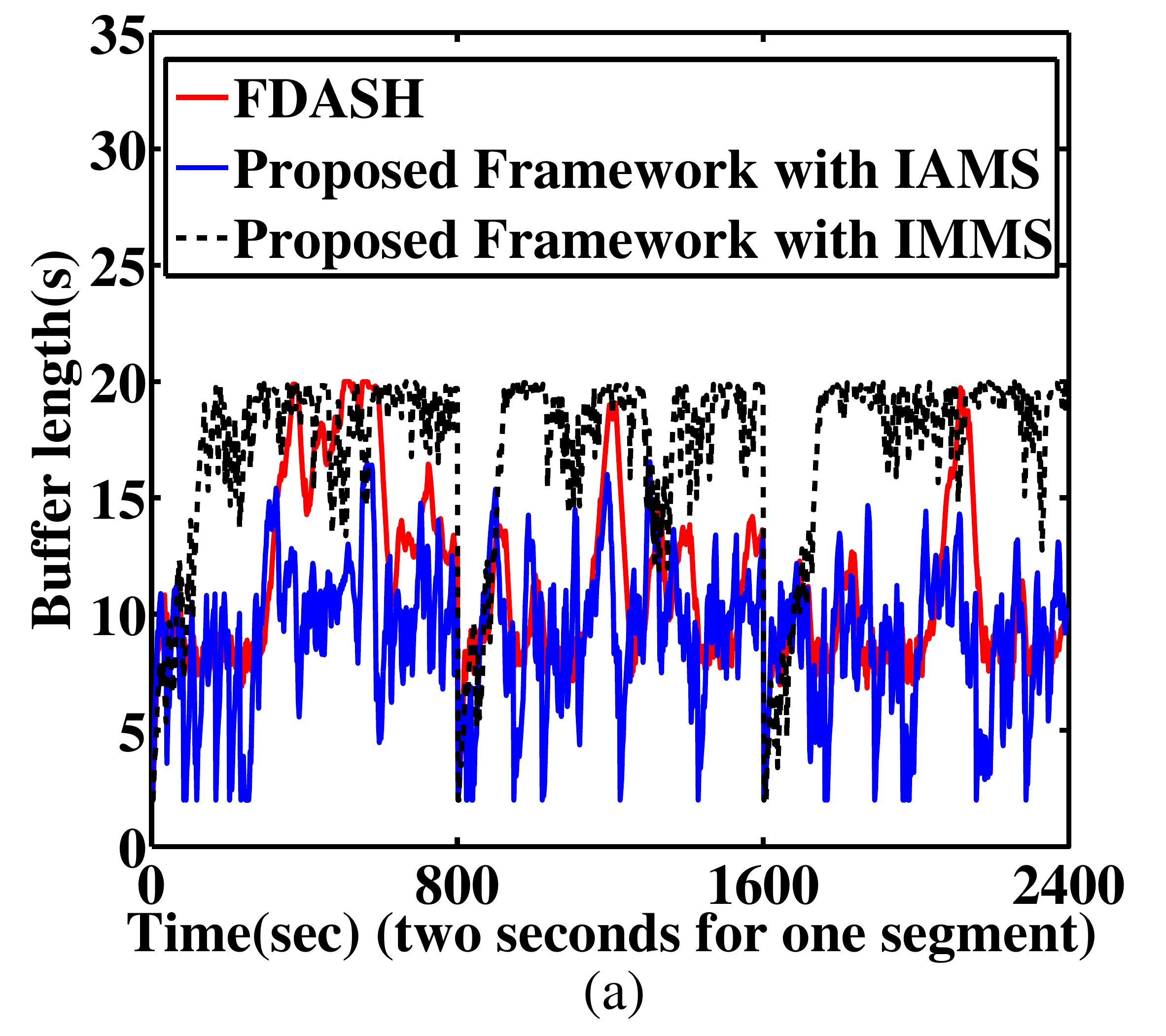}
\includegraphics[width=4.2cm]{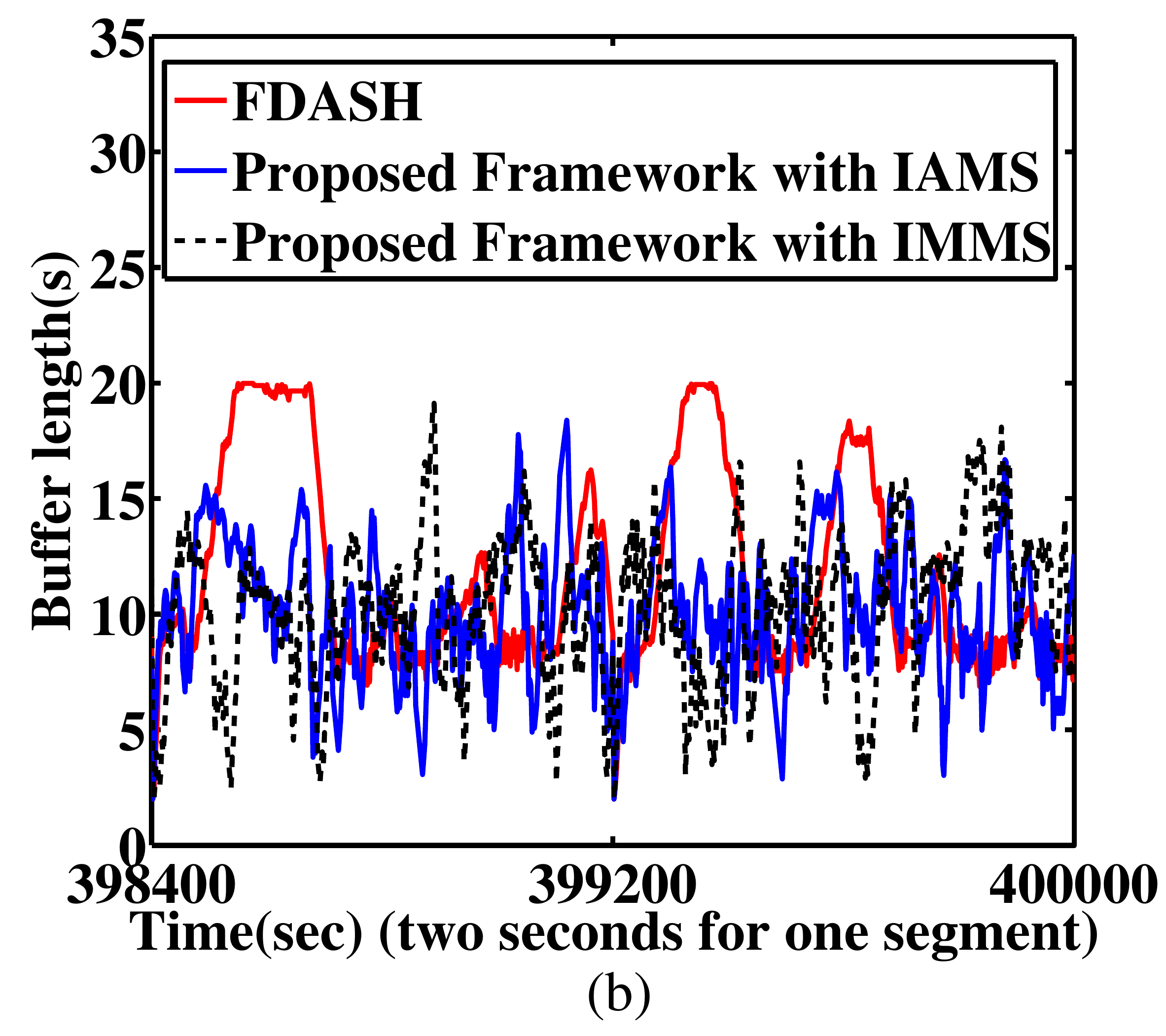}
\caption{Buffer length comparisons under the Markov channel
$(p=0.5)$.} \label{fig33}
\end{figure}

\begin{table}[htbp]
  \centering
  \caption{COMPARISONS RESULTS BY TWO EXTENSIVELY USED \emph{QoE} METRICS}
\includegraphics[width=8.76cm]{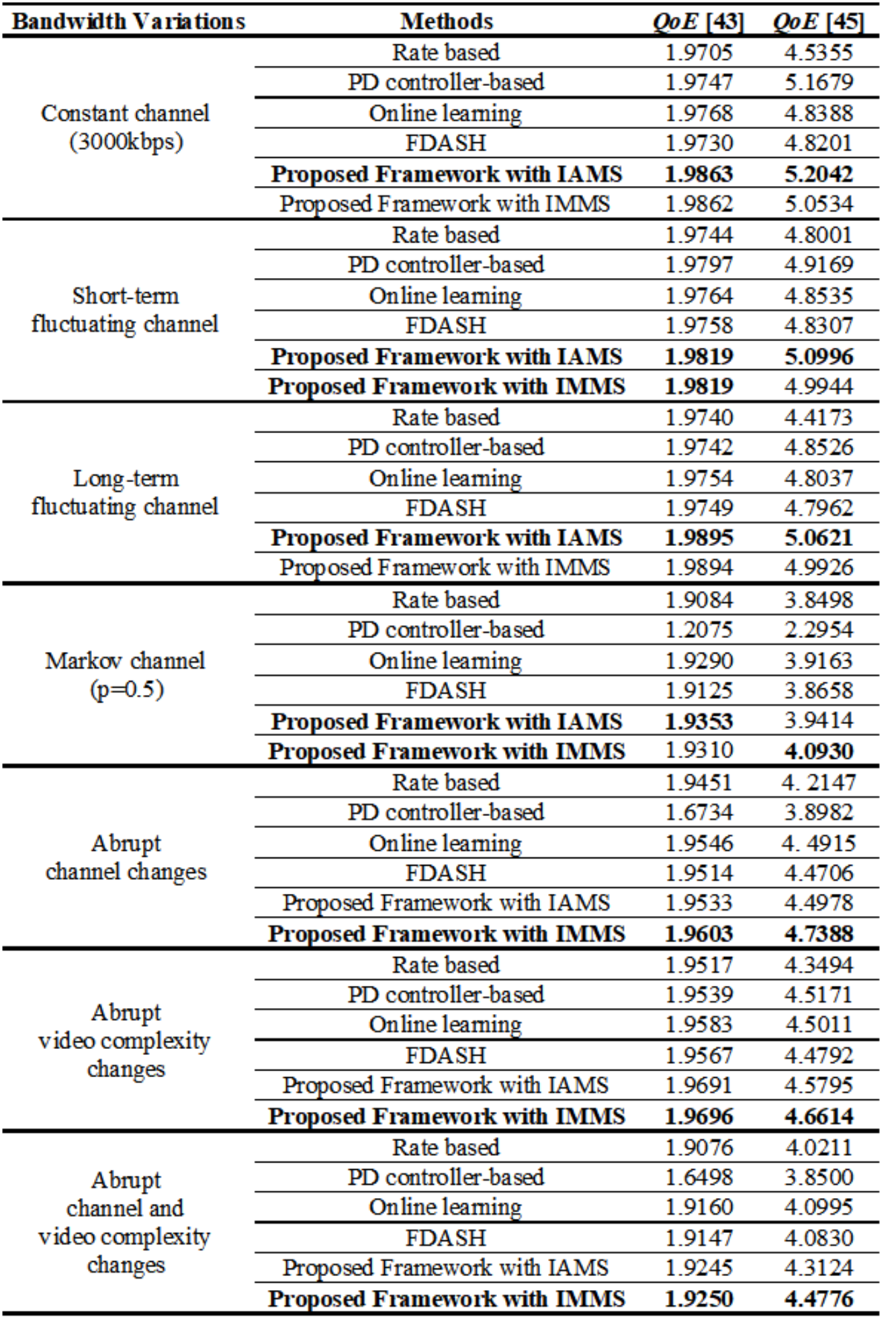}
\label{table3}
\end{table}

The \emph{QoE} model proposed in [45] can be written as
\begin{equation}\label{E14}
QoE=4.85\cdot Q_{norm}-4.95\cdot F-1.57\cdot S+0.5,
\end{equation}
where $Q_{norm}$ is the normalized average quality of all the received video segments,
\begin{equation}\label{E15}
Q_{norm}=\frac{1}{M\cdot Q_{max}}\cdot \sum_{m=1}^{M}Q_{m},
\end{equation}
$M$ is the number of requested video segments, $Q_{max}$ is the highest quality of all the received video segments, $F$ is a factor to evaluate the negative gain of playout interruption,
\begin{equation}\label{E16}
F=\frac{7}{8}\cdot \frac{\ln(f_{F}+1)}{6}+\frac{1}{8}\cdot
\frac{\min(f_{T},15)}{15},
\end{equation}
$f_{F}$ and $f_{T}$ denote freeze frequency and average freeze
duration respectively, $S$ in Eq. (14) is used to evaluate the
negative gain of video quality fluctuations,
\begin{equation}\label{E17}
S=\frac{sw_{number}}{M}\cdot \frac{sw_{depth}}{(Q_{max}-Q_{min})},
\end{equation}
where $sw_{number}$ and $sw_{depth}$ represent the number and the average amplitude of bitrate switches, respectively.


From TABLE \uppercase\expandafter{\romannumeral3}, we can see that
the proposed frameworks always produce the highest \emph{QoEs} with
respect to different types of channel bandwidth variations, no
matter what \emph{QoE} model is used. It can also be observed that
for more complex bandwidth conditions and video contents, the
performance of the proposed framework with \textbf{\emph{IMMS}} is
better than that of the proposed framework with \textbf{\emph{IAMS}}.

\section{Conclusions and Future Work}

We proposed an ensemble rate adaptation framework for
DASH clients by leveraging the advantages of multiple rate
adaptation methods to improve the \emph{QoE} of users. The proposed
framework consists of two modules: a method pool and a method
controller. At each method switching time, the method controller
will select the best method that can achieve the best reward from
the method pool. We also proposed two method switching strategies:
\textbf{\emph{IAMS}} and \textbf{\emph{IMMS}}, to select an
appropriate rate adaptation method. Three popular rate adaptation
methods (rate-based method, PD controller-based method, and online
learning-based method) are integrated into the proposed framework.
Simulation results have shown that the proposed framework can fully
use the advantages of the methods in the method pool, and always
provide the largest \emph{QoE} for users.

In the future, we will implement the algorithm in a more realistic scenario, based on a real test-bed which uses a realistic TCP transmission. The performance of the proposed ensemble learning-based rate adaptation framework largely depends on the integrated methods in the method pool. Therefore, we will add more advanced adaptation methods into the framework and improve the strategy of the method controller to adapt to more complex environments. Another future work is to cut down the system complexity and reduce the decision time as much as possible.

\section*{Acknowledgment}

The authors would like to thank Institute of Information Technology
(ITEC) at Klagenfurt University for the valuable and basis work of
DASH.

\ifCLASSOPTIONcaptionsoff
  \newpage
\fi



%

\end{document}